%% file: ms.tex
\documentclass[10pt,twocolumn,letterpaper]{article}

\usepackage{cvpr}
\usepackage{times}
\usepackage{epsfig}
\usepackage{graphicx}
\usepackage{amsmath}
\usepackage{amssymb}
\usepackage{amsfonts}
\usepackage{rotating}
\usepackage{amsthm}

\usepackage{pgfplots}
\usepackage{pgf,tikz}

\usetikzlibrary{calc,patterns,angles,quotes,positioning}

\usepackage{subcaption}
\usepackage{xcolor}
\usepackage[utf8]{inputenc}
\usepackage{overpic}
\usepackage{hhline}

\usepackage{dblfloatfix}

\newtheorem{theorem}{Theorem}

\newtheorem{myalg}[theorem]{Algorithm}

\newcommand{\mX}{\mathcal{X}}
\newcommand{\mY}{\mathcal{Y}}

\cvprfinalcopy % *** Uncomment this line for the final submission

\def\cvprPaperID{7320} % *** Enter the CVPR Paper ID here

\ifcvprfinal\fi

\begin{document}

%%%%%%%%% TITLE
\title{Smooth Shells: Multi-Scale Shape Registration with Functional Maps}

\author{Marvin Eisenberger
\and
Zorah L\"ahner\\
Technical University of Munich\\
%{\tt\small marvin.eisenberger@in.tum.de}
\and
Daniel Cremers\\
}

\maketitle
%\thispagestyle{empty}

%%%%%%%%% ABSTRACT
\begin{abstract}
    We propose a novel 3D shape correspondence method based on the iterative alignment of so-called smooth shells. Smooth shells define a series of coarse-to-fine shape approximations designed to work well with multiscale algorithms. The main idea is to first align rough approximations of the geometry and then add more and more details to refine the correspondence. We fuse classical shape registration with Functional Maps by embedding the input shapes into an intrinsic-extrinsic product space. Moreover, we disambiguate intrinsic symmetries by applying a surrogate based Markov chain Monte Carlo initialization. Our method naturally handles various types of noise that commonly occur in real scans, like non-isometry or incompatible meshing. Finally, we demonstrate state-of-the-art quantitative results on several datasets and show that our pipeline produces smoother, more realistic results than other automatic matching methods in real world applications.
\end{abstract}

	%-------------------------------------------------------------------------

	\section{Introduction}

    The wide selection of affordable 3D scanning devices in recent years has led to an enormous growth in the amount of 3D shapes and scans available.
    In contrast to synthetic shapes, real-world scans are often noisy and many properties cannot be guaranteed.
    For example, topological noise might appear in self-touching areas or the meshing density varies depending on scanning conditions.
    These distortions were proven to be difficult for state-of-the-art shape correspondence methods \cite{laehner2016shrec, melzi2019shrec}.
    Many traditional methods focus only on the (nearly) isometric case, clearly defined extensions of this like partiality \cite{litany2017fullyspectral}, or learn matching for different classes of shapes under certain perturbations \cite{groueix20183dcoded}. Unfortunately, this requires training data and knowledge about what deformations and noise are to be expected.

  In general one can distinguish between intrinsic and extrinsic correspondence methods. Intrinsic methods only use surface properties that are independent of the embedding, for example the Laplace-Beltrami operator.
  On the other hand, extrinsic methods directly use the 3D embedding of the shapes. 
  While intrinsic methods are invariant to large scale, near-isometric deformations, extrinsic alignment is often more suitable for pairs with topological changes or other non-isometric deformations.
  A natural step would be to combine both to get the best of both worlds but only few previous approaches venture in this direction \cite{bronstein2009topology,corman2017funcchara}.
  
  %Here, in order to combine intrinsic and extrinsic information we embed the input shapes in the product space of intrinsic (spectral) and extrinsic (xyz) coordinates. Then we iteratively align the two input shapes in this product space. We demonstrate that our method computes meaningful correspondences beyond the isometry assumption and in the presence of various types of noise.
  
  \paragraph{Contribution}
  In this paper we combine intrinsic and extrinsic information by embedding the input shapes into the product space of intrinsic (spectral) and extrinsic (xyz) coordinates. Then, we iteratively align smooth approximations of the two input shapes in this product space which we call smooth shells. Moreover, we propose a Markov chain Monte Carlo initialization strategy to find a meaningful local minimum and disambiguate self-similarities. Overall, we obtain a robust matching pipeline that works out of the box for a broad range of inputs beyond the isometry assumption and in the presence of various types of noise.

\begin{figure}
  \centering
 \includegraphics[width=.31\linewidth]{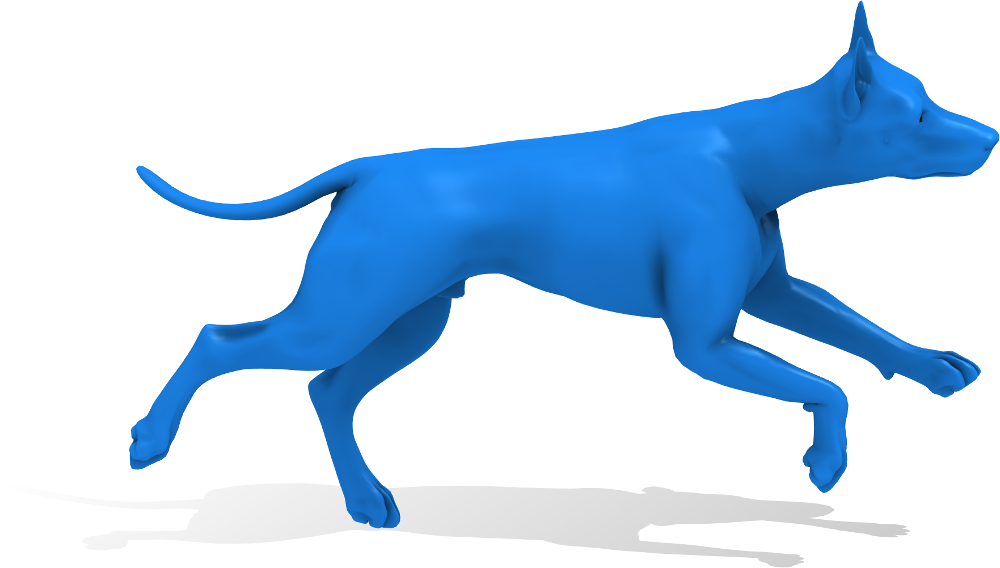}
 \includegraphics[width=.33\linewidth]{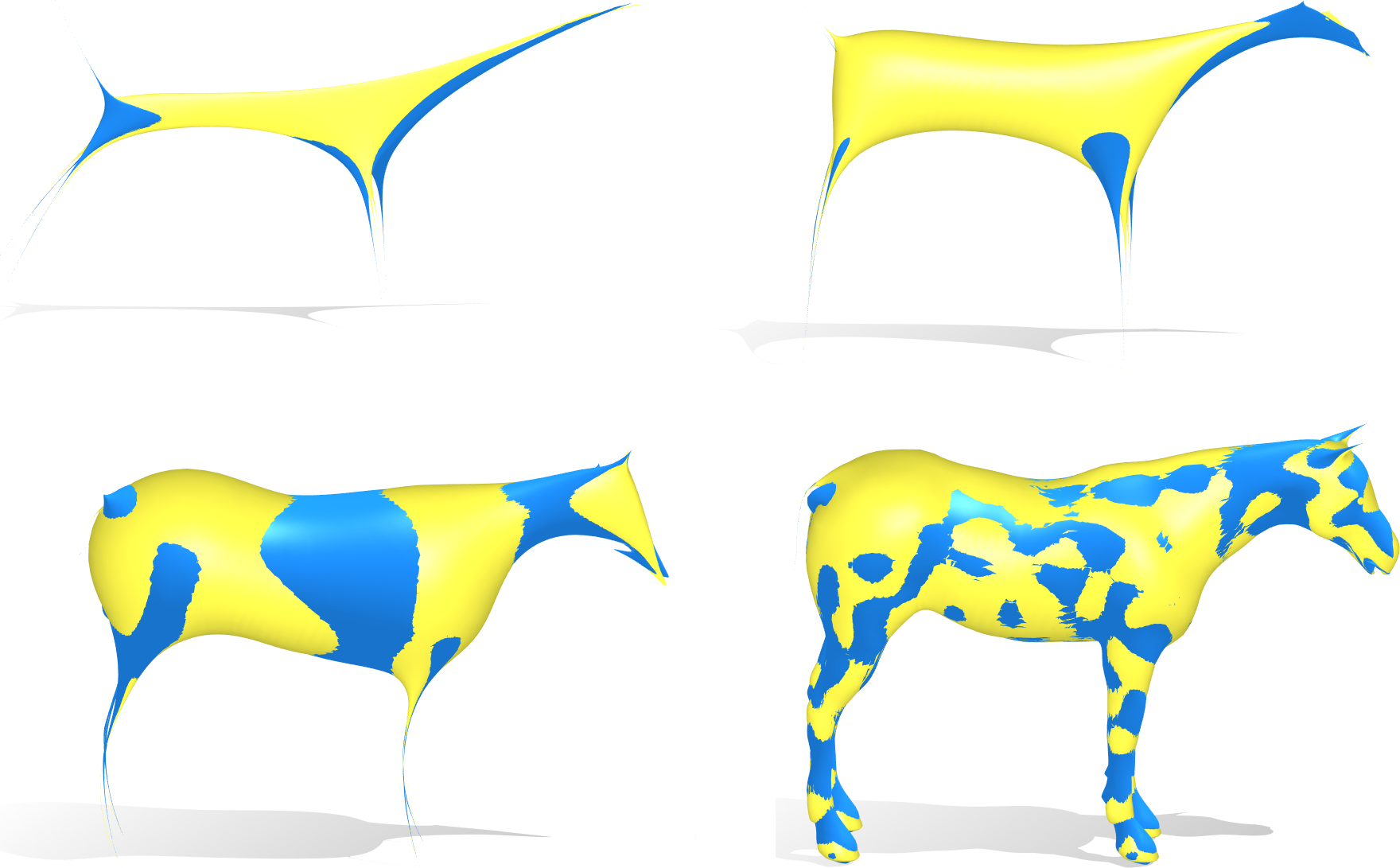}
 \includegraphics[width=.33\linewidth]{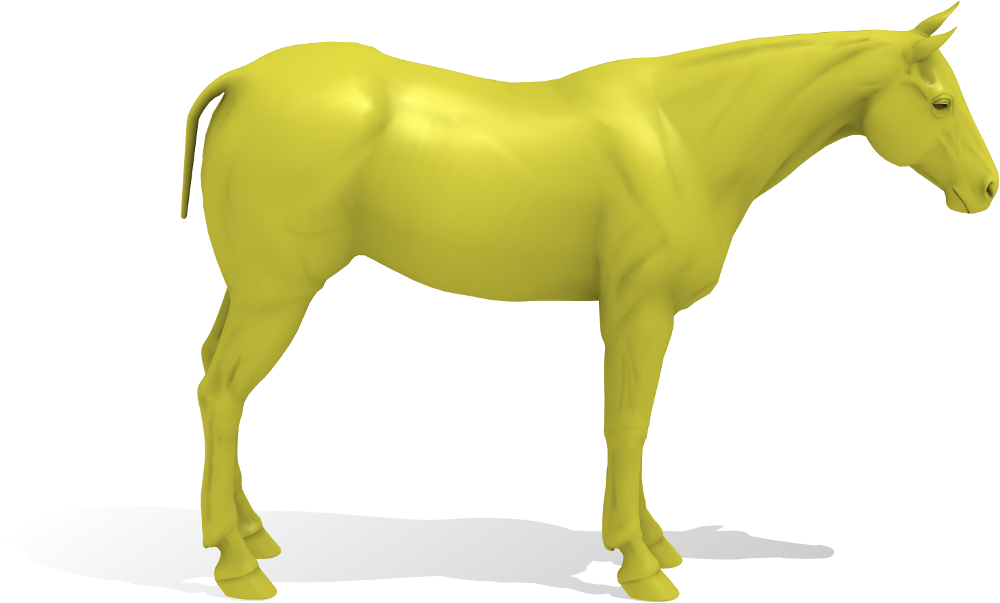}
  \caption{Given a source (left) and target (right) shape we propose a hierarchical smoothing procedure to iteratively align the inputs. First, we align very coarse approximations and then refine until we get correspondences for the original inputs.
  %In order to combine extrinsic and intrinsic shape information we concatenate the xyz coordinates with Functional Maps \cite{ovsjanikov2012functional}. 
  Among other things, we can handle challenging interclass pairs like matching a dog to a horse and our method is fully automatic, i.e. we do not use any additional information like hand-selected landmarks.}
\label{fig:teaser}
\end{figure}
  
%   \paragraph{Contribution}
  
%   We propose a coarse-to-fine approach to shape matching. In particular,
%   \begin{itemize}
%       \item we introduce smooth shells which are shape approximations with a smooth upsampling behavior.
%       \item we iteratively align those smooth shells in an extrinsic-intrinsic product space.
%       \item we propose a Markov chain Monte Carlo initialization strategy to disambiguate self-similarities.
%       \item we compute correspondences without any additional input information like annotated landmarks or an initial Functional Map.
%   \end{itemize}

  \section{Related Work} \label{sec:relatedwork}

  \subsection{Shape Correspondence and Registration}
  
  Shape correspondence is an extensively studied topic with various applications in Computer Vision and Graphics. 
  Surveys of state-of-the-art methods \cite{vankaick11correspsurvey,salvi2007rangereview,tam2013pointcloudsurvey,sahilliouglu2019recent} give a broader overview of existing approaches but here we focus on work that is immediately related to ours.
  
 The Functional Maps \cite{ovsjanikov2012functional} paper proposes an elegant formalism to model shape correspondences. The main idea is to model mappings of functions on the input shape to functions on the output shape instead of point-to-point maps. This allows for a compact matrix representation in a low rank basis. Over the last years, the original framework has been extended and applied to various applications \cite{rodola2016partial,litany2016puzzles,ren2018orientation}. A major challenge in this context is extracting a point-wise correspondence from a Functional Map \cite{rodola2015cpd}. Several methods extend the original formalism but most of them are computationally heavy or make restrictive assumptions about the inputs \cite{rodola2015cpd,ren2018orientation,nogneng2018productpreservation}. Another common approach is to take noise correspondences obtained from a Functional Map and denoise them \cite{ezuz2017deblur,vestner2017pmf}. However, this only works if the input map is sufficiently accurate. Finally, all the methods listed above are by design prone to produce faulty matches in the presence of intrinsic self-similarities.

  Extrinsic methods explicitly deform and align the input shapes in the 3D embedding space.  \cite{myronenko2010pointsetregistration,ma2014robust} model the deformation with a linear mapping in a low rank basis on the surface of one shape.
  Like our approach, \cite{eisenberger2019divergence} alternates between calculating a deformation field and correspondences but the volume-preservation constraint restricts the applicability.
  Many deformation-based methods require expensive preprocessing to apply the deformation model, for example with a deformation graph \cite{sumner2007embedded}, structural rods \cite{alhashim2017topology} or deep learning \cite{groueix20183dcoded}.
  Non-rigid ICP methods iteratively align shapes but they rely on a good initialization \cite{li2008nonrigidicp,amberg2007nricp}. However, for many applications we do not have such a previous alignment and in general there is no trivial way to obtain it.

  There exist accurate methods to register certain classes of shapes, e.g. humans \cite{bogo2017dyna}.
  Unfortunately, these are highly specialized and depend on class specific features \cite{marin2018farm}, or learn statistical models from data \cite{ponsmoll2015dyna}.
  While these methods perform extremely well for shapes within their classes, they usually do not generalize to arbitrary examples.

  \subsection{Shape Approximation and Simplification}
    The idea of mesh simplification by smoothing is investigated thoroughly in previous work. \cite{vallet2008spectral} use manifold harmonics for the smoothing.
    In surface deformation modeling this is usually a two stage algorithm.
    First, a smoothed version of a shape is deformed and then the details are added back to the surface, see \cite{botsch2007linear} for an overview.
    Some classical works on shape modeling with smoothing are \cite{guskov1999multiresolution} and \cite{kobbelt1998interactive}.
    \cite{botsch2006deformation} combines this approach with differential coordinates.
    Although our smooth shells are related to smoothing technique like \cite{vallet2008spectral}, none of the mentioned approaches use a series of approximations. We propose a novel hierarchical shape smoothing method that is particularly suitable for coarse-to-fine matching. 

  Shape skeletons offer a lower-dimensional description of the rough geometry of a shape. A recent survey of 3D skeleton methods can be found in \cite{tagliasacchi2016skeletons}. Although the skeletons are usually designed to be easily aligned between different shapes from similar classes, most methods typically only define a single, unique skeleton for each shape. This is useful for a rough matching but does not allow for an iterative refinement of the surface alignment. Similar to our method, \cite{cao2010pcskeletons} extracts a skeleton based on Laplacian-based contraction but aims at getting a unique curve skeleton. Some methods exist to create an entire class of skeletons for each shape \cite{reniers2008curves}. Our method differs from the previously mentioned in that we do not introduce a fixed skeleton for each shape. Instead we construct a whole class of approximations with an increasing level of detail.

	\section{Background}
    A correspondence between two input shapes $\mathcal{X}$ and $\mathcal{Y}$ is defined as a point-to-point mapping $\mathbf{P}:\mathcal{X}\to\mathcal{Y}$. 
    Here, a shape is a 2D Riemannian manifold with an embedding in  $\mathbb{R}^3$. We use triangular meshes to discretize the surfaces $\mathcal{X}$ and $\mathcal{Y}$ and denote the coordinate matrices as $X\in\mathbb{R}^{N \times 3}$ and $Y\in\mathbb{R}^{M \times 3}$ with $N$ and $M$ vertices respectively.
    
    %Furthermore, we consider the space of $L^2$ integrable functions on $\mathcal{X}$ denoted as $L^2(\mathcal{X})=\bigl\{f:\mathcal{X}\to\mathbb{R}\hspace{2pt}\big|\hspace{2pt}\|f\|^2=\langle f,f\rangle<\infty\bigr\}$.

	\subsection{Laplace-Beltrami Operator}
	The Laplace-Beltrami operator $\Delta=\text{div}(\nabla\cdot)$ is an extension of the standard Euclidean Laplacian to manifold domains $\mathcal{X}$.
    Computing solutions of $\Delta\phi_k=\lambda_k\phi_k$ yields the Laplace-Beltrami eigenfunctions $\{\phi_k\}_{k\in\mathbb{N}}$ which form an orthonormal basis of $L^2(\mathcal{X})$. This allows for a spectral representation of functions $f \in L^2(\mathcal{X})$:
    %Therefore, a truncated sum projecting on the first $K$ eigenfunctions $\phi_1,\dots\phi_K$ gives a low-frequency approximation that, according to the min-max principle is optimal for smooth functions $f \in L^2(\mathcal{X})$ \cite{parlett1998symmetric}.
	\begin{equation}
	\label{eq:spectraldecomptrunc}
	f\approx \tilde{f}=\sum_{k=1}^{K}\bigl\langle f,\phi_k\bigr\rangle \phi_k.
	\end{equation}
	According to the min-max principle, $\tilde{f}$ is an optimal compact approximation of smooth functions $f \in L^2(\mathcal{X})$ \cite{parlett1998symmetric} with a fixed basis size $K$.
	To compute the Laplace-Beltrami operator on triangular meshes, we use a cotangent discretization $\Delta\in\mathbb{R}^{N\times N}$ with lumped mass matrix \cite{pinkall1993discrete} and we denote its first $K$ eigenvectors as $\Phi_K=(\phi_1,\dots,\phi_N)\in\mathbb{R}^{N\times K}$ (analogously $\Psi_K\in\mathbb{R}^{M\times K}$ for $\mY$).
	
	\begin{figure*}
        \centering
        \includegraphics[width=\linewidth]{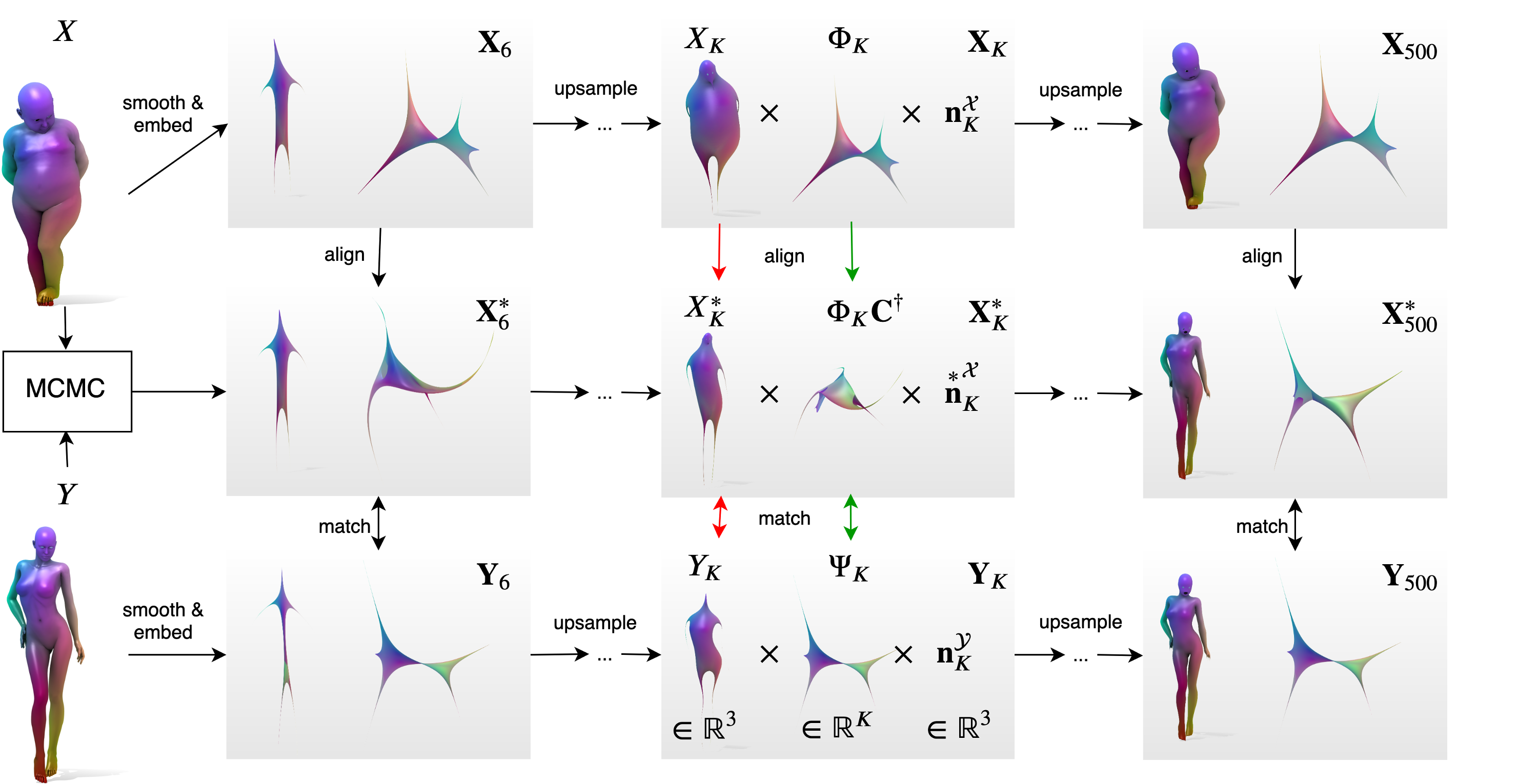}
        \caption{Overview of our pipeline. (Column 1) We initialize our method with the alignment $X^*_6$ from our Markov chain Monte Carlo initialization algorithm, see Section~\ref{sec:initial} for details. (Columns 2-4) On each level $K$ we embed both shapes in the $K+6$ dimensional product space of the smoothed extrinsic coordinates $X_K\in\mathbb{R}^{3}$, the intrinsic spectral coordinates $\Phi_K\in\mathbb{R}^{K}$ and the outer normals $\mathbf{n}^\mathcal{X}_K\in\mathbb{R}^{3}$. In this space we can align $\mathbf{X}_K$ and $\mathbf{Y}_K$ by computing an extrinsic morphing $\tau\in\mathbb{R}^{K\times 3}$ for $X_K$ and a Functional Map $\mathbf{C}\in\mathbb{R}^{K\times K}$ for $\Phi_K$. To visualize the spectral embedding $\Phi_K$ only the first three dimensions are shown. Finally, using the aligned $\mathbf{X}^*_K$ and $\mathbf{Y}_K$ we obtain a point-to-point matching $\mathbf{P}:\mathcal{X}\to\mathcal{Y}$ with a nearest neighbor search in $\mathbb{R}^{K+6}$. We repeat this process for $50$ iterations with smoothing levels on a logarithmic scale between $K=6:500$. Each iteration is initialized with the previous alignment.}
        \label{fig:diagramoverview}
    \end{figure*}

	\subsection{Functional Maps}\label{subsec:functionalmaps}

    The Functional Map framework \cite{ovsjanikov2012functional} is a popular approach to solve for correspondences $\mathbf{P}:\mathcal{X}\to\mathcal{Y}$.
    In Functional Maps $\mathbf{P}$ is replaced with a mapping of functions to functions $\mathcal{C}:L^2(\mathcal{X})\to L^2(\mathcal{Y})$.
    $\mathcal{C}$ is linear and can therefore be compactly written as a matrix $\mathbf{C}\in\mathbb{R}^{K\times K}$:
	\begin{equation}
	\label{eq:fmspectral}
	\mathcal{C}=\Psi_K\mathbf{C}\Phi_K^\dagger.
	\end{equation}
	
	To compute $\mathbf{C}$ for a pair of input shapes we need additional information to constrain the solution. Given pairs of corresponding functions $f_i\in\mathbb{R}^N$ and $g_i\in\mathbb{R}^M$ on the two surfaces an energy to optimize for $\mathbf{C}$ is: 
	\begin{equation}
	\label{eq:featureenergy}
	E_\mathrm{feat}(\mathbf{C}):=\|\mathbf{C}\Phi_K^\dagger F-\Psi_K^\dagger G\|_F^2.
	\end{equation}
	
	Here, $F,G$ are matrices whose columns are the feature functions $f_i,g_i$.
    Possible choices for those features range from pointwise descriptors or surface texture to input landmarks. 
    Another common assumption is that the mapping $\mathbf{P}$ is area preserving which leads to orthogonal Functional Maps $\mathbf{C}^\top\mathbf{C}=\mathbf{I}$, see \cite[Theorem 5.1]{ovsjanikov2012functional}.
    
    %
	%Throughout this paper it becomes clear that functional correspondences are the most natural way to encode similarities between our smooth shells because they are themselves represented in a spectral basis of the same dimension.
    %Like this we merely use Functional Maps as a tool to encode similarities between our shells but do not entirely rely on intrinsic information.

    \subsection{Shape deformation}
    
    A different approach is to align the surfaces in the embedding space instead of calculating the correspondence directly. 
    We denote the deformed version of $\mX$ with $\mX^*$ and impose that $\mX^*$ should align with $\mY$. 
    A common choice model is a linear displacement in a low rank basis \cite{myronenko2010pointsetregistration,ma2014robust}, e.g. the Laplace-Beltrami eigenbasis:
    \begin{equation}
        \label{eq:deformation}
        X^*=X+\Phi_K\tau.
    \end{equation} 
    
    $\tau\in\mathbb{R}^{K\times 3}$ are some displacement coefficients that parameterize the deformation. In the discrete case, the pointwise correspondence is represented by the matrix $\mathbf{P}\in\{0,1\}^{M\times N}$ with $\mathbf{P}^\top\mathbf{1}=\mathbf{1}$. Using the aligned shape $X^*$, we can recover $\mathbf{P}$ by minimizing the following energy:
    \begin{equation}
    \label{eq:alignmentextrinsic}
        E_\mathrm{align,3D}(\mathbf{P}):=\|\mathbf{P}X^*-Y\|_F^2.
    \end{equation}
    
    This is equivalent to a nearest neighbor search in $\mathbb{R}^3$. In order to get a meaningful correspondence with this approach we need to additionally regularize the deformations $\mX^*$. One possibility is to assume that the deformations are as-rigid-as-possible on a local scale:%To restrict the solutions to more meaningful correspondences we can impose additional assumptions about the deformation of $\mX^*$. One possible assumption is that the deformation is as-rigid-as-possible on a local scale \cite{sorkine2007rigid}:
    \begin{multline}
        \label{eq:arapenergy}
        E_\mathrm{arap}(\tau):=\int_\mX\int_{\mathcal{N}(x)}\bigl\|R(x)\bigl(X(x)-X(y)\bigr)-\\\bigl(X^*_\tau(x)-X^*_\tau(y)\bigr)\bigr\|_2^2\mathrm{d}y\mathrm{d}x.
    \end{multline}
    
    $\mathcal{N}(x)$ denotes the neighborhood of $x\in\mX$ and $R:\mX\to\mathrm{SO}(3)$ describes the local rotation, for details see \cite{sorkine2007rigid}.

    \section{Method} \label{sec:method}

    We propose to compute shape correspondences by iteratively aligning a series of coarse-to-fine approximations of the input surfaces $\mathcal{X}$ and $\mathcal{Y}$.
    This is based on the idea that the alignment of two shapes can end up in unwanted local optima in the presence of non-consistent small scale features. In many cases the rough structure of $\mathcal{X}$ and $\mathcal{Y}$, like the number of extremities, is similar while the fine scale details can differ, see Figure~\ref{fig:teaser}. After matching the global features, the local features can be used to refine the alignment.
    The smooth shells we use as coarse shape approximations are defined in Section~\ref{sec:smoothshells}. Section~\ref{subsec:intrinsicextrinsicembedding} explains how we combine extrinsic and intrinsic shape embeddings and Section~\ref{subsec:hierarchicalmatching} defines our complete matching algorithm.

    %For the first iteration, when there is no initialization from the previous iteration, we incorporate a surrogate based initialization strategy to compute an initial alignment of the coarse shells.
    %This strategy also avoids mixing self-similarities like intrinsic symmetries which are easily mistaken already in the initial state.
    %See Section~\ref{subsec:mcmc} for details.

        %\begin{figure}
         %   \centering
         %   \includegraphics[width=\linewidth]{figures/diagram_method.png}
        %    \caption{The figure above displays the main steps in our registration pipeline. Details for the individual steps are provided in the respective subsections of Chapter~\ref{sec:method}.}
        %    \label{fig:diagrammethod}
        %\end{figure}

	\subsection{Smooth Shells} \label{sec:smoothshells}

    In this section, we propose a novel shape smoothing operator $\mathcal{S}_K$ that yields smoothed shapes  similar to those from spectral surface reconstruction \cite{levy2010spectral}. In comparison, our operator leads to smoother transitions between $\mathcal{S}_K$ and $\mathcal{S}_{K+1}$ which makes it more suitable for a hierarchical alignment.

    \paragraph{Spectral Reconstruction}

    Spectral reconstruction \cite{levy2010spectral} smoothes $\mX$ by projecting its coordinate function $X$ onto the first $K$ Laplace-Beltrami eigenfunctions:
    \begin{equation}
	\label{eq:projection}
	T_K := \mathcal{T}_K(X) = \sum_{k=1}^{K}\bigl(\phi_k\otimes\phi_k\bigr)X.
	\end{equation}
    Here, $\phi_k\otimes\phi_k$ denotes the outer product of $\phi_k$ with itself which results in the projection of $X$ onto $\phi_k$.
    Since the eigenfunctions are ordered by frequency, this creates a coarse-to-fine approximation of the original shape. The level of detail is controlled by the number of eigenfunctions $K$. For small $K$ only the rough geometry is reconstructed, whereas for $K\to\infty$, $\mathcal{X}_K$ converges to the original $\mathcal{X}$.% See Figure~\ref{fig:shellexample} for an example.

    \paragraph{Shell Operator}\label{subsec:shelloperator}

    The gradual smoothing from Eq.~\eqref{eq:projection} is useful for hierarchical shape matching. In each iteration we increase $K$ and use the alignment from the previous iteration as an initialization. However, in many cases the refinement with spectral reconstruction leads to undesirable artifacts. Especially the first few $K$ projections from Eq.~\eqref{eq:projection} cause large disparities between reconstructions. This makes the alignment from the previous iteration less useful for the next step. We introduce the \emph{shell operator} $\mathcal{S}_K$ to circumvent this issue:
    \begin{equation}
	\label{eq:shelloperator}
	X_K:=\mathcal{S}_K(X):=\sum_{k=1}^{\infty}\frac{1}{1+\exp\bigl(\sigma(k-K)\bigr)}\bigl(\phi_k\otimes \phi_k\bigr)X.
	\end{equation}
	
    Just like spectral reconstruction, $\mathcal{S}_K$ smooths $X$ using a projection on $\phi_k$. However, instead of truncating the spectral coordinates at a certain $K$, we introduce a gradual truncation with sigmoid weights. Those are close to $1$ if $k\ll K$ and decay to $0$ when $k\gg K$.
    This guarantees that the displacement from $\mathcal{S}_K$ to $\mathcal{S}_{K+1}$ is reasonably bounded. For small $\sigma$ the transition becomes smoother, whereas for $\sigma\to\infty$ the sigmoid function converges to the indicator function $\mathbf{1}_{\{k\leq K\}}$ which corresponds to spectral reconstruction. In particular, we can show the following smoothness result for $\mathcal{S}$:
    \begin{theorem}\label{thm:smoothness}(Transition smoothness of $\mathcal{S}$)\\
    Let $\mathcal{X}$ be a shape with coordinate function $X\in L^2$, then the geometric difference of state $X_{K}$ and $X_{K+1}$ is bounded by the upsampling variance $\sigma$ in following way:
    \begin{equation}
    \frac{\bigl\|\mathcal{S}_{K+1}(X)-\mathcal{S}_{K}(X)\bigr\|_{L^2}}{\bigl\|\mathcal{S}_{K+1}(X)\bigr\|_{L^2}}\leq|1-e^{-\sigma}|=\mathcal{O}(\sigma),\sigma\to 0.
    \end{equation}
    \end{theorem}
    
    We provide a proof in Appendix~\ref{appendix:proof1}.
    See Figure~\ref{fig:shellexample} for an illustration of the practical implications and Table~\ref{fig:ablation} for a quantitative comparison to spectral reconstruction.
    
    \begin{figure}
    \begin{minipage}{0.47\linewidth}
        \centering
        \begin{overpic}
            [width=.47\linewidth]{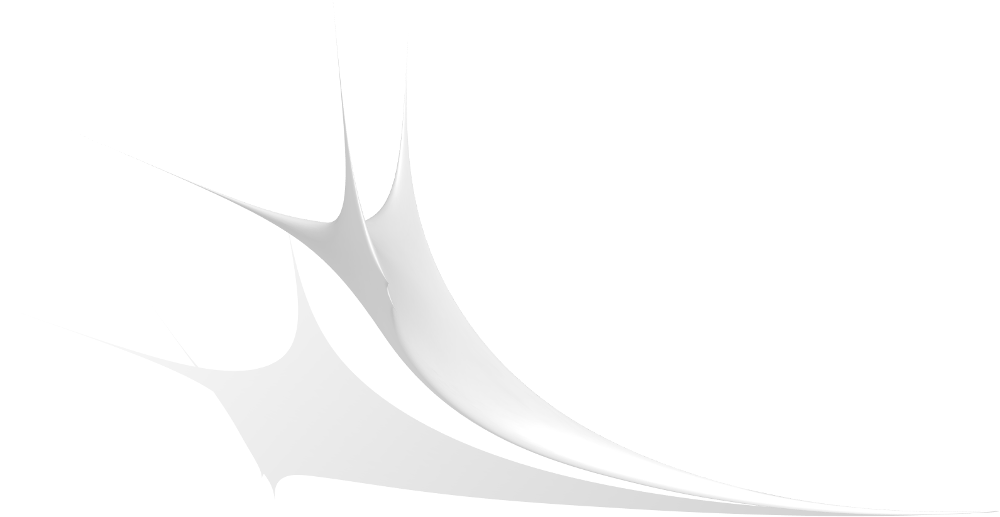}
            \put(60,35){$T_6$}
        \end{overpic}
        \begin{overpic}
            [width=.47\linewidth]{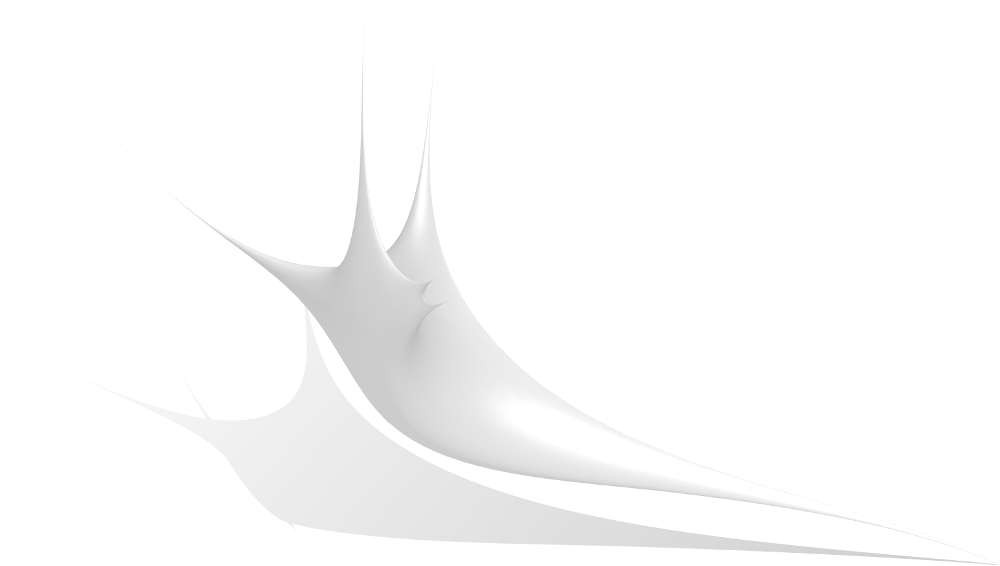}
            \put(60,35){$T_7$}
        \end{overpic}
        \begin{overpic}
            [width=.47\linewidth]{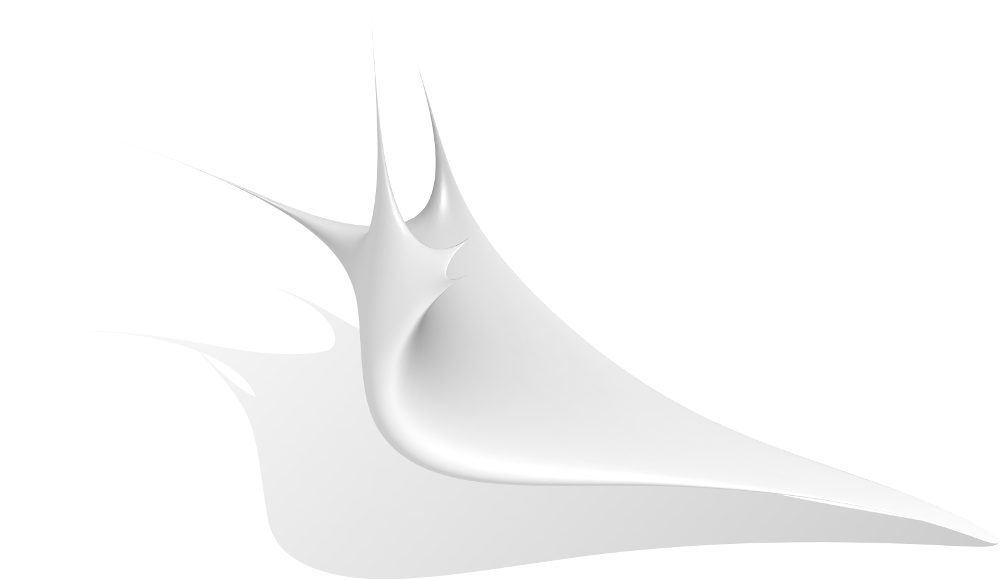}
            \put(60,35){$T_{8}$}
        \end{overpic}
        \begin{overpic}
            [width=.47\linewidth]{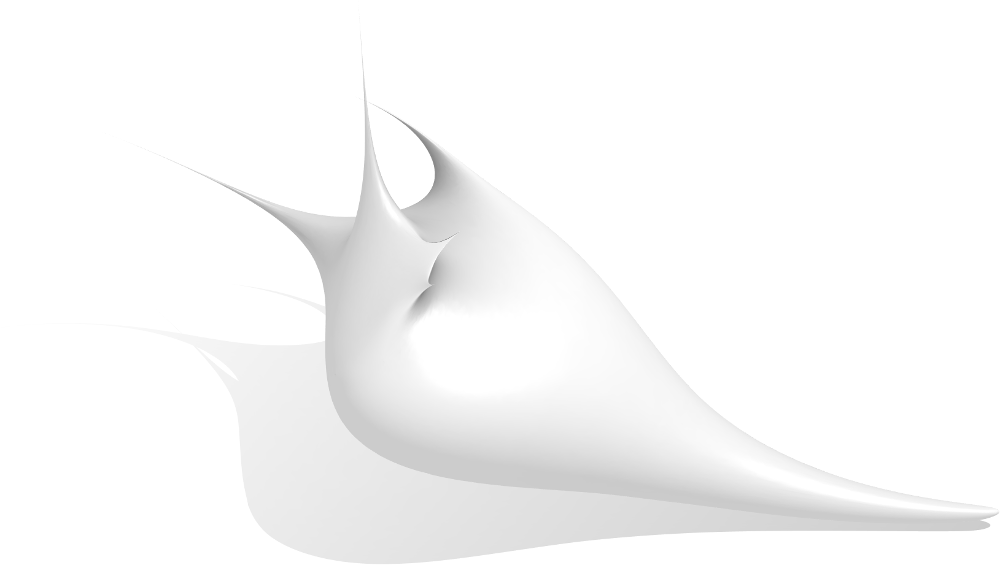}
            \put(60,35){$T_{9}$}
        \end{overpic}
        \begin{overpic}
            [width=.47\linewidth]{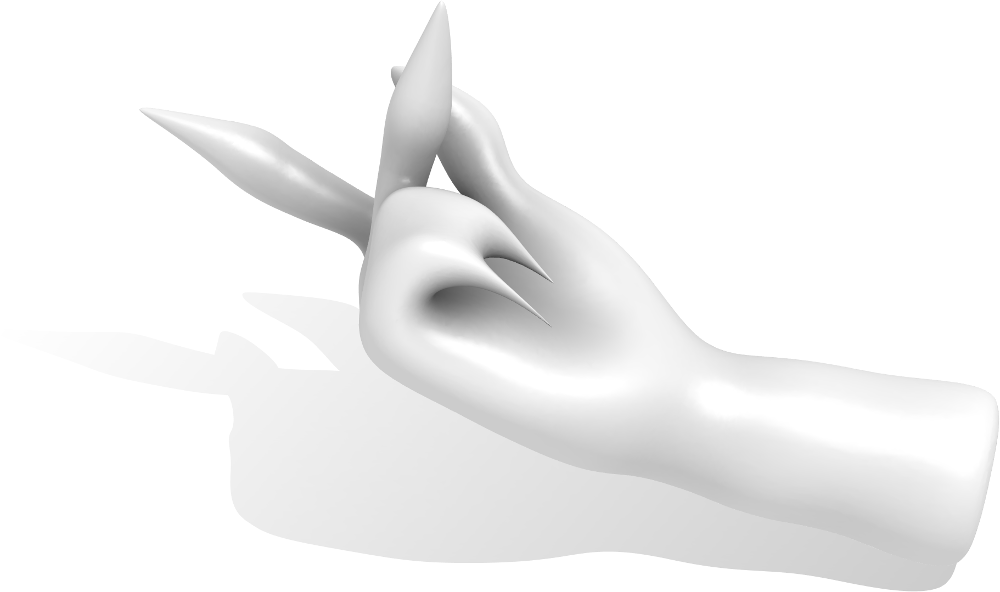}
            \put(60,45){$T_{100}$}
        \end{overpic}
        \begin{overpic}
            [width=.47\linewidth]{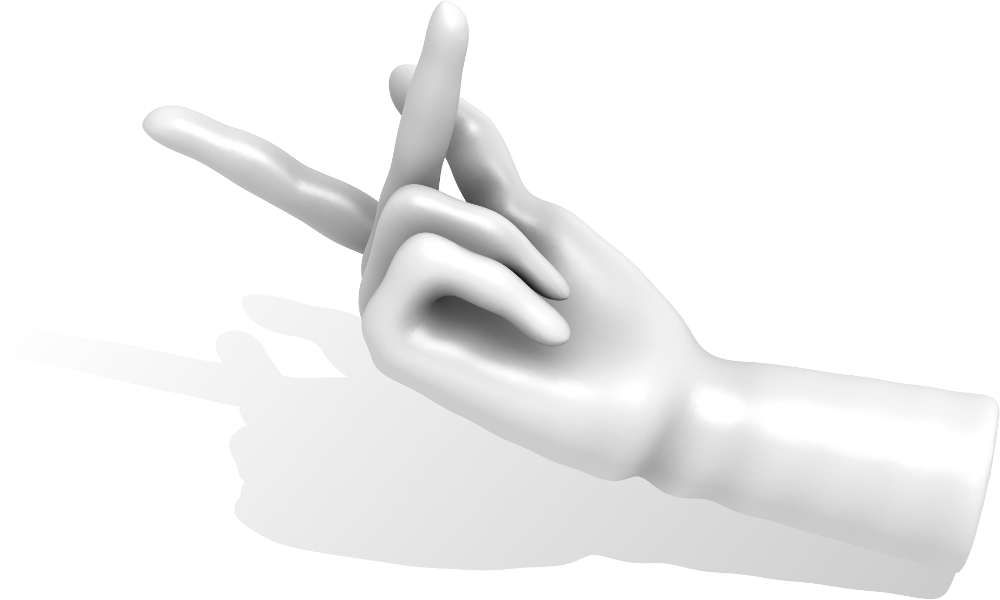}
            \put(60,45){$T_{500}$}
        \end{overpic}
    \end{minipage}~
     \rule[-45pt]{1pt}{100pt}
    \begin{minipage}{0.47\linewidth}
        \centering
        \begin{overpic}
            [width=.47\linewidth]{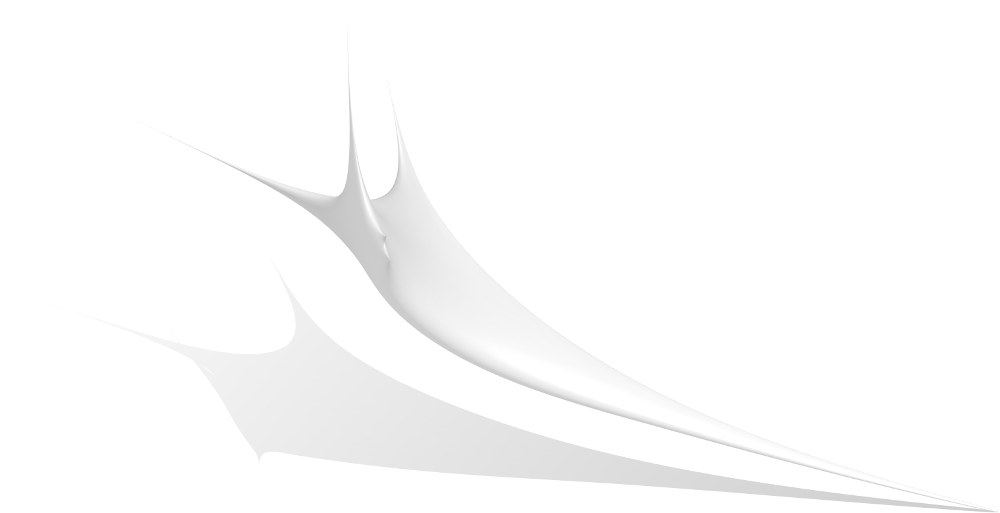}
            \put(60,35){$X_6$}
        \end{overpic}
        \begin{overpic}
            [width=.47\linewidth]{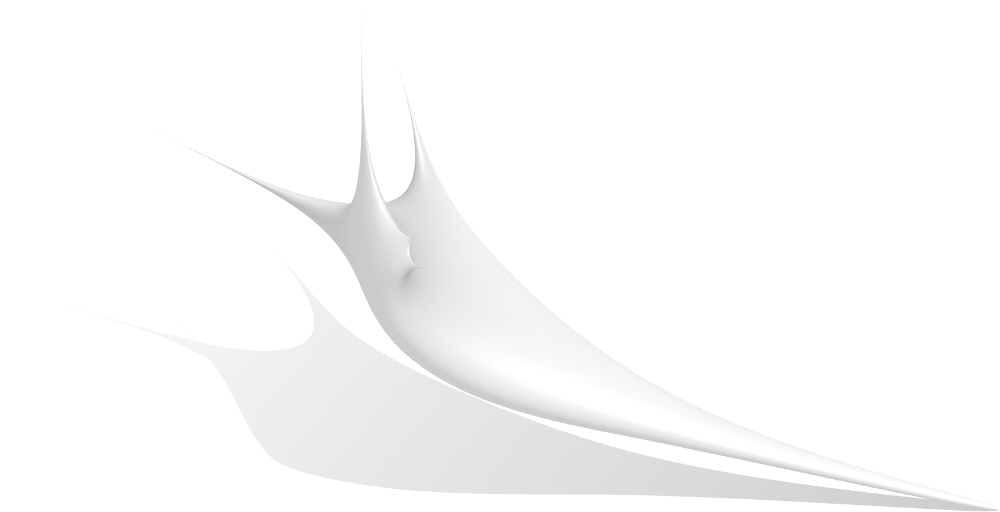}
            \put(60,35){$X_7$}
        \end{overpic}
        \begin{overpic}
            [width=.47\linewidth]{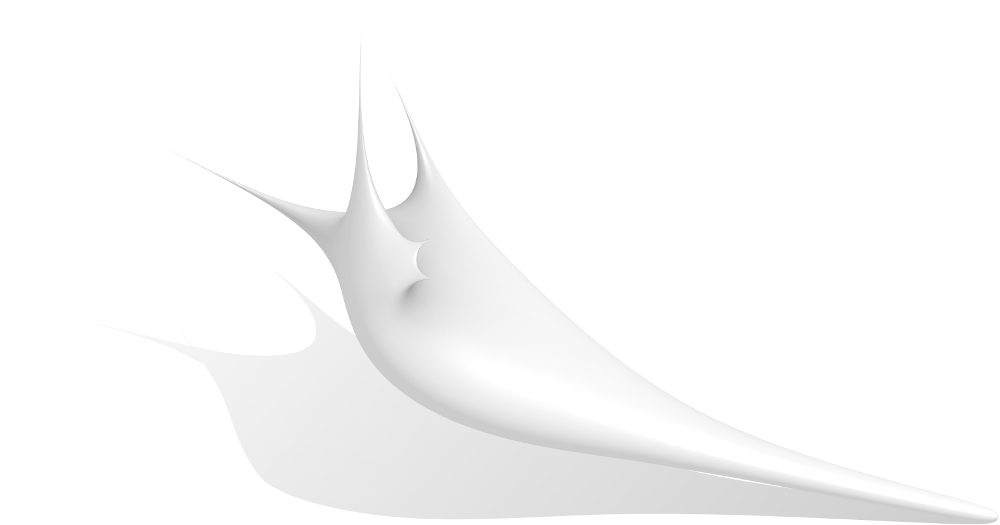}
            \put(60,35){$X_{8}$}
        \end{overpic}
        \begin{overpic}
            [width=.47\linewidth]{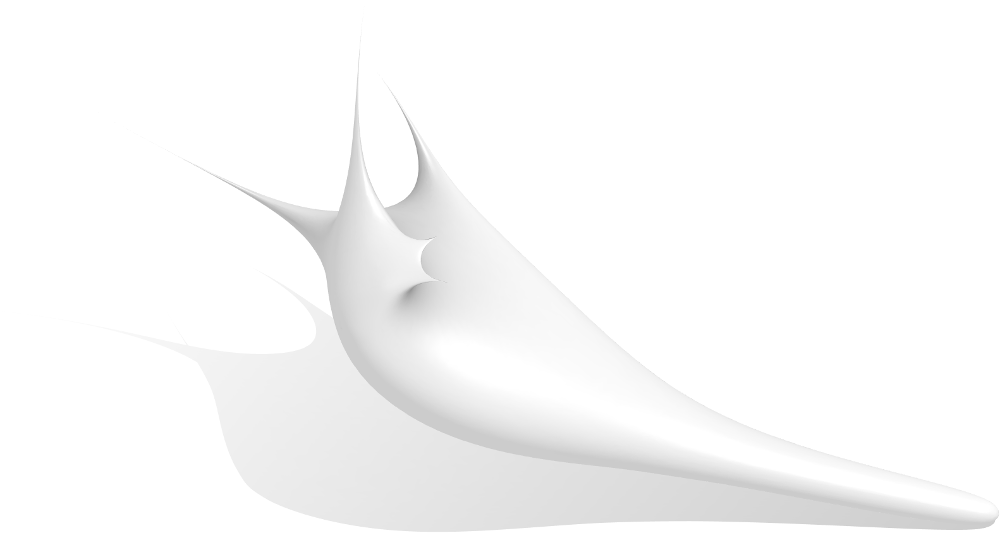}
            \put(60,35){$X_{9}$}
        \end{overpic}
        \begin{overpic}
            [width=.47\linewidth]{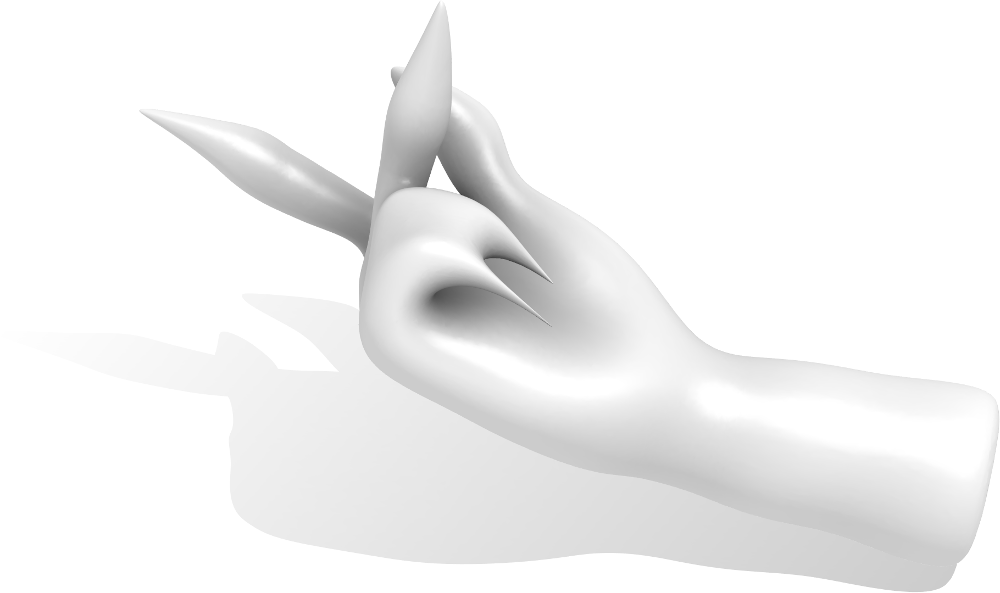}
            \put(60,45){$X_{100}$}
        \end{overpic}
        \begin{overpic}
            [width=.47\linewidth]{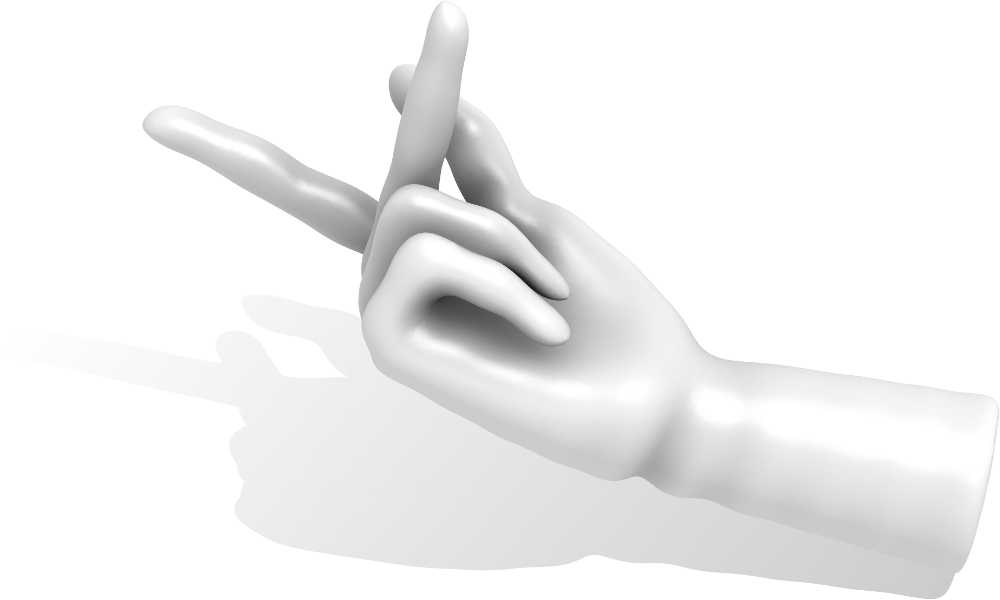}
            \put(60,45){$X_{500}$}
        \end{overpic}
    \end{minipage}
        \caption{At first glance there is no significant difference between  spectral reconstruction $T_K$ and smooth shells $X_K:=\mathcal{S}_K(X)$. They both converge to $X$ for $K\to\infty$, and for high indices $K\gg 50$ they are indistinguishable. The crucial difference lies in their upsampling behaviour. While smooth shells transition smoothly from $X_K$ to $X_{K+1}$, consecutive shapes $T_K$ tend to have large displacements and are therefore less suitable for iterative alignment methods.}
        \label{fig:shellexample}
    \end{figure}
    
    \begin{figure*}
	    \includegraphics[width=0.78\linewidth]{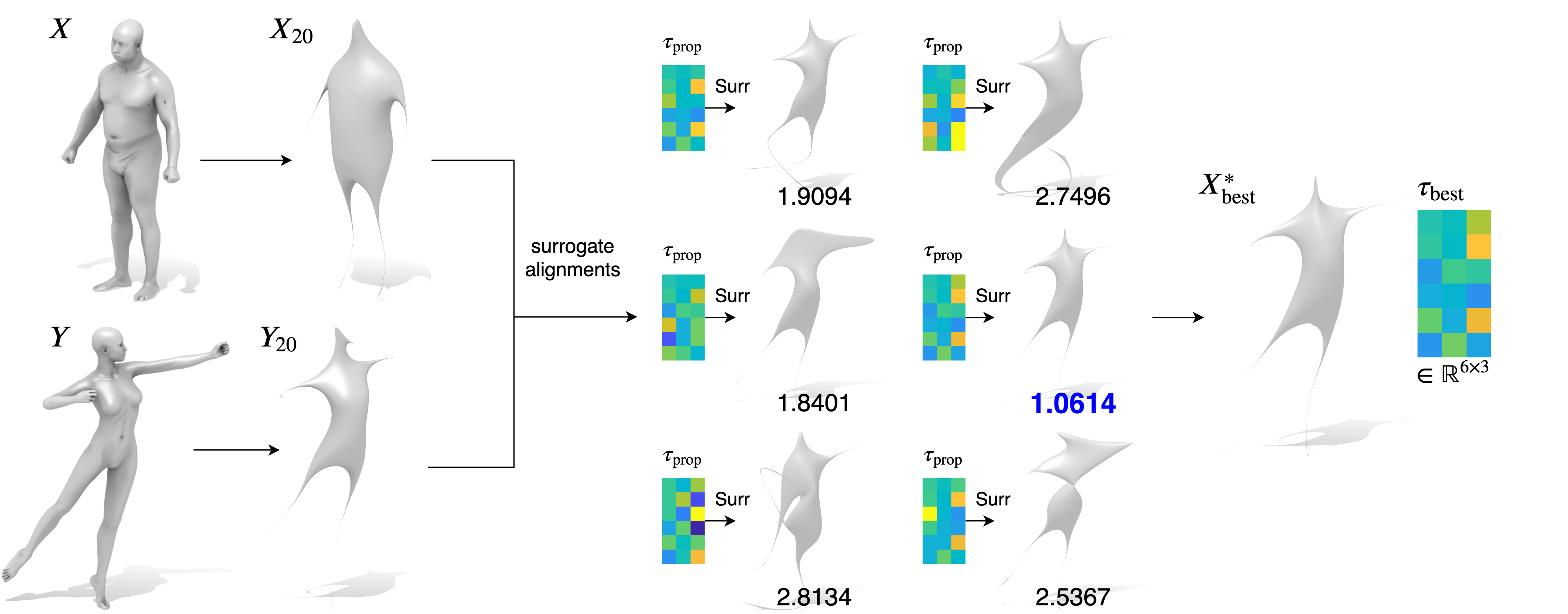}
	    \input{figures/MDS_MCMC.tikz}
		%\caption{The plot above shows an example distribution of all $100$ alignments $\tau_{K_\mathrm{max}}$ that are produced by the surrogate runs in Algorithm~\ref{alg:MCMC}. The distribution here stems from the example pair of humans that is also shown in Figure~\ref{fig:surrogate}. Each circle represents one deformation $\tau_{K_\mathrm{max}} \in \mathbb{R}^{{K_\mathrm{max}}\times 3}$. In the MCMC algorithm we choose ${K_\mathrm{max}}=20$, therefore $\tau_{K_\mathrm{max}}$ is a 60-dimensional vector. For visualization purposes we embed $\tau_{K_\mathrm{max}}$ into the 2D space with Multidimensional Scaling. For each proposal alignment the quality of the obtained alignment is measured using the energy in Eq.~\eqref{eq:energytotal} (see also Figure~\ref{fig:surrogate}). Small objective values $E(\mathbf{X}^*_{K_\mathrm{max}})$ are displayed with big, yellow circles whereas high objectives lead to smaller, blue circles. The cluster on the left side corresponds to the optimal initial alignment.}
		\caption{(left) Overview of of our MCMC initialization method. We sample potential initial alignments $\tau_\mathrm{prop}\in\mathbb{R}^{6\times 3}$ and rate them using surrogate runs (see Section~\ref{sec:initial}). Each proposal $\tau_\mathrm{prop}$ is assigned a mark $E(\mathrm{Surr}(\tau_\mathrm{prop}))>0$ based on the alignment quality of the current surrogate. In the shown example, the best objective is $E=1.0614$ and indeed this sample visually shows the tightest alignment of ${X}^*_{20}$ and ${Y}_{20}$. (right) 2D embedding with multi-dimensional scaling of all $\tau_\mathrm{prop}$ used for one initialization. Samples with small objective values have big, yellow circles and big objectives correspond to small, blue circles. Evidently there is a big cluster around the optimal (yellow) circle which shows that our algorithm is able to determine the optimal initialization with high confidence.}
		\label{fig:MCMC_overview}
	\end{figure*}

    \subsection{Intrinsic-extrinsic Embedding}\label{subsec:intrinsicextrinsicembedding}
    
    Intrinsic and extrinsic methods are often depicted as opposing viewpoints and, although there are some notable exceptions \cite{bronstein2009topology,corman2017funcchara}, only few methods try to combine them. 
    Our deformation model combines shape alignment in both intrinsic and extrinsic space. 
    Functional Maps is based on rigid ICP alignment of the spectral coordinates $\Phi_K\in\mathbb{R}^{N\times K}$ of $\mathcal{X}$ and $\Psi_K\in\mathbb{R}^{M\times K}$ of $\mathcal{Y}$ in the $K$-dimensional spectral domain \cite{ovsjanikov2012functional}.
    \begin{equation*}
        \Phi_K\mathbf{C}^\dagger\approx\Psi_K.
    \end{equation*}
    On the other hand, extrinsic methods typically align the 3-dimensional geometry as described in Eq.~\eqref{eq:deformation}:
    \begin{equation*}
        X^*_K=X_K +\Phi_K\tau\approx Y_K.
    \end{equation*}
    
    We combine intrinsic and extrinsic alignment in order to gain both their advantages. To this end, we embed the inputs $\mathcal{X}$ and $\mathcal{Y}$ in the product space of the \emph{intrinsic} (spectral coordinates $\Phi_K$) and the \emph{extrinsic} (smoothed Cartesian coordinates $X_K$ and outer normals $\mathbf{n}^\mathcal{X}_{K}$ of $\mathcal{X}$) coordinates: 
    \begin{subequations}
		\label{eq:productspace}
		\begin{equation}
		\label{eq:productspaceX}
		\mathbf{X}_K:=\begin{pmatrix}
		\Phi_K,X_{K},{\mathbf{n}}^\mathcal{X}_{K}
		\end{pmatrix}\in\mathbb{R}^{N\times (K+6)}.
		\end{equation}
		\begin{equation}
		\label{eq:productspaceY}
		\mathbf{Y}_K:=\begin{pmatrix}\Psi_K,Y_{K},\mathbf{n}^\mathcal{Y}_{K}
		\end{pmatrix}\in\mathbb{R}^{M\times (K+6)}.
		\end{equation}
	\end{subequations}
    
    Using the normals makes the embedding more descriptive because they convey information about the inside-outside orientation of each point.
    Using both the Functional Map $\mathbf{C}$ and the extrinsic deformation $\tau$ (see Eq.~\eqref{eq:deformation}) now yields the morphed embedding $\mathbf{X}^*_K$:
    \begin{equation}
        \label{eq:productspacedeformed}
        \mathbf{X}^*_K:=\begin{pmatrix}
		\Phi_{K}{\mathbf{C}}^{\dagger},X_K+\Phi_K\tau,{\overset{*}{\mathbf{n}}}^\mathcal{X}_{K}
		\end{pmatrix}\in\mathbb{R}^{N\times (K+6)}.
    \end{equation}
    
    ${\overset{*}{\mathbf{n}}}^\mathcal{X}_{K}$ are the normals of $X_K+\Phi_K\tau$. The next section will go into detail on how to compute $C$ and $\tau$.
    
    %\paragraph{Remark}
    %Our embedding is motivated by the observation that purely intrinsic or purely extrinsic methods only make use of %either one of those domains -- we put both approaches together.
    %The Functional Maps framework is based on a rigid ICP alignment algorithm of $\Phi_K\in\mathbb{R}^{N\times K}$ %and $\Psi_K\in\mathbb{R}^{M\times K}$ in the $K$-dimensional spectral domain \cite{ovsjanikov2012functional}.
    %\begin{equation*}
    %    \mathbf{P}\Phi_K\mathbf{C}^\dagger\approx\Psi_K.
    %\end{equation*}
    %In contrast to that, extrinsic correspondence methods typically align the 3-dimensional geometry by deforming the %first input shape $X_K$, see Eq.~\eqref{eq:deformation}:
    %\begin{equation*}
    %    \mathbf{P}X^*_K=\mathbf{P}\bigl(X_K +\Phi_K\tau\bigr)\approx Y_K.
    %\end{equation*}
    
    \begin{figure*}[b]
        \centering
        \input{figures/curve_tosca.tikz}
        \input{figures/curve_scape.tikz}
        \input{figures/curve_topkids.tikz}
        \input{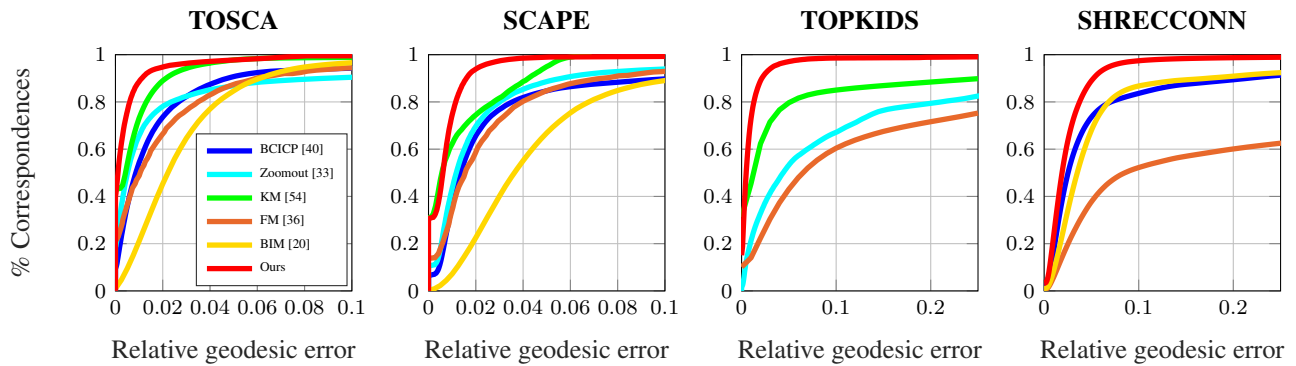}
        \caption{Our matching accuracies for four datasets in comparison to other popular fully automatic shape correspondence methods. For more details on the datasets, see Section~\ref{subsec:shapecorrespondence}. }
        \label{fig:matching_curves}
    \end{figure*}
    
    \subsection{Hierarchical Matching}\label{subsec:hierarchicalmatching}
    
    Putting everything together, we can define a hierarchical correspondence algorithm with the following energy:
    \begin{multline}
        \label{eq:energytotal}
        E(\mathbf{P},\mathbf{C},\tau):=\|\mathbf{P}\mathbf{X}^*_K-\mathbf{Y}_K\|^2_F+\\\lambda_\mathrm{feat}E_\mathrm{feat}(\mathbf{C})+\lambda_\mathrm{arap}E_\mathrm{arap}(\tau).
    \end{multline}
    
    The regularization terms $E_\mathrm{feat}$ and $E_\mathrm{arap}$ are defined in Eq.~\eqref{eq:featureenergy} and Eq.~\eqref{eq:arapenergy} respectively. For the former we use the SHOT \cite{tombari2010SHOT} and HKS \cite{sun2009concise} descriptors. 
	To minimize the energy $E$ we choose an alternating optimization strategy. In particular, we first fix the correspondences $\mathbf{P}$ and optimize for the alignment $(\mathbf{C},\tau)$ and then do the same vice versa. This is a common approach for both intrinsic \cite{ovsjanikov2012functional} and extrinsic \cite{ma2014robust,ma2016non,myronenko2010pointsetregistration} matching methods. Our overall matching algorithm is the following:
	\begin{myalg}\label{alg:matching}(Hierarchical Matching)
		\begin{enumerate}
		\itemsep-0.3em
		    \item[1.] Input: $\mathcal{X},\mathcal{Y}$
			\item[2.] For $K\in\mathcal{K}$:
			\begin{enumerate}
				\item[2.1] $\mathbf{P}:=\underset{\begin{smallmatrix}\mathbf{P}_{mn}\in\{0,1\},\mathbf{P}^\top\mathbf{1}=\mathbf{1}\end{smallmatrix}}{\arg\min}~E(\mathbf{P},\mathbf{C},\tau)$.
				\item[2.2] $(\tau,\mathbf{C}):=\underset{\mathbf{C}^\top\mathbf{C}=\mathbf{I}}{\arg\min}~\quad \ \ E(\mathbf{P},\mathbf{C},\tau)$.
			\end{enumerate}
			%\item[3.] Apply the final deformation to the original shape $X^*:=X+\Phi_K\tau$.
			\item[3.] Output: $\mathbf{P},\mathbf{X}^*$.
		\end{enumerate}
	\end{myalg}
	The decomposition of the optimization problem $E$ results in more tractable subproblems. For $\mathbf{P}$ this is a nearest neighbor search, for $\mathbf{C}$ a Procrustes problem and for $\tau$ a nonlinear least squares problem. The first two can be solved in closed form, for the last one we use Gauss-Newton optimization. Our method now repeatedly solves those optimization problems with shells of an increasing level of detail $K\in\mathcal{K}$ on a logarithmic scale between $K_\mathrm{init}=6$ and $K_\mathrm{max}=500$. See Figure~\ref{fig:diagramoverview} for a visualization of Algorithm~\ref{alg:matching}.

	\section{Initialization: Surrogate based Markov chain Monte Carlo Sampling}\label{sec:initial}

    Self-similarities are still a challenging problem for state-of-the-art shape correspondence methods and many struggle to distinguish them without proper initialization \cite{kernel17,ezuz2019elastic,melzi2019zoomout}. Even for humans it is difficult to distinguish between the legs/arms of an animal without any context.
    In other words, our energy from Eq.~\eqref{eq:energytotal} is highly non-convex with a multitude of local minima. 
    Unfortunately, there is no obvious way to compute a meaningful initial alignment for all classes of shapes.
    We propose an indirect approach to this using Markov chain Monte Carlo (MCMC) sampling.
    
    \paragraph{Surrogate runs}
    This approach is based on efficiently exploring the space of initial poses instead of heuristically picking one. We assign a probability distribution to the displacement parameter $\tau\in\mathbb{R}^{K_\mathrm{init}\times 3}$ and sample from this distribution. In particular, we set the prior for $\tau$ to the standard normal distribution $\mathcal{N}(0,\mathbf{I})$ and the negative log likelihood proportional to the objective value $E$. By design, this yields samples $\tau$ that have a high objective value $E$.
	Each $\tau$ is ranked according to the objective function $E$ from Eq.~\eqref{eq:energytotal} and the lowest energy result is used to initialize the full pipeline.
	
	To evaluate $E$, we run a low cost version of the full pipeline, a \emph{surrogate run}, with $K_\mathrm{max}=20$, no regularizers $\lambda_\mathrm{feat},\lambda_\mathrm{arap}:=0$ and downsampled versions of the input shapes to $1000$ vertices.
	We evaluate $N_\mathrm{prop}=100$ different proposals $\tau_\mathrm{prop}$. Those can be run in parallel with an average runtime of $0.46$ seconds per surrogate. See Figure~\ref{fig:MCMC_overview} for a visualization of this strategy and see Appendix~\ref{appendix:pseudocode} for pseudo code of our MCMC algorithm as well as the implementation in the supplementary material.

    \section{Experiments} \label{sec:experiments}
    
    We apply our pipeline to various, challenging matching tasks using two metrics to measure the quality of a matching. The first one is the \emph{accuracy}, defined as the geodesic distance to ground truth matches, see Section~\ref{subsec:shapecorrespondence}. 
    The second is the \emph{smoothness} of the correspondence $\mathbf{P}$ which we quantify using the conformal distortion of triangles, see Section~\ref{subsec:informationtransfer}. 
    To show that our method can be used out of the box, we use the same set of parameters for all experiments and do not require additional information except for the inputs $\mathcal{X}$ and $\mathcal{Y}$. 
    See our implementation in the supplementary material for more details. Additionally, we perform an ablation study in Section~\ref{sec:ablation} and a runtime analysis in the Appendix~\ref{appendix:ablation} to further investigate our method.
    Finally, there are more qualitative examples of matchings and style transfer in the Appendix.
    
    %introduction to experiments, refer to ablation study, refer to runtime analysis (and mention that our method has a favorable scaling behaviour)

    \subsection{Shape correspondence}\label{subsec:shapecorrespondence}
    
    \begin{figure}
        \centering
        \begin{overpic}
            [width=.99\linewidth]{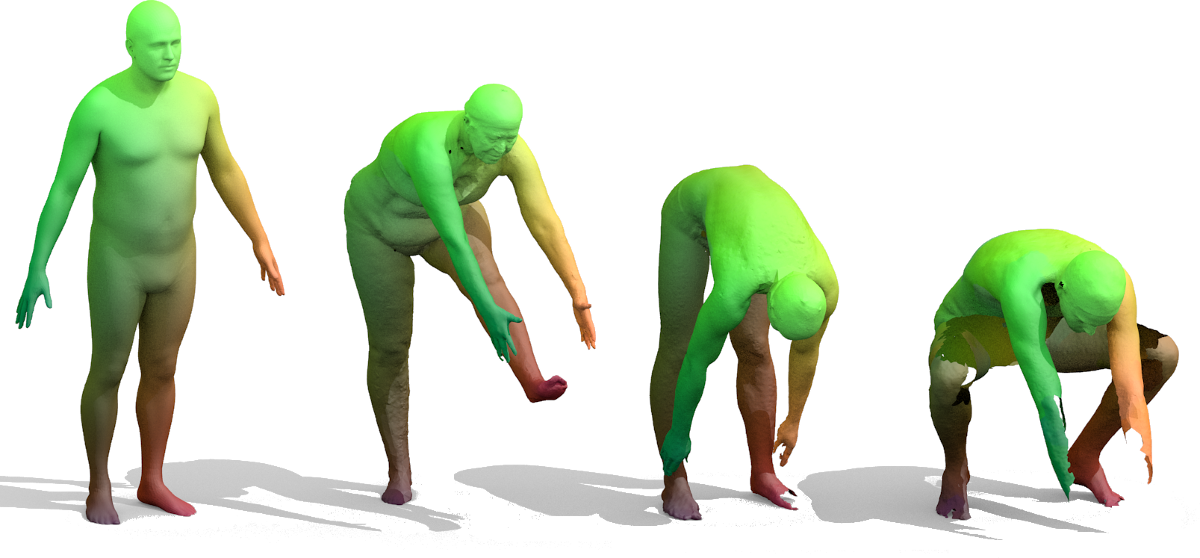}
            \put(35,45){$(a)$}
            \put(60,45){$(b)$}
            \put(85,45){$(c)$}
        \end{overpic}
        
        \caption{Example matchings for real scans from the FAUST \cite{Bogo:CVPR:2014} dataset. The shapes have very high resolution (200k vertices) and contain scanning noise. The FAUST interclass challenge consists of $(a)$ different humans that are $(b)$ subject to topological changes and $(c)$ extreme degrees of noise and partiality. Like \cite{groueix20183dcoded} we match a template (left) to each target. Here, correspondences are color coded such that matching points have the same color.}
        \label{fig:faust_qual}
    \end{figure}

    \begin{figure}
        \centering
        \includegraphics[width=.99\linewidth]{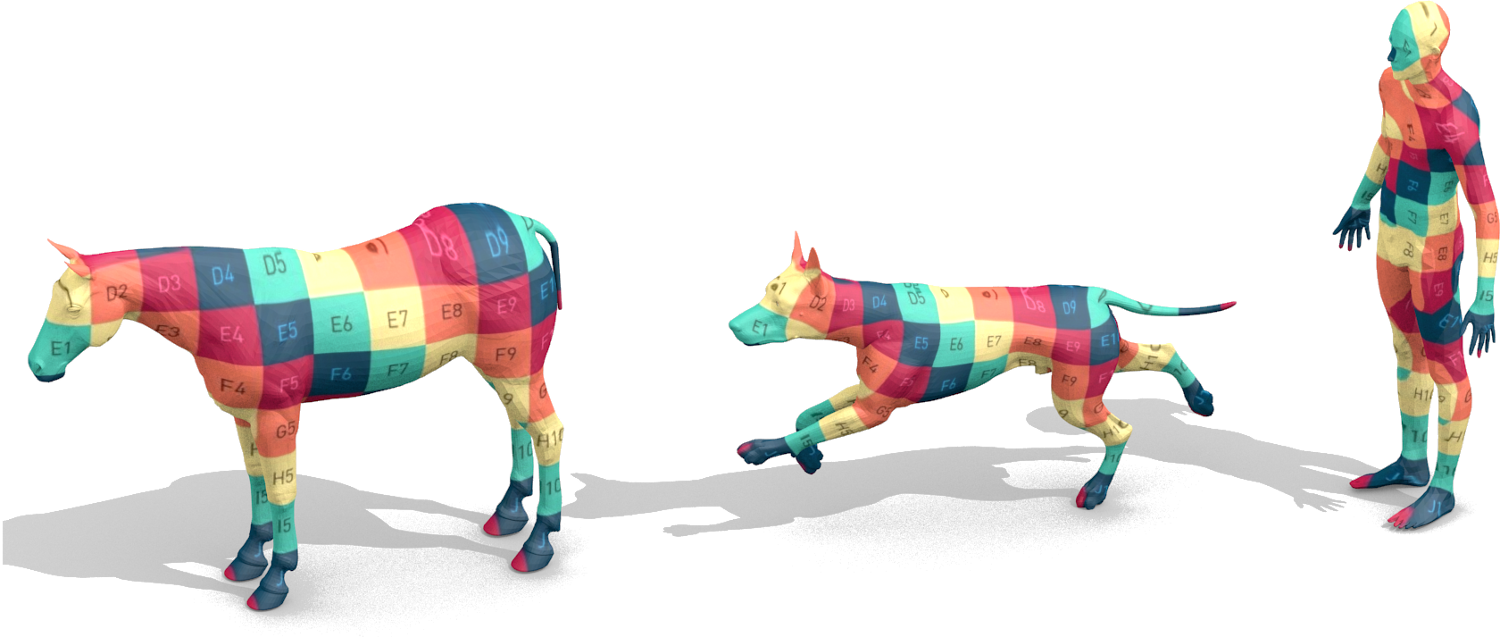}
        \caption{Example texture transfers with our method for challenging interclass examples. The texture defined on the source shape (horse) is transferred to two individual target shapes (dog and human).}
        \label{fig:interclass_texture}
    \end{figure}
    
    \begin{figure*}
        \centering
        \begin{overpic}
            [width=.7\linewidth]{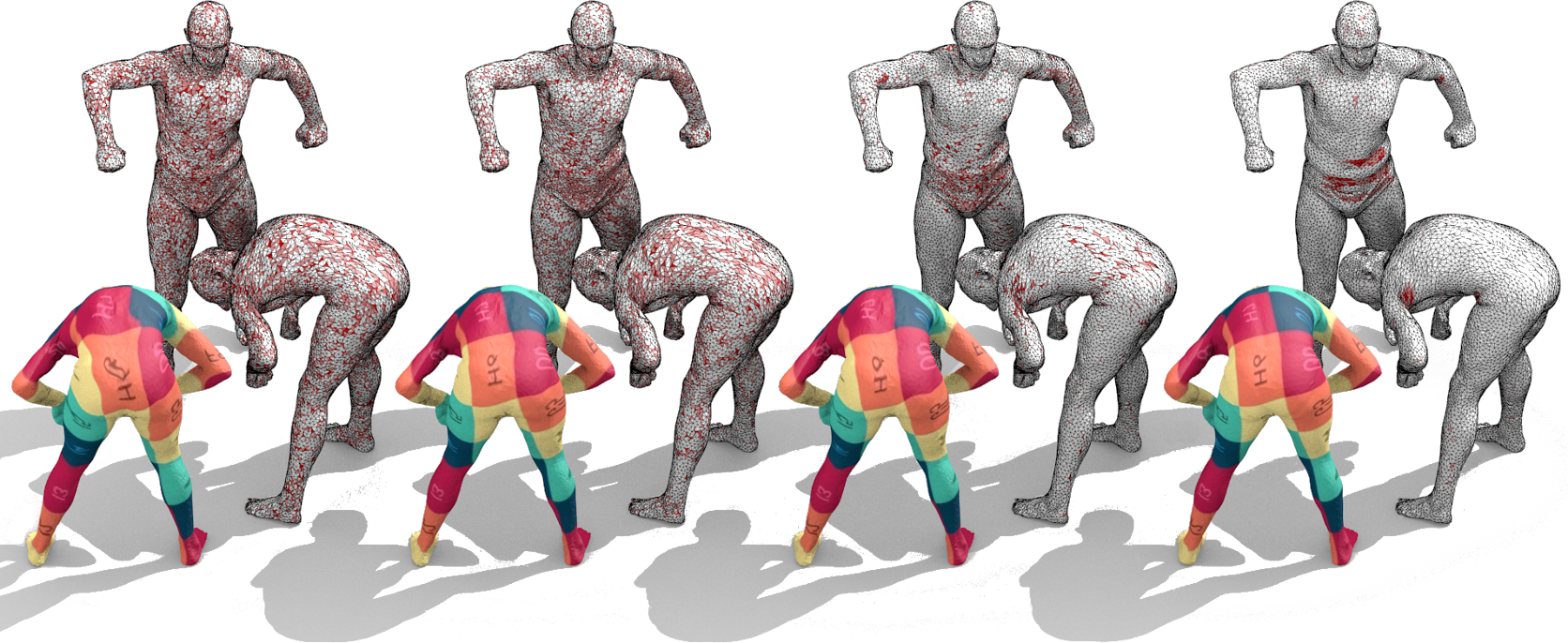}
            \put(5,45){Kernel matching}
            \put(35,45){Ours}
            \put(55,45){Ground truth}
            \put(84,45){Ours$^*$}
        \end{overpic}
        \input{figures/conf_dist_SCAPE.tikz}
        
        \caption{Correspondence smoothness measured on the SCAPE dataset using conformal distortion of triangles. (Left) Qualitative comparison -- red triangles are distorted whereas white triangles preserve the angles. Our deformed mesh shows the most meaningful, artifact-free result. Additionally, we provide an example texture transfer to prove that our deformed mesh (Ours$^*$) is the most useful one. (Right) Accumulated distortions for all $71$ pairs in the dataset. There are two possibilities to transfer the mesh of $\mathcal{X}$ to the reference pose $\mathcal{Y}$. Either we use the deformed geometry $\mathcal{X}^*$ directly (Ours$^*$) or we snap it to the surface $\mathcal{Y}$ post alignment using the map $\mathbf{P}:\mathcal{X}\to\mathcal{Y}$ (Ours). The former is only possible for our method because it is the only here one to calculate a deformation instead of only a correspondence. See Section~\ref{subsec:informationtransfer} for more details. } 
        \label{fig:scape_meshing}
    \end{figure*}
    
%     \begin{figure}
% 		\scalebox{0.75}{
% 			\begin{tabular}{|l||l|l|l|l|l|l|}
% 				\cline{1-7}
% 				Method & 3D-CODED & SP & Ours & LBS-AE & FARM & FMNet  \\ \cline{1-7}
% 				Avg. error & 2.878 (4.883$^*$) & 3.126 & 3.929 & 4.079 & 4.123 & 4.826  \\
% 				\cline{1-7}
% 			\end{tabular}
% 		}
% 		\caption{The accuracies (in cm) of the best methods for the FAUST \cite{Bogo:CVPR:2014} interclass challenge as reported on the website \emph{faust.is.tue.mpg.de}. For 3D-CODED two different versions exist, the accuracy in brackets (*) is the unsupervised version of the method. Remarkably, our method is on par with the state-of-the-art, although it is the only method listed here that did not either train on human shapes (3D-CODED, FMNet) or makes strong modelling assumptions that only hold for humans (The Stiched Puppet, LBS-AE, FARM). Our method is an out of the box tool and not specialized on human shapes.}
% 		\label{fig:faust_results}
%  	\end{figure}
    
	We evaluate the matching accuracy of our method according to the Princeton benchmark protocol \cite{kim11} on multiple datasets. Given the ground-truth match $(x, y^*)$, 
	the error of the calculated match $(x, y)$ is given by the geodesic distance between $y$ and $y^*$ normalized by the diameter of $\mathcal{Y}$:
	\begin{equation}
	    \label{eq:geoerror}
	    \epsilon(x) = \frac{\text{d}_{geo}(y, y^*)}{\sqrt{\text{area}(\mathcal{Y})}}
	\end{equation}

    \paragraph{TOSCA, SCAPE, TOPKIDS, SHRECCONN} Two datasets contain synthetic shapes with isometric pairs, TOSCA \cite{bronstein2008numerical} contains 76 shapes of humans and animals, SCAPE \cite{anguelov2005scape} contains 72 poses of the same person. TOPKIDS \cite{laehner2016shrec} contains 25 poses of a human child with self intersections. These shapes are also synthetic but topological changes from real scans are simulated with merged meshing at self-touching areas. The SHREC'19 Connectivity \cite{melzi2019shrec} dataset contains 430 pairs of human shapes from different classes with severe differences in the meshing ranging from template sized shapes ($N\approx 5000$ vertices) to real scans ($N>200$K), and varying vertex densities in different areas. We compare our matching accuracy on these datasets to other fully-automatic matching methods, see Figure~\ref{fig:matching_curves}.

    \paragraph{FAUST} The FAUST \cite{Bogo:CVPR:2014} dataset contains 300 real scans of different humans in various poses, see Figure~\ref{fig:faust_qual}. Besides being high resolution $N\approx200$K with non-compatible meshing, the shapes are noisy and highly non-isometric. Additionally, there are various poses with topological changes due to self touching parts. To address this issue, we do not compute the correspondence directly for a given pair of shapes but, like 3D-CODED \cite{groueix20183dcoded}, FARM \cite{marin2018farm} and LBS-AE \cite{li2019lbs}, use an intermediate template from \cite{SMPL:2015} to compute correspondences for two scans. This allows our method to separate topological changes and deal with noisy geometry, otherwise the as-rigid-as-possible assumption leads to faulty deformations. The accuracy (in cm) of the best methods for the FAUST \cite{Bogo:CVPR:2014} interclass challenge reported on the website \emph{faust.is.tue.mpg.de} are:\\\\
    \scalebox{0.75}{
			\begin{tabular}{|l||l|l|l|l|l|l|}
				\cline{1-7}
				Method & 3D-CODED & SP & Ours & LBS & FARM & FMNet  \\ \cline{1-7}
				Error & 2.878 (4.883$^*$) & 3.126 & 3.929 & 4.079 & 4.123 & 4.826  \\
				\cline{1-7}
			\end{tabular}
		}\\
		
    For 3D-CODED (*) refers to the unsupervised version. The striking observation is that our method is on par with the state-of-the-art without specializing on the class of human shapes. Ours is the only method listed here that does not train on human shapes (3D-CODED, FMNet) or makes strong modelling assumptions holding only for humans (Stitched Puppet (SP), LBS-AE, FARM). We did not specifically tune parameters for this challenge.

    %matching curves, plots where our method has no symmetry swaps, counting symmetry swaps for (TOSCA&)SCAPE and showing who has more/less paragraphs: isometries, different meshing, topological noise, 
    
    \begin{table*}
		\scalebox{0.87}{
			\begin{tabular}{|l||l|l|l|l|l|l|l|l|l|}
				\cline{1-10}
				\textbf{SCAPE} & Ours & $\lambda_\mathrm{feat}=0$ & $\lambda_\mathrm{arap}=0$ & Extr. only & Intr. only & w/o normals & w/o MCMC & random rigid & spectral rec.  \\ 
				\cline{1-10}
				Avg. error & 0.0088 & 0.0211 & 0.0147 & 0.0344 & 0.0121 & 0.0115 & 0.0568 & 0.1163 & 0.0139 \\
				\cline{1-10}
				\% Failure & 0 & 0.2676 & 0.0282 & 0.7606 & 0.2254 & 0.0141 & 0.8310 & 0.4930 & 0.0282 \\
				\cline{1-10}
				Avg. Distortion & 0.1287 & 0.1171 & 0.1604 & 0.1322 & 0.1539 & 0.1310 & 0.2594 & 0.2055 & 0.1305 \\
				\hhline{==========}
				\textbf{TOSCA} & Ours & $\lambda_\mathrm{feat}=0$ & $\lambda_\mathrm{arap}=0$ & Extr. only & Intr. only & w/o normals & w/o MCMC & random rigid & spectral rec.  \\ 
				\cline{1-10}
				Avg. error & 0.0056 & 0.0075 & 0.0066 & 0.0441 & 0.0205 & 0.0076 & 0.1039 & 0.0694 & 0.0098 \\
				\cline{1-10}
				\% Failure & 0 & 0.2083 & 0.0833 & 0.7500 & 0.4028 & 0.0694 & 0.8611 & 0.3889 & 0.1111 \\
				\cline{1-10}
				Avg. Distortion & 0.1654 & 0.1641 & 0.1926 & 0.1710 & 0.2239 & 0.1716 & 0.3485 & 0.2829 & 0.1666 \\
				\cline{1-10}
			\end{tabular}
		}
		\caption{Ablation study on TOSCA and SCAPE. We turn off certain parts of the method or replace it with an alternative to assess its necessity and compare the average geodesic error in $\%$ of the diameter, the failure rate and the average conformal distortion for each setting. The failure rate is the number of pairs in $\%$ where the geodesic error is twice as high than (Ours). $\lambda_\mathrm{feat},\lambda_\mathrm{arap}=0$ turns off the regularizers, \emph{Extr./Intr. only, w/o normals} removes one part of the embedding (see Eq.~\eqref{eq:productspace}), \emph{w/o MCMC} removes the initialization, \emph{random rigid} replaces our rigid alignment strategy with random rigid poses, and \emph{spectral rec.} replaces smooth shells with spectral reconstruction (see Eq.~\eqref{eq:projection}). The main insight is that the accuracy decreases whenever one of the components is removed. The conformal distortion is rather stable, except when a big percentage of the results are totally broken (e.g. w/o MCMC) or the as-rigid-as-possible regularizer is removed.} 
		\label{fig:ablation}
 	\end{table*}
    
    \subsection{Map Smoothness}\label{subsec:informationtransfer}
	
	A map with good accuracy can still produce artifacts when transferring information form surface $\mathcal{X}$ to $\mathcal{Y}$ because small scale noise typically does not have a severe effect on the geodesic matching error \eqref{eq:geoerror}. 
	This behavior is, however, prohibitive for applications like meshing, texture or normal map transfer, see Figure~\ref{fig:interclass_texture}.
	The conformal distortion of each triangle after deformation measures the local consistency of a matching \cite[Eq. (3)]{hormann2000mips}. This allows for a quantification of the smoothness of the map $\mathbf{P}$.
	
	Figure~\ref{fig:scape_meshing} shows the conformal triangle distortion of our method on the SCAPE dataset. Remarkably, the deformations obtained with our method are even smoother than the ground-truth provided in \cite{anguelov2005scape}. The reason for that lies in the way the authors construct this ground-truth. In order to transfer the meshes they use a classical nonrigid registration algorithm \cite{haehnel2003extension} to register a template in a canonical pose to 71 noisy scans of a person. This method requires $\sim$ 150 markers to get a faithful alignment, some of which are hand-selected. The main concern was to obtain a possibly tight alignment of the markers. However, in practice the markers are not perfectly placed and these small deviations lead to distorted triangles. In comparison to that, we align the templates without any markers while explicitly using an ARAP penalization term. This evidently leads to smoother deformations and the few triangles that get distorted are typically not artifacts of random noise but rather in meaningful places like the armpits or the abdomen of the person in Figure~\ref{fig:scape_meshing}.
	
	%mesh transfers, texture transfers, conformal distortion

    \subsection{Ablation study}\label{sec:ablation}
    
    We assess the effect of the different components of our method in the ablation study in Table~\ref{fig:ablation}. The main insight is that there is an intricate interplay of the different subparts of our method and the accuracy drops significantly if any part is removed. In particular, the MCMC initialization strategy is vital. Without it our deformation based approach is extremely prone to run into suboptimal local minima which leads to a failure rate of over $83\%$. Remarkably, even when our rigid initialization strategy (see end of Appendix~\ref{appendix:pseudocode}) is replaced with random rigid poses the failure rate is only around $38-50\%$. In many cases, our MCMC algorithm is able to find the correct pose, even in the presence of large scale rotational displacements of the inputs. 

	\section{Conclusion} \label{sec:conclusion}

    We have presented a novel approach to shape correspondence that combines geometric and spectral alignment by embedding the input shapes into an extrinsic-intrinsic product space.
    Our method introduces smooth shells as a coarse-to-fine shape approximation with minimal geometry changes between iterations. This is valuable for hierarchical approaches.
    Furthermore, we solve the problem of self-similarities by starting with an efficient surrogate based Markov chain Monte Carlo approach in which the deformation energy is used to find the optimal initialization.
    Finally, our method produces state-of-the-art results on established isometry datasets as well as two datasets which focus on specific noise, namely different meshing and topology changes. In the FAUST real-scan interclass challenge we are on par with the state-of-the-art although we do not train specifically for this set-up. All results were achieved with the same set of parameters which shows great generality.

    %In future work we would like to extend our framework to using more general morphing models like affine mappings.
    %This might allow our method to work with less transformation basis functions.
    %Furthermore, we would like to incorporate partial functional correspondences \cite{rodola2016partial} to apply our framework to partial data.

{\small
\bibliographystyle{ieee_fullname}
\bibliography{refs}
}

\newpage
\appendix

\section{MCMC - pseudo code}\label{appendix:pseudocode}

    \begin{figure*}
        \centering
        \input{figures/michael_runtimes.tikz}
        \input{figures/michael_errors.tikz}
        \includegraphics[width=.3\linewidth]{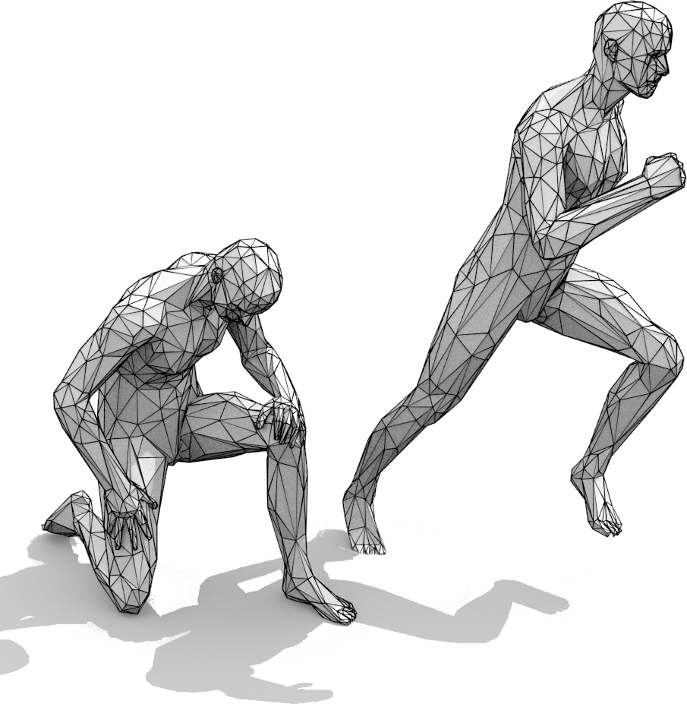}
        
        \caption{We compare the runtimes of our method with BCICP \cite{ren2018orientation}, Kernel matching \cite{kernel17} and Zoomout \cite{melzi2019zoomout}. To this end, we remesh the Michael shape from TOSCA to different resolutions, on the right side we display the pair for $N=1000$. Besides the runtime we also compare the matching accuracies of all methods. Here, our method is the most accurate one and stable across resolutions, whereas our runtime is the second best after Zoomout.}
        \label{fig:runtime}
    \end{figure*}

    In Section~\ref{sec:initial} we already gave a rough description of our MCMC initialization algorithm. Here, we provide a more detailed pseudo code:
	\begin{myalg}\label{alg:MCMC}
		(MCMC)
		\begin{enumerate}
		\itemsep0em
			\item[1.] Initialize $\tau_\mathrm{best}:=0,\mathbf{X}^*_\mathrm{best}:=\mathbf{X}$.
			\item[2.] For $i=1,...,N_\mathrm{prop}$:
			\begin{enumerate}
				\item[2.1] Sample new proposal $\tau_\mathrm{prop}\sim\mathcal{N}(0,\mathbf{I})$.
				\item[2.2] Compute the current alignment $\mathbf{X}^*_\mathrm{prop}$ by making a surrogate run with the initial guess $\tau_\mathrm{prop}$.
				\item[2.3] Compute the acceptance probability $\alpha:=\\\exp\biggl(-\frac{1}{2\sigma_\mathrm{match}^2}\bigl(E(\mathbf{X}^*_\mathrm{prop})-E(\mathbf{X}^*_\mathrm{best})\bigr)\biggr)$ using the energy $E$ from  Eq.~\eqref{eq:energytotal}.
				\item[2.4] Sample $u\sim\mathrm{Unif}(0,1)$ and either accept or reject the new sample $\tau_\mathrm{prop}\in\mathbb{R}^{K_\mathrm{init}\times 3}$: \begin{align*}&(\tau_\mathrm{best},\mathbf{X}^*_\mathrm{best}):= \\ &\quad\quad\quad\quad\quad\begin{cases} (\tau_\mathrm{prop},\mathbf{X}^*_\mathrm{prop}), & u\leq\alpha \text{ (accept)} \\ (\tau_\mathrm{best},\mathbf{X}^*_\mathrm{best}), & u>\alpha\text{ (reject)} \end{cases}\end{align*}
			\end{enumerate}
		\end{enumerate}
	\end{myalg}

    We usually set the number of surrogates to $N_\mathrm{prop}:=100$. In the majority of cases in our experiments this is more than sufficient. Furthermore, we usually choose a small objective variance $\sigma_\mathrm{match}^2:=0.001$ to get a sharp distribution and therefore more accurate samples $\tau$.

    \paragraph{Remarks}
    
    One aspect of our method that we did not talk about yet is how to compute a good initial rigid pose. For most datasets in our experiments this is a requirement, e.g. SHREC'19 \cite{melzi2019shrec} connectivity has random rigid poses for all inputs. In theory, our MCMC algorithm can account for rigidly displaced inputs $\mX$ and $\mY$ but in practice our $N_\mathrm{prop}=100$ surrogates are not enough for extreme cases. Therefore, we initially apply another surrogate based method that initializes with different rigid poses and determines the best one according to the objective $E$ from Eq.~\eqref{eq:energytotal}. A thorough description of this is beyond the scope of this paper, but all the details can be found in our implementation. 
    
\section{Proof of Theorem~\ref{thm:smoothness}}\label{appendix:proof1}

    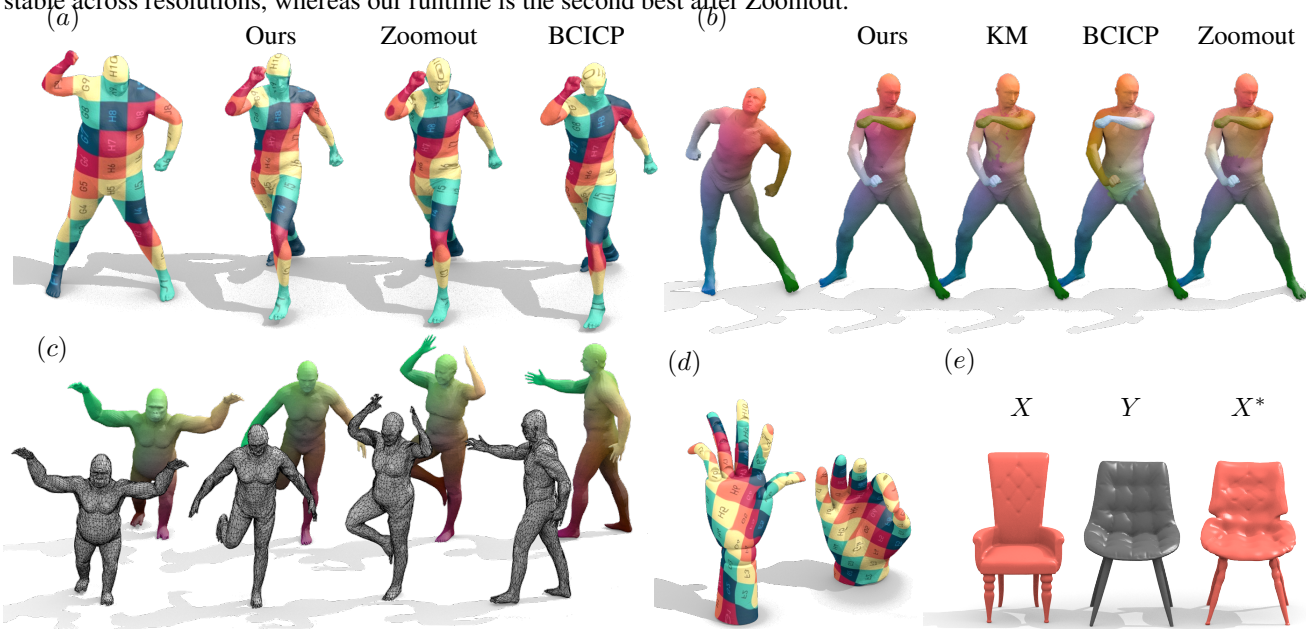
\begin{figure*}
        \centering
        \begin{overpic}
            [width=.49\linewidth]{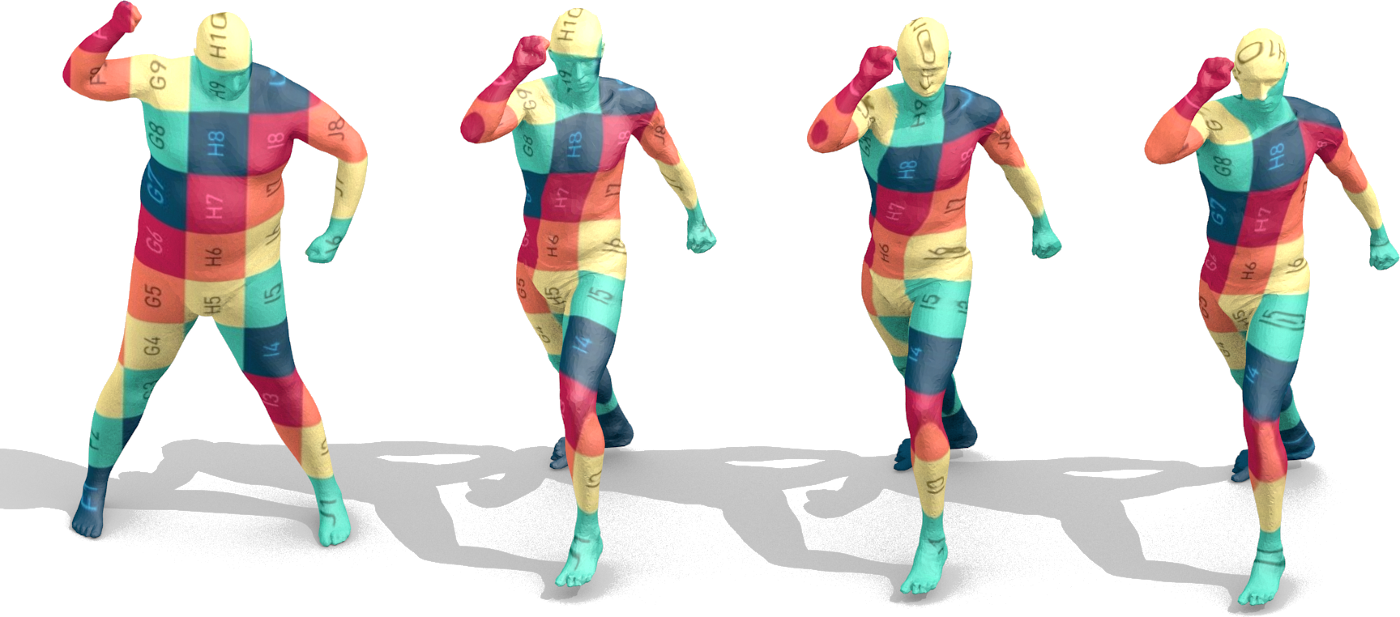}
            \put(5,48){$(a)$}
            \put(36,45){Ours}
            \put(57,45){Zoomout}
            \put(83,45){BCICP}
        \end{overpic}
        \begin{overpic}
            [width=.49\linewidth]{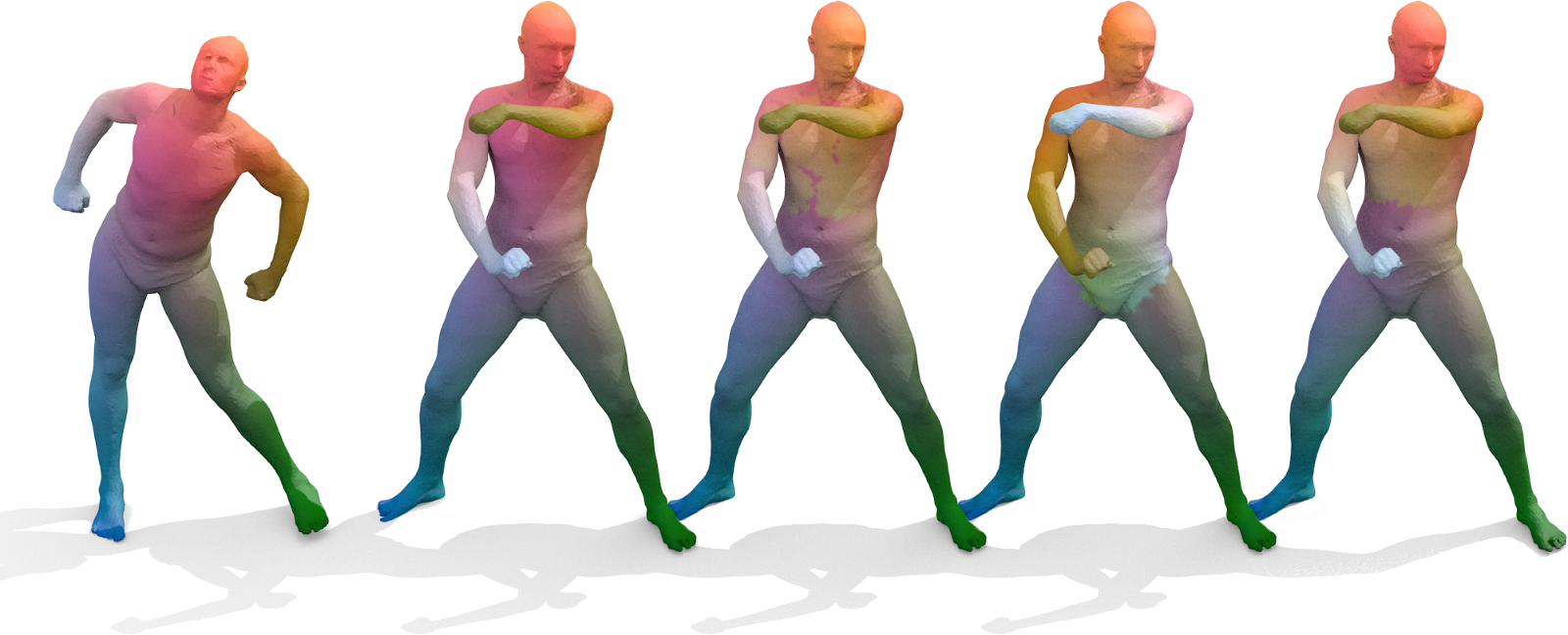}
            \put(5,48){$(b)$}
            \put(30,45){Ours}
            \put(50,45){KM}
            \put(65,45){BCICP}
            \put(83,45){Zoomout}
        \end{overpic}
        \begin{overpic}
            [width=.49\linewidth]{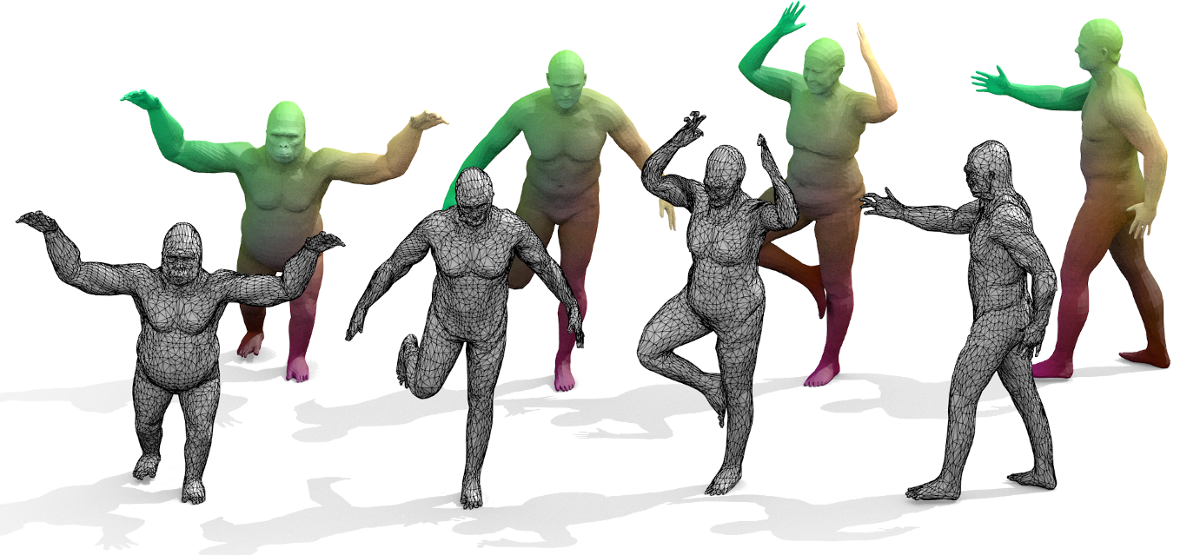}
            \put(5,43){$(c)$}
        \end{overpic}
        \begin{overpic}
            [width=.2\linewidth]{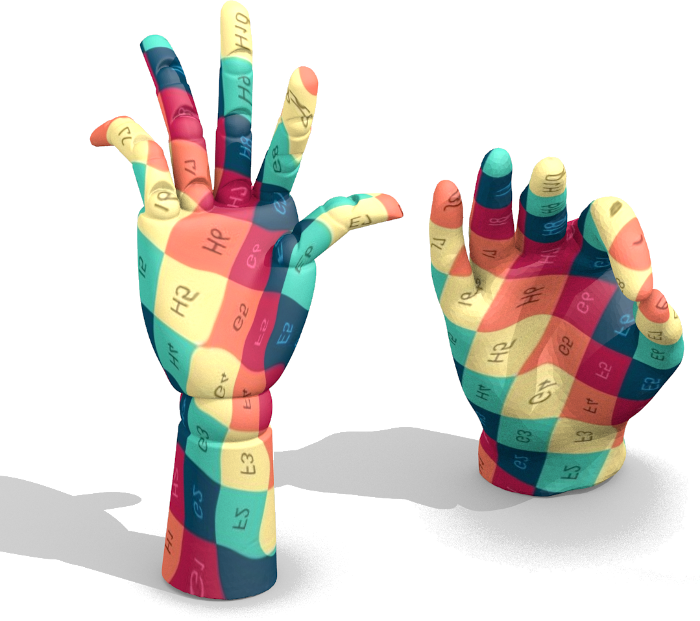}
            \put(5,100){$(d)$}
        \end{overpic}
        \begin{overpic}
            [width=.3\linewidth]{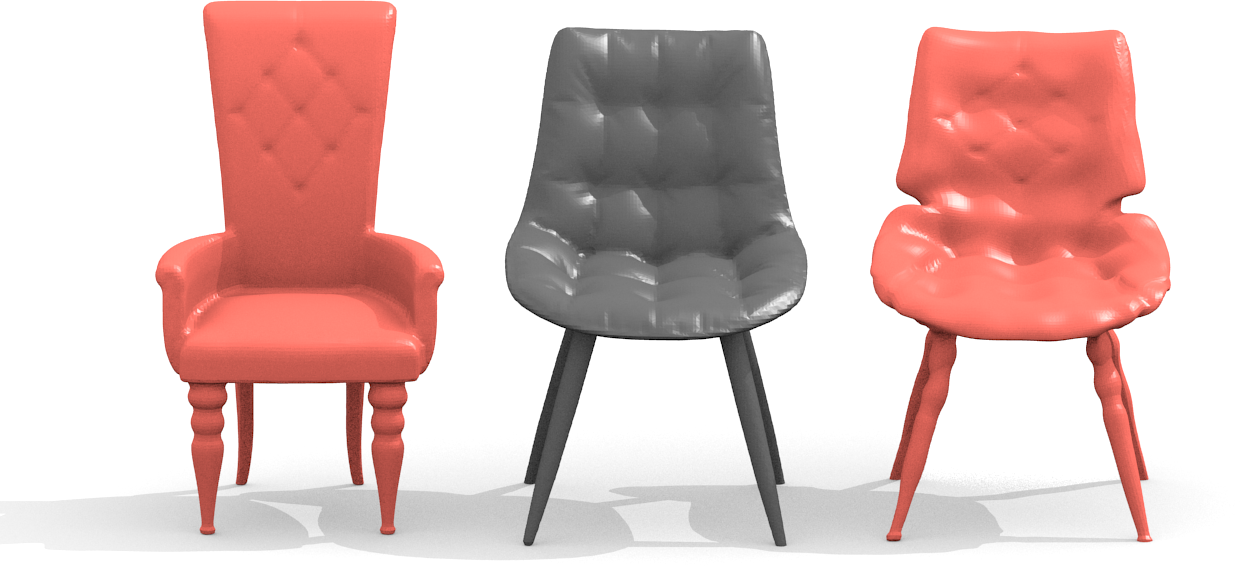}
            \put(5,67){$(e)$}
            \put(22,55){$X$}
            \put(50,55){$Y$}
            \put(78,55){$X^*$}
        \end{overpic}
        \caption{Here, we show additional qualitative evaluations of our method. $(a)$ and $(b)$ are comparisons for a pair from the SHREC'19 and the SCAPE dataset respectively. $(a)$ is challenging due to an incompatible meshing, $(b)$ has equivalent meshing but is still susceptible to mismatches due to self-similarities (left-right: BCICP, front-back: KM, Zoomout). $(c)$ shows how our method can be used to smoothly transfer meshings for interclass pairs, here for a gorilla from TOSCA to humans from FAUST. The maps are smooth in the sense that local structures are preserved and deformations only occur in the form of uniform, global stretching of parts. I.e. the face still looks like a gorilla after deformation although the rest of the body adapts to the human form. $(d)$ shows a texture transfer from a template hand (right) to a scanned hand of a puppet (left). The latter is a scan of a real world object from \cite{dyke2019shape}, obtained with the handheld Space Spider scanner from Artec. This is a challenging example due to different resultions of the inputs, different small scale features and a different size of the residual part at the bottom. $(e)$ shows how our method can be applied to deform an object (red chair) and align it with a reference shape (black chair) to create a new object. The deformed red chair $X^*$ has the global structure of $Y$ and the fine scale details of $X$.}
        \label{fig:additional_qualitative}
    \end{figure*}

    Theorem~\ref{thm:smoothness} gives an upper bound on how much the geometry of our smooth shells can change between two states $K$ and $K+1$. For spectral reconstruction, a projection on a new eigenfunction is added in each iteration. Depending on the magnitude of the new projection $\bigl(\phi_{K+1}\otimes\phi_{K+1}\bigr)$, this can lead to arbitrarily high changes :
    \begin{equation}
    \label{eq:magnitudeupsamplingspectralrec}
         \|\mathcal{T}_{K+1}(X)-\mathcal{T}_K(X)\|_{L^2}=\|\bigl(\phi_{K+1}\otimes\phi_{K+1}\bigr)X\|_{L^2}.
    \end{equation}
   
   In comparison, Theorem~\ref{thm:smoothness} states that the change from $\mathcal{S}_K(X)$ to $\mathcal{S}_{K+1}(X)$ can be bounded by choosing a small upsampling variance $\sigma$.
   
   \begin{proof}
        We will proof the statement for scalar functions $X\in L^2(\mathcal{X})$. The extension to vector valued functions $L^2(\mathcal{X},\mathbb{R}^3)$ is trivial -- we just need to apply the identity to each component at a time. Now let $K>0$ and $\sigma>0$. For brevity we will denote the sigmoid weights with $s_k^K:=\frac{1}{1+\exp\bigl(\sigma(k-K)\bigr)}$. Using the spectral decomposition of operators, we can deduce:
        \begin{flalign*}
            &\bigl\|\mathcal{S}_{K+1}(X)-\mathcal{S}_{K}(X)\bigr\|_{L^2}^2=\\
            &\sum_{k=1}^{\infty}\bigl|\bigl(s_k^{K+1}-s_k^K\bigr)\bigl\langle\phi_k,X\bigr\rangle_{L^2}\bigr|^2=\\
            &\sum_{k=1}^{\infty}\bigl|\biggl(\bigl(s_k^K\bigr)^{-1}-\bigl(s_k^{K+1}\bigr)^{-1}\biggr)s_k^{K+1}s_k^K\bigl\langle\phi_k,X\bigr\rangle_{L^2}\bigr|^2=\\
            &\sum_{k=1}^{\infty}\bigl|(1-e^{-\sigma})\exp\bigl(\sigma(k-K)\bigr)s_k^{K+1}s_k^K\bigl\langle\phi_k,X\bigr\rangle_{L^2}\bigr|^2\leq\\
            &\sum_{k=1}^{\infty}\bigl|(1-e^{-\sigma})\biggl(1+\exp\bigl(\sigma(k-K)\bigr)\biggr)s_k^{K+1}s_k^K\bigl\langle\phi_k,X\bigr\rangle_{L^2}\bigr|^2=\\
            &\sum_{k=1}^{\infty}\bigl|(1-e^{-\sigma})s_k^{K+1}\bigl\langle\phi_k,X\bigr\rangle_{L^2}\bigr|^2=\\
            &\bigl|(1-e^{-\sigma})\bigr|^2\bigl\|\mathcal{S}_{K+1}(X)\bigr\|_{L^2}^2.
        \end{flalign*}
        Taking the square root on both sides then yields the desired identity.
   \end{proof}
   
    Remarkably, this bound is independent of the index $K$. Small eigenfunctions $\phi_k$ typically represent coarse structures like limbs. Therefore, in particular the first iterations using spectral reconstruction lead to big changes in the geometry, see Eq.~\eqref{eq:magnitudeupsamplingspectralrec}.
    
    \section{Runtime Analysis}\label{appendix:ablation}
    
    We analyze the time complexity of our method in comparison to other popular matching methods in Figure~\ref{fig:runtime}. In particular, we compare the runtime of the whole pipelines for instances of the same pair of Michael shapes from the TOSCA dataset that was remeshed to different resolutions between $500$ and $50k$ vertices.

    \section{Additional Qualitative Evaluations}\label{appendix:qualitative}
    
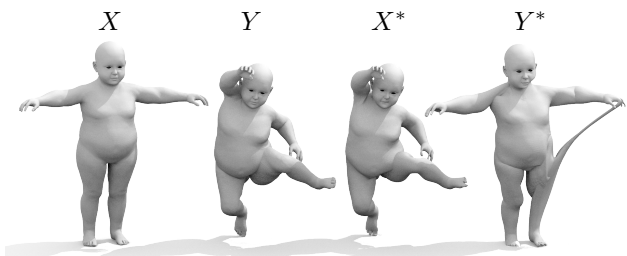
\begin{figure}
        \centering
        \begin{overpic}
            [width=.99\linewidth]{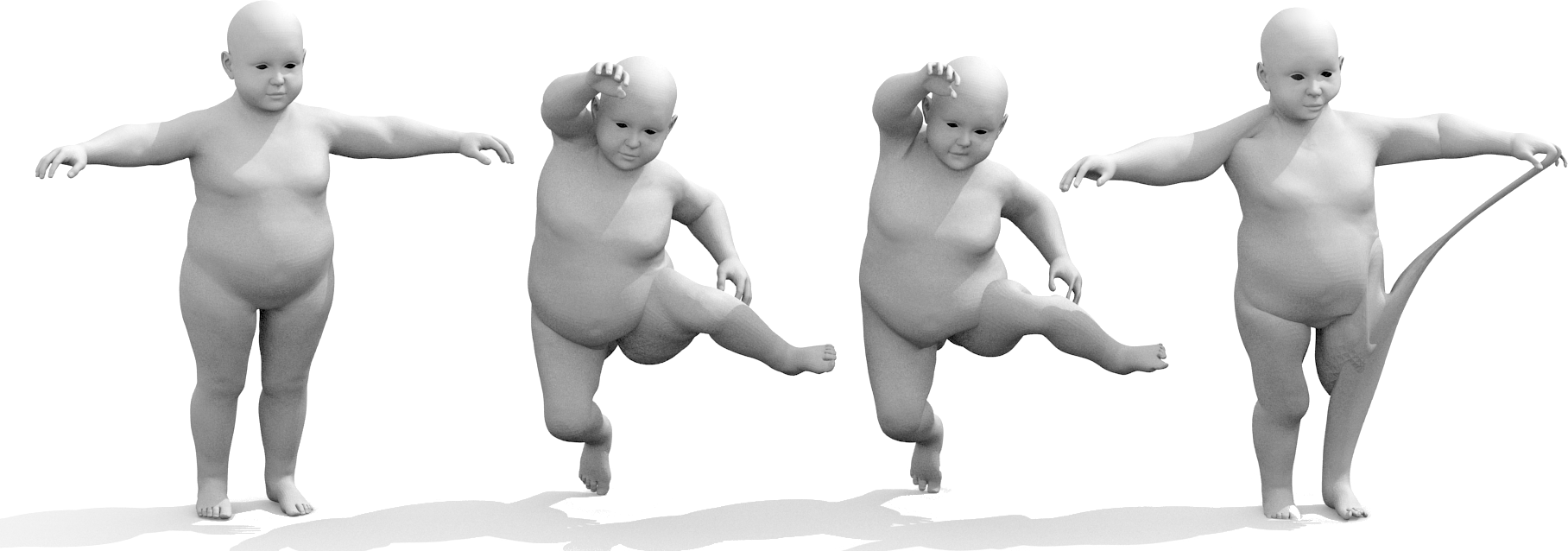}
            \put(15,37){$X$}
            \put(38,37){$Y$}
            \put(59,37){$X^*$}
            \put(82,37){$Y^*$}
        \end{overpic}
        \caption{A failure case of our method for a pair of shapes from TOPKIDS. If we align the the canonical pose $X$ with the reference $Y$ we get a meaningful alignment $X^*$ and high quality correspondences. However, if we try to apply our method the other way around, we get undesirable "cheese pull" effects at the left hand of the deformed kid $Y^*$. The reason for that is that in the pose $Y$ the fingers touch the left knee and the meshing connects. In order to avoid this effect, we either need a mesh separation policy or use an intermediate template where the original topology is known. We prefer the latter approach in our quantitative evaluations on FAUST and TOPKIDS because finding a meaningful topological cut is a complicated problem on its own.}
        \label{fig:failure_case}
    \end{figure}

    We provide some additional qualitative evaluations and comparisons of our pipeline in order to give the reader a better understanding about the merits of our method, see Figure~\ref{fig:additional_qualitative}. Additionally, we provide a failure case in Figure~\ref{fig:failure_case}. Our method is deformation based with an as-rigid-as-possible assumption. This means that in places of topological changes the meshing cannot be separated. Our method still tries to align the shape as good as possible with the reference which invariably leads to a "cheese pull" effect. This is also the main reason why we use an intermediate template to match the FAUST and TOPKIDS shapes in our quantitative evaluations.
    
\end{document}

%% file: figures/MDS_MCMC.tikz
\pgfplotsset{ticks=none}
\begin{tikzpicture}
\begin{axis}[%
width=0.2\linewidth,
height=0.27\linewidth,
at={(0,0)},
scale only axis,
colormap={mymap}{[1pt] rgb(0pt)=(0.2422,0.1504,0.6603); rgb(1pt)=(0.25039,0.164995,0.707614); rgb(2pt)=(0.257771,0.181781,0.751138); rgb(3pt)=(0.264729,0.197757,0.795214); rgb(4pt)=(0.270648,0.214676,0.836371); rgb(5pt)=(0.275114,0.234238,0.870986); rgb(6pt)=(0.2783,0.255871,0.899071); rgb(7pt)=(0.280333,0.278233,0.9221); rgb(8pt)=(0.281338,0.300595,0.941376); rgb(9pt)=(0.281014,0.322757,0.957886); rgb(10pt)=(0.279467,0.344671,0.971676); rgb(11pt)=(0.275971,0.366681,0.982905); rgb(12pt)=(0.269914,0.3892,0.9906); rgb(13pt)=(0.260243,0.412329,0.995157); rgb(14pt)=(0.244033,0.435833,0.998833); rgb(15pt)=(0.220643,0.460257,0.997286); rgb(16pt)=(0.196333,0.484719,0.989152); rgb(17pt)=(0.183405,0.507371,0.979795); rgb(18pt)=(0.178643,0.528857,0.968157); rgb(19pt)=(0.176438,0.549905,0.952019); rgb(20pt)=(0.168743,0.570262,0.935871); rgb(21pt)=(0.154,0.5902,0.9218); rgb(22pt)=(0.146029,0.609119,0.907857); rgb(23pt)=(0.138024,0.627629,0.89729); rgb(24pt)=(0.124814,0.645929,0.888343); rgb(25pt)=(0.111252,0.6635,0.876314); rgb(26pt)=(0.0952095,0.679829,0.859781); rgb(27pt)=(0.0688714,0.694771,0.839357); rgb(28pt)=(0.0296667,0.708167,0.816333); rgb(29pt)=(0.00357143,0.720267,0.7917); rgb(30pt)=(0.00665714,0.731214,0.766014); rgb(31pt)=(0.0433286,0.741095,0.73941); rgb(32pt)=(0.0963952,0.75,0.712038); rgb(33pt)=(0.140771,0.7584,0.684157); rgb(34pt)=(0.1717,0.766962,0.655443); rgb(35pt)=(0.193767,0.775767,0.6251); rgb(36pt)=(0.216086,0.7843,0.5923); rgb(37pt)=(0.246957,0.791795,0.556743); rgb(38pt)=(0.290614,0.79729,0.518829); rgb(39pt)=(0.340643,0.8008,0.478857); rgb(40pt)=(0.3909,0.802871,0.435448); rgb(41pt)=(0.445629,0.802419,0.390919); rgb(42pt)=(0.5044,0.7993,0.348); rgb(43pt)=(0.561562,0.794233,0.304481); rgb(44pt)=(0.617395,0.787619,0.261238); rgb(45pt)=(0.671986,0.779271,0.2227); rgb(46pt)=(0.7242,0.769843,0.191029); rgb(47pt)=(0.773833,0.759805,0.16461); rgb(48pt)=(0.820314,0.749814,0.153529); rgb(49pt)=(0.863433,0.7406,0.159633); rgb(50pt)=(0.903543,0.733029,0.177414); rgb(51pt)=(0.939257,0.728786,0.209957); rgb(52pt)=(0.972757,0.729771,0.239443); rgb(53pt)=(0.995648,0.743371,0.237148); rgb(54pt)=(0.996986,0.765857,0.219943); rgb(55pt)=(0.995205,0.789252,0.202762); rgb(56pt)=(0.9892,0.813567,0.188533); rgb(57pt)=(0.978629,0.838629,0.176557); rgb(58pt)=(0.967648,0.8639,0.16429); rgb(59pt)=(0.96101,0.889019,0.153676); rgb(60pt)=(0.959671,0.913457,0.142257); rgb(61pt)=(0.962795,0.937338,0.12651); rgb(62pt)=(0.969114,0.960629,0.106362); rgb(63pt)=(0.9769,0.9839,0.0805)},
xmin=-0.1,
xmax=0.25,
ymin=-0.1,
ymax=0.15,
xminorgrids,
yminorgrids,
axis background/.style={fill=white},
%axis x line*=bottom,
%axis y line*=left,
legend style={legend cell align=left,align=left,draw=white!15!black}
]
\addplot[scatter,line width=1pt,only marks,scatter src=explicit,scatter/use mapped color={mark=o,mark options={},draw=mapped color},visualization depends on={\thisrowno{2} \as \perpointmarksize},scatter/@pre marker code/.append style={/tikz/mark size=\perpointmarksize}] plot table[row sep=crcr,meta index=3]{%
-0.0554497244133511	-0.0113124400560085	1.22307405659037	23.9345623664706\\
-0.0498928734983961	0.0216818289675928	4.88159671475484	381.279783768084\\
-0.047725000104371	0.0211501872554976	3.07530417508205	151.319932308434\\
-0.0376178267711825	0.040185510792498	1.70119050775188	46.3047862986418\\
-0.0309851393326795	-0.035582455007005	3.22923198907127	166.847027827859\\
0.0139227461809431	0.0322183506284629	2.17815897557951	75.9100243663614\\
-0.0407018302480801	-0.0263793123073893	2.69700022394591	116.380963327429\\
-0.0286199559364562	-0.0378913377027148	3.50022486305078	196.025185470701\\
-0.0530771037880731	0.0218334427725904	2.91420048301173	135.881031282973\\
-0.0611812632045267	0.0159036660748075	4.27022004387792	291.756467570188\\
-0.0344100725403958	-0.0443703387805879	3.00053708234042	144.051564519999\\
-0.0549730477248865	0.0220444729476648	1.1370728784788	20.6869556955529\\
-0.049695235704374	0.0212410557071638	5.12071429424505	419.547438132569\\
-0.0203186694313825	-0.0383153750443979	1.27469876380743	25.9977110152349\\
-0.03577327321623	-0.0178310363047565	2.73603096626326	119.773847173623\\
-0.0661009932669714	-0.00843609334922902	0.851218416861056	11.5931646912551\\
-0.0501541809252095	0.0208116381658998	5.05201631827902	408.36590208252\\
-0.0232724708212576	-0.0441994397905365	2.04503108581412	66.9144342711371\\
-0.0402769212988962	0.0255117010710557	1.24411385341793	24.7651084842624\\
-0.012699248708639	-0.00181439152273573	1.52473588558334	37.1971123325696\\
0.0180135476747218	-0.0837577159457944	1.83485607502121	53.8671490566759\\
0.186745679382981	0.130463744321653	0.796052585856121	10.1391955111715\\
-0.0555570979093662	0.0206756675149754	1.07312012769165	18.4253889353111\\
-0.0502510245162159	0.0155638823283687	5.09009483705961	414.545047204174\\
-0.052799025622798	0.00290381149854301	1.85383056089286	54.9870039760055\\
0.044349590093664	-0.0671525735360444	1.51220552136355	36.5882486214785\\
-0.0327515423976042	-0.033268943700648	3.35720189476111	180.332872995002\\
0.0165967877230568	-0.0813965567608065	0.932375599815905	13.9091881461131\\
-0.0491210994251878	0.0170329889999997	5.58670944440246	499.38115865881\\
-0.0519102411262762	0.0298923918379861	1.06862679093281	18.2714114927898\\
0.0124092655636802	-0.0844690571899372	3.30623719284469	174.899270005592\\
-0.0630648740289165	-0.00679918677651569	1.7478249275931	48.8782716402534\\
-0.0470298805580216	0.0198069116192101	1.97302506466274	62.2852464925986\\
0.0278104601605324	-0.0832923368480986	1.50608326723915	36.292588925724\\
0.201283489307556	-0.0440038072426277	1.25454136755157	25.1819846863705\\
-0.0488018077867289	0.019875152221689	5.24063801428523	439.428588748344\\
0.0756817486743228	0.0739868801582361	2.01883097873058	65.2108563309165\\
-0.0207173317924067	-0.041013449230275	1.40613906078179	31.6356329321006\\
-0.0571292094453687	-0.00608841069550216	1.78386867043263	50.9149989336171\\
0.224872374768095	-0.00570999699258882	2.31053316010244	85.4170157429272\\
-0.042901850873182	-0.000685666837160534	1.38681747880522	30.7722035123147\\
0.216996259530119	0.0004381766458346	2.59322359971443	107.596938209854\\
0.0416129903482625	-0.0849767403754518	1.18101378182168	22.316696845644\\
-0.0520367129462199	0.0126783691299992	0.8758422976495	12.2735956856313\\
-0.0400699354357481	0.0192454671881776	1.70950052775711	46.7582728704292\\
-0.0489883885419864	0.0299431081401826	2.04606511376198	66.9821191960611\\
-0.0483444174342028	0.0178704212478123	1.49268514691733	35.6497431652417\\
-0.0448874883577283	0.00374157169440109	4.33351146017158	300.469145207014\\
0.0199077117252226	-0.0850234514291602	2.72519125017999	118.826677600921\\
-0.0458484246380033	0.0132927615702814	1.4763161416882	34.8721496033461\\
-0.0493323483873538	0.0213309911601246	4.89172462164509	382.86351638414\\
0.245623871310488	0.0179039193880069	1.0597758195771	17.9699966041652\\
-0.0501377143569093	0.0248975528737071	3.41608531216934	186.714221760306\\
-0.045960444721024	0.00553127803352646	4.28611566378021	293.932599732833\\
0.241613848524257	0.0125921948790127	1.59841665611966	40.8789729049722\\
-0.0543381341407826	0.0121947923910005	2.31986375224642	86.1082852637891\\
0.106886796686275	-0.0182521768712334	0.458869092140036	3.36897349954273\\
0.0132936875964954	0.02859195928776	2.49569026641601	99.6555184941377\\
-0.0457452599972578	0.00534207431646399	4.0684662704785	264.83868470434\\
-0.0431344934843709	0.0220125342106815	3.0068021748149	144.653749095545\\
-0.046329154536774	0.0163759638462384	5.33125801467316	454.756992304269\\
-0.0506684060212507	0.0247785226225571	2.91396866538304	135.859414125347\\
-0.0434007635356947	0.0216170304141836	3.06513569115349	150.320908882928\\
-0.0470355857476445	0.0164464558670549	5.59016994374947	500\\
0.049012307672016	0.0890767681286275	2.01201561200246	64.7713091670659\\
-0.0493243369806236	0.0209344664057121	5.32168878914134	453.125945095562\\
-0.048710639428369	-0.0152048476348955	2.69074945501918	115.842122074976\\
-0.0316261882118033	-0.0241215369658234	2.00077386221803	64.0495367637576\\
-0.058386994064527	0.0222164024350569	2.4724650576661	97.8093353820774\\
-0.0650650815623514	0.0139024315062261	1.76068991680179	49.6004637300397\\
0.193794531585857	0.0198329136258329	0.873546953093559	12.2093484681446\\
-0.0397812184237494	-0.0247059227438812	3.12156593016648	155.906781702018\\
-0.0502380433092463	0.0203158052788272	5.08251225007744	413.310892354996\\
0.161357945556153	-0.0422764315217647	1.31920670928754	27.8449014692683\\
0.0246481413978197	0.0579229102551232	1.62638190451342	42.3218895892593\\
-0.0497739782405756	0.0212974254383119	5.13727064416818	422.264794742914\\
0.103787182361468	-0.0724032295046046	1.68599804622481	45.4814305899819\\
0.0116757597124112	-0.0822142407844848	1.71388822155759	46.9986053759012\\
0.0102015269586971	-0.0762828150634375	1.4155857732574	32.0621293031802\\
0.00752085996128663	0.0504742784544144	2.32510786656523	86.4980254586165\\
-0.00754704358525823	0.0343563134716106	1.55951402317084	38.9133438154639\\
-0.0702091857576478	0.0117173857884689	1.04189581208606	17.3687501318794\\
-0.0492842529734347	0.0209826026543338	5.25964850762764	442.622438780635\\
0.244157150266062	0.013076500848823	1.42595213680424	32.5334319433054\\
-0.0500454668660796	0.0181135250546282	4.00202555713708	256.259336959653\\
-0.048514870649497	0.0206175618571469	5.39934104394629	466.446139341488\\
-0.0459040189495168	0.00470811306856205	4.38622685139161	307.8237758699\\
0.0205333394781351	-0.0805765440391712	2.76998354709548	122.764941618874\\
-0.0458157667975001	0.00453486252017143	4.21602108079362	284.397340059139\\
-0.0452686444970915	-0.00328354147733719	2.36941316100479	89.825899640683\\
-0.0556822676309757	-0.0270379515587659	2.91900294228481	136.329250833078\\
0.0393300109369344	-0.0817089490523325	0.993191585956277	15.7828724226295\\
0.0989858758061194	0.117630993539831	1.25606253024072	25.2430892779956\\
0.0932748691038398	0.085710186722527	1.91981659181197	58.9711319391443\\
-0.0352375591552785	0.00218702232844883	1.97655371690807	62.5082335331695\\
-0.0269604451038599	-0.0389475070256479	3.1797620812695	161.77419029567\\
0.0191302080263282	0.0551609641263535	1.19281637322204	22.7649744036251\\
0.221713003372155	0.00337928121714548	2.33068806319044	86.9137095663745\\
-0.0345746952655409	0.0328721859934162	1.0038672718755	16.1239919926842\\
0.0224246187073525	-0.0798435228411483	1.49442382964881	35.7328413219552\\
};
\end{axis}

\end{tikzpicture}%

%% file: figures/curve_tosca.tikz
\definecolor{mycolor1}{rgb}{0.00000,1.00000,1.00000}%
\definecolor{mycolor2}{rgb}{0.91000,0.41000,0.17000}%
\definecolor{mycolor3}{rgb}{1.00000,0.84314,0.00000}%
\begin{tikzpicture}

\begin{axis}[%
width=0.18\linewidth,
height=0.18\linewidth,
at={(0,0)},
scale only axis,
xmin=0,
xmax=0.1,
xlabel style={font=\color{white!15!black}},
xlabel={Relative geodesic error},
ymin=0,
ymax=1,
ylabel style={font=\color{white!15!black}},
ylabel={\% Correspondences},
every x tick label/.append style={font=\color{black}, font=\footnotesize},
every y tick label/.append style={font=\color{black}, font=\footnotesize},
x tick label style={/pgf/number format/fixed},
axis background/.style={fill=white},
title style={font=\bfseries},
title={TOSCA},
xmajorgrids,
ymajorgrids,
legend style={at={(0.34,0.02)}, anchor=south west, legend cell align=left, align=left, draw=white!15!black,font=\tiny}
]

\addplot [color=blue, line width=2.0pt]
  table[row sep=crcr]{%
0	0\\
0	0.00502784760899457\\
0	0.0100556952179891\\
0	0.0150835428269837\\
0	0.0201113904359783\\
0	0.0251392380449728\\
0	0.0301531965721746\\
0	0.0351810441811692\\
0	0.0402088917901638\\
0	0.0452367393991583\\
0	0.0502645870081529\\
0	0.0552785455353547\\
0	0.0603063931443492\\
0	0.0653342407533438\\
0	0.0703620883623384\\
0	0.0753899359713329\\
0	0.0804038944985347\\
0	0.0854317421075293\\
0	0.0904595897165238\\
0	0.0954874373255184\\
0.000257131530147251	0.100515284934513\\
0.000407296426956153	0.105529243461715\\
0.000518325794254367	0.110557091070709\\
0.000635278123878182	0.115584938679704\\
0.00072114424170548	0.120612786288698\\
0.000801670493963633	0.125640633897693\\
0.000880216403044906	0.130654592424895\\
0.000964963459886286	0.135682440033889\\
0.00104864852864642	0.140710287642884\\
0.00112933424292103	0.145738135251879\\
0.00120588924159862	0.150765982860873\\
0.00128895421013746	0.155779941388075\\
0.00136919676144879	0.160807788997069\\
0.0014540316367113	0.165835636606064\\
0.00153722490390749	0.170863484215059\\
0.00161278691987809	0.175891331824053\\
0.00169252477106228	0.180905290351255\\
0.0017662268427893	0.185933137960249\\
0.00185655481384439	0.190960985569244\\
0.00194498120111045	0.195988833178239\\
0.00203749774994929	0.201016680787233\\
0.00212017935501049	0.206030639314435\\
0.00220549735095986	0.211058486923429\\
0.00228851865241022	0.216086334532424\\
0.00236839119144015	0.221114182141419\\
0.00245230623049831	0.226142029750413\\
0.00253305644424019	0.231155988277615\\
0.00261414282502786	0.23618383588661\\
0.00270594276780885	0.241211683495604\\
0.00280472454928588	0.246239531104599\\
0.00290142383898178	0.251267378713593\\
0.0029927543199381	0.256295226322588\\
0.00308652110575642	0.26130918484979\\
0.00317494327612971	0.266337032458784\\
0.00326249582508418	0.271364880067779\\
0.00335171424054114	0.276392727676773\\
0.00344958389816862	0.281420575285768\\
0.00353732068858351	0.28643453381297\\
0.00364433072935591	0.291462381421964\\
0.00374610191328627	0.296490229030959\\
0.00384660836151392	0.301518076639953\\
0.0039391428781895	0.306545924248948\\
0.00403380249592447	0.31155988277615\\
0.0041280412022353	0.316587730385144\\
0.00422205354139218	0.321615577994139\\
0.00432916359951812	0.326643425603133\\
0.00442704607597927	0.331671273212128\\
0.00453173541967968	0.33668523173933\\
0.00464007772897756	0.341713079348324\\
0.00475156454649358	0.346740926957319\\
0.00485180532503534	0.351768774566313\\
0.0049552761164641	0.356796622175308\\
0.00506917213653685	0.36181058070251\\
0.00518348670093341	0.366838428311504\\
0.00529823416413078	0.371866275920499\\
0.00540689700169881	0.376894123529493\\
0.00552217659982179	0.381921971138488\\
0.00563089369087022	0.38693592966569\\
0.00574675613809363	0.391963777274684\\
0.00586269502046722	0.396991624883679\\
0.00598481141597866	0.402019472492674\\
0.00609728210445221	0.407047320101668\\
0.00620875436408157	0.41206127862887\\
0.00632701773381605	0.417089126237864\\
0.00645199955117067	0.422116973846859\\
0.00657525120465624	0.427144821455854\\
0.00670346341872085	0.432172669064848\\
0.00683326227870864	0.43718662759205\\
0.00696883337750775	0.442214475201044\\
0.00710698495395714	0.447242322810039\\
0.00725125042569994	0.452270170419034\\
0.00738991655219672	0.457298018028028\\
0.00752529860715972	0.46231197655523\\
0.00767082620060962	0.467339824164224\\
0.00781018658353298	0.472367671773219\\
0.00794907742784676	0.477395519382214\\
0.00809143782587992	0.482423366991208\\
0.00822465587126736	0.48743732551841\\
0.00838422817631567	0.492465173127405\\
0.00853127013452885	0.497493020736399\\
0.00869439139639933	0.502520868345394\\
0.00884168450693294	0.507548715954388\\
0.00901656888713912	0.512576563563383\\
0.00918054854611639	0.517590522090585\\
0.00933171347031947	0.522618369699579\\
0.00950050665144086	0.527646217308574\\
0.00968148352778091	0.532674064917568\\
0.00983475840905105	0.537701912526563\\
0.0100021905880357	0.542715871053765\\
0.0101835825974635	0.547743718662759\\
0.010381837326065	0.552771566271754\\
0.0105872580911671	0.557799413880748\\
0.0107824523303727	0.562827261489743\\
0.0109745211431337	0.567841220016945\\
0.0111674217207258	0.572869067625939\\
0.0113669979758017	0.577896915234934\\
0.0115495059159947	0.582924762843928\\
0.0117520847430925	0.587952610452923\\
0.0119523028233697	0.592966568980125\\
0.012149160452431	0.597994416589119\\
0.0123644293979118	0.603022264198114\\
0.0125566438647028	0.608050111807108\\
0.012769675780794	0.613077959416103\\
0.0129847735316194	0.618091917943305\\
0.0131996652602999	0.623119765552299\\
0.0134366231768274	0.628147613161294\\
0.0136461240660289	0.633175460770289\\
0.0139064334935003	0.638203308379283\\
0.0141423564391346	0.643217266906485\\
0.0143897214011669	0.648245114515479\\
0.0146564083053912	0.653272962124474\\
0.0149181709595721	0.658300809733468\\
0.0151820996601321	0.663328657342463\\
0.0154347261845821	0.668342615869665\\
0.0157165955464892	0.673370463478659\\
0.0159964016557499	0.678398311087654\\
0.01633400042882	0.683426158696649\\
0.0166522013908734	0.688454006305643\\
0.0169468660156892	0.693467964832845\\
0.0172579234197309	0.698495812441839\\
0.0175980459009701	0.703523660050834\\
0.0179344627922446	0.708551507659829\\
0.0182805220950963	0.713579355268823\\
0.0186475532385592	0.718593313796025\\
0.0190299766094059	0.72362116140502\\
0.0194234833403564	0.728649009014014\\
0.0198348988334954	0.733676856623009\\
0.0202410364001045	0.738704704232003\\
0.0206858620034234	0.743718662759205\\
0.0211218456082221	0.7487465103682\\
0.021578798837665	0.753774357977194\\
0.0220565012971807	0.758802205586189\\
0.0225231175872462	0.763830053195183\\
0.0230382476157469	0.768857900804178\\
0.0235074884657143	0.77387185933138\\
0.02406275041421	0.778899706940374\\
0.0246947985046681	0.783927554549369\\
0.0253223724777234	0.788955402158363\\
0.0259182935042693	0.793983249767358\\
0.0265525135190524	0.79899720829456\\
0.0272258542078413	0.804025055903554\\
0.0279360659845883	0.809052903512549\\
0.0286781089848584	0.814080751121543\\
0.0294494478860647	0.819108598730538\\
0.0302389374234567	0.82412255725774\\
0.0310727288596119	0.829150404866734\\
0.0318805479278944	0.834178252475729\\
0.0327164061390561	0.839206100084723\\
0.0336393754323989	0.844233947693718\\
0.0345576277071173	0.84924790622092\\
0.0354811659494562	0.854275753829914\\
0.0365890223060552	0.859303601438909\\
0.0376644137752196	0.864331449047903\\
0.0387530967948992	0.869359296656898\\
0.0400238153430112	0.8743732551841\\
0.0412549940662743	0.879401102793094\\
0.0426396807251332	0.884428950402089\\
0.0439581991507599	0.889456798011083\\
0.0453688027415859	0.894484645620078\\
0.0471527823983274	0.89949860414728\\
0.0489220593612389	0.904526451756274\\
0.0509548180089862	0.909554299365269\\
0.0535284973490263	0.914582146974264\\
0.0565022761287715	0.919609994583258\\
0.0603161952690016	0.92462395311046\\
0.0656647695432615	0.929651800719454\\
0.0753860167625859	0.934679648328449\\
0.0897906658365281	0.939707495937444\\
0.105947023966573	0.944735343546438\\
0.141118303763939	0.94974930207364\\
0.183348931950423	0.954777149682634\\
0.274062281158704	0.959804997291629\\
0.334049826122441	0.964832844900624\\
0.372705364368043	0.969860692509618\\
0.397973958302976	0.97487465103682\\
0.453551929259964	0.979902498645815\\
0.602770429009736	0.984930346254809\\
0.734797499754583	0.989958193863804\\
0.885518206255568	0.994986041472798\\
1.01774052113879	1\\
};
\addlegendentry{BCICP \cite{ren2018orientation}}

\addplot [color=mycolor1, line width=2.0pt]
  table[row sep=crcr]{%
0	0\\
0	0.00506979651364678\\
0	0.0100701437599833\\
0	0.0151399402736301\\
0	0.0201402875199667\\
0	0.0251406347663032\\
0	0.03021043127995\\
0	0.0352107785262865\\
0	0.0402111257726231\\
0	0.0452809222862699\\
0	0.0502812695326064\\
0	0.055281616778943\\
0	0.0603514132925898\\
0	0.0653517605389263\\
0	0.0703521077852629\\
0	0.0754219042989096\\
0	0.0804222515452462\\
0	0.085492048058893\\
0	0.0904923953052295\\
0	0.0954927425515661\\
0	0.100562539065213\\
0	0.105562886311549\\
0	0.110563233557886\\
0	0.115633030071533\\
0	0.120633377317869\\
0	0.125633724564206\\
0	0.130703521077853\\
0	0.135703868324189\\
0	0.140704215570526\\
0	0.145774012084173\\
0	0.150774359330509\\
0	0.155844155844156\\
0	0.160844503090492\\
0	0.165844850336829\\
0	0.170914646850476\\
0	0.175914994096812\\
0	0.180915341343149\\
0	0.185985137856796\\
0.000211491684327367	0.190985485103132\\
0.000268285132405657	0.195985832349469\\
0.000350015076280111	0.201055628863115\\
0.000443729308122369	0.206055976109452\\
0.000524622680064517	0.211056323355789\\
0.000626891811183491	0.216126119869435\\
0.000685960337747124	0.221126467115772\\
0.000754972247080468	0.226196263629419\\
0.000826188574539561	0.231196610875755\\
0.000895715013928667	0.236196958122092\\
0.000952981655436017	0.241266754635739\\
0.0010229109174016	0.246267101882075\\
0.00107868568006396	0.251267449128412\\
0.00114380513597542	0.256337245642058\\
0.0012045518662313	0.261337592888395\\
0.0012694593784611	0.266337940134732\\
0.00132175910570458	0.271407736648378\\
0.00137832008546441	0.276408083894715\\
0.00144520758197457	0.281408431141051\\
0.00149517716518658	0.286478227654698\\
0.0015603635485161	0.291478574901035\\
0.00162647103553094	0.296548371414682\\
0.00169333067637252	0.301548718661018\\
0.00176117299177856	0.306549065907355\\
0.00182842557761709	0.311618862421001\\
0.00189768258448746	0.316619209667338\\
0.00197303952804189	0.321619556913675\\
0.00206946128191517	0.326689353427321\\
0.00213219420464622	0.331689700673658\\
0.00219362883650537	0.336690047919994\\
0.0022623478131518	0.341759844433641\\
0.00232700584646438	0.346760191679978\\
0.0024017896464506	0.351760538926314\\
0.00248175871760537	0.356830335439961\\
0.00255360248262575	0.361830682686298\\
0.00262982390959788	0.366900479199944\\
0.00271316546916585	0.371900826446281\\
0.00280744436686697	0.376901173692618\\
0.00288824869364294	0.381970970206264\\
0.00296236196851463	0.386971317452601\\
0.00305242008867686	0.391971664698937\\
0.00312654591230901	0.397041461212584\\
0.00321399128975203	0.402041808458921\\
0.0032976878220469	0.407042155705257\\
0.00339305412871929	0.412111952218904\\
0.00349151950050527	0.417112299465241\\
0.00356086092696019	0.422112646711577\\
0.00364788961037842	0.427182443225224\\
0.00372653510996384	0.432182790471561\\
0.00380960516109109	0.437252586985207\\
0.00387971689651452	0.442252934231544\\
0.00397413217723985	0.44725328147788\\
0.00405845496317836	0.452323077991527\\
0.00417389668547289	0.457323425237864\\
0.00428236697755998	0.4623237724842\\
0.00439153232720963	0.467393568997847\\
0.00449971357930709	0.472393916244184\\
0.00461414851096018	0.47739426349052\\
0.00472888916353342	0.482464060004167\\
0.00481904990181714	0.487464407250503\\
0.00491382076861796	0.49246475449684\\
0.00502829815761864	0.497534551010487\\
0.00514237570159087	0.502534898256823\\
0.00526388723797943	0.50760469477047\\
0.00536794038309807	0.512605042016807\\
0.00551114897365404	0.517605389263143\\
0.00561437740013147	0.52267518577679\\
0.00573271414073742	0.527675533023127\\
0.00586611929885472	0.532675880269463\\
0.00600891348225966	0.53774567678311\\
0.00612043114339127	0.542746024029446\\
0.00623329909547943	0.547746371275783\\
0.00635741366265495	0.55281616778943\\
0.00648170982135432	0.557816515035766\\
0.00664778381877101	0.562816862282103\\
0.00680488828137149	0.56788665879575\\
0.00697100271493342	0.572887006042086\\
0.00712832579782044	0.577956802555733\\
0.00728181040368032	0.58295714980207\\
0.00742681706739877	0.587957497048406\\
0.00758979311810369	0.593027293562053\\
0.00775331900430358	0.598027640808389\\
0.00793849761325233	0.603027988054726\\
0.00812786910963573	0.608097784568373\\
0.00830523055852995	0.613098131814709\\
0.00845175996819029	0.618098479061046\\
0.00861316507429807	0.623168275574693\\
0.00878837367364694	0.628168622821029\\
0.00897050550571022	0.633168970067366\\
0.00918656860394439	0.638238766581013\\
0.00939055009843879	0.643239113827349\\
0.00960690757355657	0.648308910340996\\
0.00984277868560223	0.653309257587332\\
0.0100625726281654	0.658309604833669\\
0.0103032778301821	0.663379401347316\\
0.0105567387637197	0.668379748593652\\
0.010771424740844	0.673380095839989\\
0.0110221951453958	0.678449892353636\\
0.0112267978242985	0.683450239599972\\
0.0114985554789773	0.688450586846309\\
0.0118071222623848	0.693520383359956\\
0.0120421481662788	0.698520730606292\\
0.0123659269614689	0.703521077852629\\
0.0127484439877861	0.708590874366275\\
0.0131104983415373	0.713591221612612\\
0.0134794729281155	0.718661018126259\\
0.013853009575797	0.723661365372595\\
0.0142515485084785	0.728661712618932\\
0.014721404856269	0.733731509132579\\
0.0151353752526218	0.738731856378915\\
0.0156129118101009	0.743732203625252\\
0.0160731725163199	0.748802000138899\\
0.0166367422751772	0.753802347385235\\
0.0171721170398412	0.758802694631572\\
0.0178086400622861	0.763872491145218\\
0.0184037536322682	0.768872838391555\\
0.0190230678559715	0.773873185637892\\
0.0197230016497132	0.778942982151538\\
0.0204372594154921	0.783943329397875\\
0.0214340973147315	0.789013125911522\\
0.0224013426990296	0.794013473157858\\
0.0233188935336849	0.799013820404195\\
0.0245695318781723	0.804083616917841\\
0.0258127787632336	0.809083964164178\\
0.0272181537679028	0.814084311410515\\
0.0284203446262206	0.819154107924161\\
0.030131907243096	0.824154455170498\\
0.0316099502238728	0.829154802416834\\
0.0333154126833632	0.834224598930481\\
0.035018171602118	0.839224946176818\\
0.0370481636137637	0.844225293423154\\
0.038938050588893	0.849295089936801\\
0.0415039374628655	0.854295437183138\\
0.0444336906087614	0.859365233696785\\
0.0474997116438575	0.864365580943121\\
0.0500737537238062	0.869365928189458\\
0.0532429733158724	0.874435724703104\\
0.0568502470355206	0.879436071949441\\
0.0616877656212274	0.884436419195778\\
0.0687730832510373	0.889506215709424\\
0.0770976439360148	0.894506562955761\\
0.0875264088466823	0.899506910202097\\
0.100297831111397	0.904576706715744\\
0.119892005085683	0.909577053962081\\
0.143308993295846	0.914577401208417\\
0.188793209766472	0.919647197722064\\
0.297943994351594	0.924647544968401\\
0.404238192441421	0.929717341482047\\
0.455773436451906	0.934717688728384\\
0.496329298574712	0.93971803597472\\
0.555424061376775	0.944787832488367\\
0.595909588402267	0.949788179734704\\
0.636134408068075	0.95478852698104\\
0.670747480156507	0.959858323494687\\
0.69413159498016	0.964858670741024\\
0.713639358673941	0.96985901798736\\
0.731417750160565	0.974928814501007\\
0.754136717230853	0.979929161747344\\
0.775518972000808	0.98492950899368\\
0.798771174097792	0.989999305507327\\
0.8318472391631	0.994999652753663\\
1.01055920601411	1\\
};
\addlegendentry{Zoomout \cite{melzi2019zoomout}}

\addplot [color=green, line width=2.0pt]
  table[row sep=crcr]{%
0	0.432027397260274\\
0.001	0.432205479452055\\
0.002	0.434205479452055\\
0.003	0.446698630136986\\
0.004	0.487123287671233\\
0.005	0.537109589041096\\
0.006	0.585808219178082\\
0.007	0.629561643835616\\
0.008	0.667397260273973\\
0.009	0.701397260273973\\
0.01	0.731315068493151\\
0.011	0.756205479452055\\
0.012	0.778643835616438\\
0.013	0.799164383561644\\
0.014	0.816657534246576\\
0.015	0.832520547945206\\
0.016	0.846630136986301\\
0.017	0.859123287671233\\
0.018	0.870493150684931\\
0.019	0.880917808219178\\
0.02	0.889986301369863\\
0.021	0.897479452054795\\
0.022	0.904794520547945\\
0.023	0.911301369863014\\
0.024	0.917232876712329\\
0.025	0.922506849315068\\
0.026	0.926972602739726\\
0.027	0.93127397260274\\
0.028	0.935301369863013\\
0.029	0.938780821917808\\
0.03	0.941904109589041\\
0.031	0.945383561643836\\
0.032	0.948219178082191\\
0.033	0.95086301369863\\
0.034	0.95331506849315\\
0.035	0.955438356164384\\
0.036	0.957493150684932\\
0.037	0.959520547945206\\
0.038	0.961205479452055\\
0.039	0.962794520547945\\
0.04	0.96458904109589\\
0.041	0.966191780821918\\
0.042	0.96772602739726\\
0.043	0.969027397260274\\
0.044	0.970342465753425\\
0.045	0.971589041095891\\
0.046	0.972739726027397\\
0.047	0.973958904109589\\
0.048	0.975287671232876\\
0.049	0.976260273972602\\
0.05	0.977506849315068\\
0.051	0.97827397260274\\
0.052	0.97913698630137\\
0.053	0.97986301369863\\
0.054	0.98058904109589\\
0.055	0.981191780821917\\
0.056	0.981671232876712\\
0.057	0.98213698630137\\
0.058	0.982684931506849\\
0.059	0.983041095890411\\
0.06	0.983369863013698\\
0.061	0.983767123287671\\
0.062	0.984109589041095\\
0.063	0.984438356164383\\
0.064	0.984712328767123\\
0.065	0.984876712328767\\
0.066	0.985123287671233\\
0.067	0.985397260273973\\
0.068	0.985561643835616\\
0.069	0.985794520547945\\
0.07	0.985931506849315\\
0.071	0.986041095890411\\
0.072	0.986164383561644\\
0.073	0.986246575342466\\
0.074	0.986342465753425\\
0.075	0.986397260273973\\
0.076	0.986534246575342\\
0.077	0.986671232876713\\
0.078	0.986808219178082\\
0.079	0.986904109589041\\
0.08	0.987\\
0.081	0.987068493150685\\
0.082	0.987164383561644\\
0.083	0.987246575342466\\
0.084	0.987315068493151\\
0.085	0.987465753424657\\
0.086	0.987547945205479\\
0.087	0.987616438356164\\
0.088	0.987753424657534\\
0.089	0.987794520547945\\
0.09	0.987835616438356\\
0.091	0.987876712328767\\
0.092	0.987945205479452\\
0.093	0.988\\
0.094	0.988123287671233\\
0.095	0.988205479452055\\
0.096	0.98827397260274\\
0.097	0.988369863013699\\
0.098	0.988424657534247\\
0.099	0.988534246575342\\
0.1	0.988561643835616\\
0.101	0.988657534246575\\
0.102	0.988684931506849\\
0.103	0.988698630136986\\
0.104	0.988767123287671\\
0.105	0.988849315068493\\
0.106	0.988958904109589\\
0.107	0.989082191780822\\
0.108	0.989164383561644\\
0.109	0.989191780821918\\
0.11	0.989205479452055\\
0.111	0.989260273972603\\
0.112	0.989315068493151\\
0.113	0.989328767123288\\
0.114	0.98941095890411\\
0.115	0.989465753424657\\
0.116	0.989534246575342\\
0.117	0.98958904109589\\
0.118	0.989657534246575\\
0.119	0.989739726027397\\
0.12	0.989835616438356\\
0.121	0.989890410958904\\
0.122	0.989958904109589\\
0.123	0.99\\
0.124	0.990013698630137\\
0.125	0.990082191780822\\
0.126	0.990164383561644\\
0.127	0.990232876712329\\
0.128	0.99027397260274\\
0.129	0.990342465753425\\
0.13	0.990424657534247\\
0.131	0.990479452054795\\
0.132	0.990547945205479\\
0.133	0.990602739726027\\
0.134	0.990616438356164\\
0.135	0.990712328767123\\
0.136	0.99072602739726\\
0.137	0.990739726027397\\
0.138	0.990821917808219\\
0.139	0.990849315068493\\
0.14	0.990876712328767\\
0.141	0.990904109589041\\
0.142	0.990945205479452\\
0.143	0.991\\
0.144	0.991013698630137\\
0.145	0.991109589041096\\
0.146	0.99113698630137\\
0.147	0.991219178082192\\
0.148	0.991301369863014\\
0.149	0.991356164383562\\
0.15	0.991438356164383\\
0.151	0.991561643835616\\
0.152	0.99158904109589\\
0.153	0.991630136986301\\
0.154	0.991712328767123\\
0.155	0.991753424657534\\
0.156	0.991780821917808\\
0.157	0.991849315068493\\
0.158	0.991890410958904\\
0.159	0.991890410958904\\
0.16	0.991945205479452\\
0.161	0.991958904109589\\
0.162	0.992\\
0.163	0.992027397260274\\
0.164	0.992068493150685\\
0.165	0.992123287671233\\
0.166	0.992123287671233\\
0.167	0.992164383561644\\
0.168	0.992219178082192\\
0.169	0.992246575342466\\
0.17	0.992301369863014\\
0.171	0.992369863013699\\
0.172	0.992424657534247\\
0.173	0.992479452054794\\
0.174	0.992534246575342\\
0.175	0.992561643835616\\
0.176	0.992616438356164\\
0.177	0.992643835616438\\
0.178	0.992657534246575\\
0.179	0.992753424657534\\
0.18	0.992808219178082\\
0.181	0.99286301369863\\
0.182	0.992904109589041\\
0.183	0.992972602739726\\
0.184	0.993041095890411\\
0.185	0.993095890410959\\
0.186	0.993109589041096\\
0.187	0.99313698630137\\
0.188	0.993150684931507\\
0.189	0.993178082191781\\
0.19	0.993191780821918\\
0.191	0.993191780821918\\
0.192	0.993232876712329\\
0.193	0.993301369863014\\
0.194	0.993342465753425\\
0.195	0.993342465753425\\
0.196	0.993369863013699\\
0.197	0.993383561643836\\
0.198	0.993397260273973\\
0.199	0.99341095890411\\
0.2	0.993465753424658\\
0.201	0.993465753424658\\
0.202	0.993520547945205\\
0.203	0.993534246575342\\
0.204	0.99358904109589\\
0.205	0.993602739726027\\
0.206	0.993616438356164\\
0.207	0.993630136986301\\
0.208	0.993643835616438\\
0.209	0.993657534246575\\
0.21	0.993657534246575\\
0.211	0.993684931506849\\
0.212	0.993684931506849\\
0.213	0.99372602739726\\
0.214	0.993767123287671\\
0.215	0.993780821917808\\
0.216	0.993835616438356\\
0.217	0.99386301369863\\
0.218	0.993917808219178\\
0.219	0.993931506849315\\
0.22	0.993986301369863\\
0.221	0.994\\
0.222	0.994013698630137\\
0.223	0.994041095890411\\
0.224	0.994054794520548\\
0.225	0.994095890410959\\
0.226	0.994109589041096\\
0.227	0.994123287671233\\
0.228	0.994123287671233\\
0.229	0.99413698630137\\
0.23	0.99413698630137\\
0.231	0.994164383561644\\
0.232	0.994178082191781\\
0.233	0.994219178082192\\
0.234	0.994219178082192\\
0.235	0.994232876712329\\
0.236	0.994260273972603\\
0.237	0.994287671232877\\
0.238	0.994301369863014\\
0.239	0.994315068493151\\
0.24	0.994328767123288\\
0.241	0.994342465753425\\
0.242	0.994356164383562\\
0.243	0.99441095890411\\
0.244	0.994438356164383\\
0.245	0.994493150684931\\
0.246	0.994493150684931\\
0.247	0.994520547945205\\
0.248	0.994547945205479\\
0.249	0.994575342465753\\
0.25	0.994602739726027\\
};
\addlegendentry{KM \cite{kernel17}}

\addplot [color=mycolor2, line width=2.0pt]
  table[row sep=crcr]{%
0	0.210526315789474\\
0.00114810562571757	0.229018492176387\\
0.00172215843857635	0.2475106685633\\
0.00229621125143513	0.266002844950213\\
0.00287026406429392	0.285917496443812\\
0.00315729047072331	0.30298719772404\\
0.00373134328358209	0.322901849217639\\
0.00430539609644087	0.341394025604552\\
0.00487944890929966	0.361308677098151\\
0.00545350172215844	0.378378378378378\\
0.00602755453501722	0.398293029871977\\
0.00660160734787601	0.415362731152205\\
0.00717566016073479	0.435277382645804\\
0.00832376578645235	0.453769559032717\\
0.00889781859931114	0.469416785206259\\
0.0100459242250287	0.487908961593172\\
0.0106199770378875	0.5049786628734\\
0.0111940298507463	0.51778093883357\\
0.0120551090700344	0.532005689900427\\
0.0126291618828932	0.547652916073969\\
0.0134902411021814	0.560455192034139\\
0.0143513203214696	0.577524893314367\\
0.0149253731343284	0.593172119487909\\
0.0160734787600459	0.610241820768137\\
0.0172215843857635	0.627311522048364\\
0.0183696900114811	0.642958748221906\\
0.0195177956371986	0.658605974395448\\
0.0206659012629162	0.672830725462304\\
0.0215269804822044	0.688477951635846\\
0.0229621125143513	0.702702702702703\\
0.0241102181400689	0.714082503556188\\
0.0255453501722158	0.72972972972973\\
0.0269804822043628	0.7425320056899\\
0.0287026406429392	0.753911806543386\\
0.0301377726750861	0.765291607396871\\
0.0315729047072331	0.776671408250356\\
0.0335820895522388	0.789473684210526\\
0.0350172215843858	0.800853485064011\\
0.0373134328358209	0.813655761024182\\
0.0390355912743972	0.825035561877667\\
0.041044776119403	0.834992887624467\\
0.0433409873708381	0.847795163584637\\
0.0456371986222733	0.857752489331437\\
0.0482204362801378	0.867709815078236\\
0.0510907003444317	0.877667140825036\\
0.054247990815155	0.884779516358464\\
0.0574052812858783	0.893314366998578\\
0.0605625717566016	0.903271692745377\\
0.0637198622273249	0.907539118065434\\
0.067451205510907	0.913229018492176\\
0.0714695752009185	0.920341394025605\\
0.0749138920780712	0.920341394025605\\
0.0783582089552239	0.924608819345662\\
0.0812284730195178	0.930298719772404\\
0.0849598163030999	0.934566145092461\\
0.0889781859931114	0.937411095305832\\
0.0932835820895522	0.940256045519203\\
0.0967278989667049	0.943100995732575\\
0.100746268656716	0.94452347083926\\
0.104477611940299	0.945945945945946\\
0.107634902411022	0.948790896159317\\
0.111653272101033	0.953058321479374\\
0.115384615384615	0.955903271692745\\
0.119402985074627	0.958748221906117\\
0.12284730195178	0.960170697012802\\
0.12743972445465	0.963015647226173\\
0.131458094144661	0.965860597439545\\
0.134615384615385	0.965860597439545\\
0.138059701492537	0.967283072546231\\
0.140929965556831	0.965860597439545\\
0.144087256027555	0.967283072546231\\
0.147531572904707	0.967283072546231\\
0.151549942594719	0.967283072546231\\
0.154994259471871	0.968705547652916\\
0.159299655568312	0.970128022759602\\
0.163030998851894	0.970128022759602\\
0.167336394948335	0.968705547652916\\
0.171354764638347	0.968705547652916\\
0.175373134328358	0.970128022759602\\
0.179678530424799	0.970128022759602\\
0.18398392652124	0.971550497866287\\
0.188289322617681	0.971550497866287\\
0.192020665901263	0.971550497866287\\
0.196900114810563	0.972972972972973\\
0.200918484500574	0.972972972972973\\
0.205223880597015	0.974395448079659\\
0.209242250287026	0.974395448079659\\
0.213260619977038	0.972972972972973\\
0.21699196326062	0.974395448079659\\
0.221297359357061	0.974395448079659\\
0.224741676234214	0.975817923186344\\
0.228473019517796	0.975817923186344\\
0.232204362801378	0.975817923186344\\
0.23593570608496	0.97724039829303\\
0.239380022962113	0.97724039829303\\
0.243398392652124	0.978662873399716\\
0.247416762342135	0.97724039829303\\
0.25	0.978662873399716\\
};
\addlegendentry{FM \cite{ovsjanikov2012functional}}

\addplot [color=mycolor3, line width=2.0pt]
  table[row sep=crcr]{%
0	0.00950928\\
0.0025	0.0464478\\
0.005	0.0949219\\
0.0075	0.150183\\
0.01	0.210095\\
0.0125	0.271948\\
0.015	0.333997\\
0.0175	0.395203\\
0.02	0.453137\\
0.0225	0.506812\\
0.025	0.559033\\
0.0275	0.604321\\
0.03	0.646362\\
0.0325	0.68252\\
0.035	0.71571\\
0.0375	0.744446\\
0.04	0.77002\\
0.0425	0.792944\\
0.045	0.812939\\
0.0475	0.830432\\
0.05	0.846729\\
0.0525	0.862305\\
0.055	0.875488\\
0.0575	0.886487\\
0.06	0.896265\\
0.0625	0.904895\\
0.065	0.914148\\
0.0675	0.92146\\
0.07	0.928247\\
0.0725	0.934058\\
0.075	0.939404\\
0.0775	0.943921\\
0.08	0.947803\\
0.0825	0.950964\\
0.085	0.953918\\
0.0875	0.956604\\
0.09	0.958862\\
0.0925	0.96084\\
0.095	0.962659\\
0.0975	0.964355\\
0.1	0.96593\\
0.1025	0.967273\\
0.105	0.968311\\
0.1075	0.969165\\
0.11	0.970288\\
0.1125	0.971191\\
0.115	0.971826\\
0.1175	0.972656\\
0.12	0.973376\\
0.1225	0.97417\\
0.125	0.974915\\
0.1275	0.9755\\
0.13	0.976172\\
0.1325	0.976758\\
0.135	0.97738\\
0.1375	0.977954\\
0.14	0.978394\\
0.1425	0.978943\\
0.145	0.979468\\
0.1475	0.979944\\
0.15	0.980322\\
0.1525	0.980688\\
0.155	0.981079\\
0.1575	0.98158\\
0.16	0.982104\\
0.1625	0.982422\\
0.165	0.98291\\
0.1675	0.983374\\
0.17	0.983765\\
0.1725	0.984143\\
0.175	0.984497\\
0.1775	0.984863\\
0.18	0.985205\\
0.1825	0.985706\\
0.185	0.986084\\
0.1875	0.986438\\
0.19	0.986853\\
0.1925	0.987097\\
0.195	0.987549\\
0.1975	0.987817\\
0.2	0.988123\\
0.2025	0.988416\\
0.205	0.988745\\
0.2075	0.988989\\
0.21	0.989197\\
0.2125	0.989417\\
0.215	0.989709\\
0.2175	0.98988\\
0.22	0.990088\\
0.2225	0.990417\\
0.225	0.990674\\
0.2275	0.99082\\
0.23	0.990979\\
0.2325	0.991138\\
0.235	0.991235\\
0.2375	0.991382\\
0.24	0.991516\\
0.2425	0.991638\\
0.245	0.991797\\
0.2475	0.99187\\
};
\addlegendentry{BIM \cite{kim11}}

\addplot [color=red, line width=2.0pt]
  table[row sep=crcr]{%
0	0\\
0	0.00506979651364678\\
0	0.0100701437599833\\
0	0.0151399402736301\\
0	0.0201402875199667\\
0	0.0251406347663032\\
0	0.03021043127995\\
0	0.0352107785262865\\
0	0.0402111257726231\\
0	0.0452809222862699\\
0	0.0502812695326064\\
0	0.055281616778943\\
0	0.0603514132925898\\
0	0.0653517605389263\\
0	0.0703521077852629\\
0	0.0754219042989096\\
0	0.0804222515452462\\
0	0.085492048058893\\
0	0.0904923953052295\\
0	0.0954927425515661\\
0	0.100562539065213\\
0	0.105562886311549\\
0	0.110563233557886\\
0	0.115633030071533\\
0	0.120633377317869\\
0	0.125633724564206\\
0	0.130703521077853\\
0	0.135703868324189\\
0	0.140704215570526\\
0	0.145774012084173\\
0	0.150774359330509\\
0	0.155844155844156\\
0	0.160844503090492\\
0	0.165844850336829\\
0	0.170914646850476\\
0	0.175914994096812\\
0	0.180915341343149\\
0	0.185985137856796\\
0	0.190985485103132\\
0	0.195985832349469\\
0	0.201055628863115\\
0	0.206055976109452\\
0	0.211056323355789\\
0	0.216126119869435\\
0	0.221126467115772\\
0	0.226196263629419\\
0	0.231196610875755\\
0	0.236196958122092\\
0	0.241266754635739\\
0	0.246267101882075\\
0	0.251267449128412\\
0	0.256337245642058\\
0	0.261337592888395\\
0	0.266337940134732\\
0	0.271407736648378\\
0	0.276408083894715\\
0	0.281408431141051\\
0	0.286478227654698\\
0	0.291478574901035\\
0	0.296548371414682\\
0	0.301548718661018\\
0	0.306549065907355\\
0	0.311618862421001\\
0	0.316619209667338\\
0	0.321619556913675\\
0	0.326689353427321\\
0	0.331689700673658\\
0	0.336690047919994\\
0	0.341759844433641\\
0	0.346760191679978\\
0	0.351760538926314\\
0	0.356830335439961\\
0	0.361830682686298\\
0	0.366900479199944\\
0	0.371900826446281\\
0	0.376901173692618\\
0	0.381970970206264\\
0	0.386971317452601\\
0	0.391971664698937\\
0	0.397041461212584\\
0	0.402041808458921\\
0.000120165042398057	0.407042155705257\\
0.000204366632860666	0.412111952218904\\
0.00025147014649241	0.417112299465241\\
0.000308868209887815	0.422112646711577\\
0.000372302024723879	0.427182443225224\\
0.000423710624968058	0.432182790471561\\
0.000470630410465524	0.437252586985207\\
0.000515054400112601	0.442252934231544\\
0.000565352772541189	0.44725328147788\\
0.000613515072763741	0.452323077991527\\
0.000659621548640031	0.457323425237864\\
0.000709173870295065	0.4623237724842\\
0.000757087077104579	0.467393568997847\\
0.000802749057997354	0.472393916244184\\
0.000848218421135539	0.47739426349052\\
0.000890531662208552	0.482464060004167\\
0.000943350849346319	0.487464407250503\\
0.000998373679301422	0.49246475449684\\
0.00104286009292054	0.497534551010487\\
0.001093367820654	0.502534898256823\\
0.0011496866827371	0.50760469477047\\
0.00119196129142866	0.512605042016807\\
0.00123644223115159	0.517605389263143\\
0.0012892800832975	0.52267518577679\\
0.00134000634638883	0.527675533023127\\
0.00138350815592851	0.532675880269463\\
0.00143279949079583	0.53774567678311\\
0.001488900637742	0.542746024029446\\
0.00154260106871545	0.547746371275783\\
0.00159684812625581	0.55281616778943\\
0.00165498224344074	0.557816515035766\\
0.0017221883553527	0.562816862282103\\
0.00177834009960805	0.56788665879575\\
0.00183734716352677	0.572887006042086\\
0.0019049616012232	0.577956802555733\\
0.00196117053193163	0.58295714980207\\
0.00202995847427025	0.587957497048406\\
0.00209379440387002	0.593027293562053\\
0.00217062092742298	0.598027640808389\\
0.00222395774190835	0.603027988054726\\
0.00229239388071125	0.608097784568373\\
0.00235460979063479	0.613098131814709\\
0.00242457579830396	0.618098479061046\\
0.00249469923602282	0.623168275574693\\
0.00256596467219958	0.628168622821029\\
0.00263839026628041	0.633168970067366\\
0.00271951931098941	0.638238766581013\\
0.00280400723394929	0.643239113827349\\
0.00286870364304875	0.648308910340996\\
0.00295485732501506	0.653309257587332\\
0.00303657696423647	0.658309604833669\\
0.00310989384124812	0.663379401347316\\
0.00318150645289108	0.668379748593652\\
0.00326476520274154	0.673380095839989\\
0.00334217691737825	0.678449892353636\\
0.00342334771524016	0.683450239599972\\
0.00349853486720237	0.688450586846309\\
0.00359373864093642	0.693520383359956\\
0.00367671059829149	0.698520730606292\\
0.00376660578479391	0.703521077852629\\
0.00387382778533386	0.708590874366275\\
0.00398037958771791	0.713591221612612\\
0.00408241624751255	0.718661018126259\\
0.00418861435939462	0.723661365372595\\
0.0042877208982912	0.728661712618932\\
0.00440702137244091	0.733731509132579\\
0.00450559714781565	0.738731856378915\\
0.00463327536052779	0.743732203625252\\
0.0047272570974601	0.748802000138899\\
0.00484675976012676	0.753802347385235\\
0.0049783654828535	0.758802694631572\\
0.00509990507769869	0.763872491145218\\
0.00520379101239078	0.768872838391555\\
0.00533403464172847	0.773873185637892\\
0.00545020917197964	0.778942982151538\\
0.00555723419177567	0.783943329397875\\
0.00569630720312722	0.789013125911522\\
0.0058461812489431	0.794013473157858\\
0.00600151085498278	0.799013820404195\\
0.00615262089270361	0.804083616917841\\
0.00629591766533095	0.809083964164178\\
0.00644191652749851	0.814084311410515\\
0.00661027281824026	0.819154107924161\\
0.00679711697341277	0.824154455170498\\
0.00700571637868548	0.829154802416834\\
0.00717127876722064	0.834224598930481\\
0.00735183568212329	0.839224946176818\\
0.00758429670510893	0.844225293423154\\
0.0078309617215551	0.849295089936801\\
0.00806815301402858	0.854295437183138\\
0.00824709403922861	0.859365233696785\\
0.00854638460895628	0.864365580943121\\
0.0088496148028419	0.869365928189458\\
0.00916096254409954	0.874435724703104\\
0.00950403629439547	0.879436071949441\\
0.00991264543578549	0.884436419195778\\
0.0102519708493892	0.889506215709424\\
0.0106345776676444	0.894506562955761\\
0.0111147141353728	0.899506910202097\\
0.0115890380881005	0.904576706715744\\
0.0120232085824306	0.909577053962081\\
0.0125690088506031	0.914577401208417\\
0.0130619485212835	0.919647197722064\\
0.0137903981107612	0.924647544968401\\
0.0145753757049278	0.929717341482047\\
0.0156009471036631	0.934717688728384\\
0.0169892952077474	0.93971803597472\\
0.0189394672423567	0.944787832488367\\
0.0207768029824162	0.949788179734704\\
0.0229876179730339	0.95478852698104\\
0.0270957989869927	0.959858323494687\\
0.0313281051554231	0.964858670741024\\
0.0373060447324219	0.96985901798736\\
0.045582199209792	0.974928814501007\\
0.0545146705447825	0.979929161747344\\
0.0640537143683283	0.98492950899368\\
0.0713977514090491	0.989999305507327\\
0.0810333965885433	0.994999652753663\\
0.735038296421969	1\\
};
\addlegendentry{Ours}

\end{axis}
\end{tikzpicture}%

%% file: figures/curve_scape.tikz
\definecolor{mycolor1}{rgb}{0.00000,1.00000,1.00000}%
\definecolor{mycolor2}{rgb}{0.91000,0.41000,0.17000}%
\definecolor{mycolor3}{rgb}{1.00000,0.84314,0.00000}%
\begin{tikzpicture}

\begin{axis}[%
width=0.18\linewidth,
height=0.18\linewidth,
at={(0,0)},
scale only axis,
xmin=0,
xmax=0.1,
xlabel style={font=\color{white!15!black}},
xlabel={Relative geodesic error},
every x tick label/.append style={font=\color{black}, font=\footnotesize},
every y tick label/.append style={font=\color{black}, font=\footnotesize},
x tick label style={/pgf/number format/fixed},
ymin=0,
ymax=1,
ylabel style={font=\color{white!15!black}},
axis background/.style={fill=white},
title style={font=\bfseries},
title={SCAPE},
xmajorgrids,
ymajorgrids
]

\addplot [color=blue, line width=2.0pt]
  table[row sep=crcr]{%
0	0\\
0	0.00502823983436386\\
0	0.0100564796687277\\
0	0.0150847195030916\\
0	0.0201129593374555\\
0	0.0251271144664009\\
0	0.0301553543007648\\
0	0.0351835941351287\\
0	0.0402118339694925\\
0	0.0452400738038564\\
0	0.0502542289328019\\
0	0.0552824687671657\\
0	0.0603107086015296\\
0	0.0653389484358935\\
0.00246078931165245	0.0703531035648389\\
0.00321544796964408	0.0753813433992028\\
0.00356668809841241	0.0804095832335667\\
0.00388870538295035	0.0854378230679305\\
0.00412386715315813	0.0904660629022944\\
0.00432105828727712	0.0954802180312399\\
0.00448250745418172	0.100508457865604\\
0.00463365410549867	0.105536697699968\\
0.00479007794182088	0.110564937534331\\
0.00491773515843845	0.115579092663277\\
0.00505128193183506	0.120607332497641\\
0.00517871869902929	0.125635572332005\\
0.00529759560975811	0.130663812166369\\
0.00541035244827271	0.135692052000732\\
0.00551689603300282	0.140706207129678\\
0.00562379944059836	0.145734446964042\\
0.00573195812793884	0.150762686798406\\
0.00582617752438012	0.155790926632769\\
0.00592719844269546	0.160805081761715\\
0.00603570113009391	0.165833321596079\\
0.00613618142232177	0.170861561430443\\
0.00623809469212751	0.175889801264807\\
0.00634351671695026	0.18091804109917\\
0.00643655800058295	0.185932196228116\\
0.00653799150305161	0.19096043606248\\
0.00663485285104433	0.195988675896844\\
0.00673076523381786	0.201016915731207\\
0.00683554826600609	0.206031070860153\\
0.00692468799332381	0.211059310694517\\
0.00701297079060169	0.216087550528881\\
0.00709919589369016	0.221115790363245\\
0.00720419448202987	0.226144030197608\\
0.00729853146358662	0.231158185326554\\
0.00739599303239833	0.236186425160918\\
0.00749435701598021	0.241214664995282\\
0.0075942697356131	0.246242904829645\\
0.00768886258891723	0.251257059958591\\
0.00778228905965416	0.256285299792955\\
0.0078807910690683	0.261313539627319\\
0.00798485183987567	0.266341779461683\\
0.00808735181545032	0.271370019296046\\
0.00819447471162215	0.276384174424992\\
0.00829512783901383	0.281412414259356\\
0.00838796856546219	0.28644065409372\\
0.00848376721247489	0.291468893928084\\
0.00858434229169948	0.296483049057029\\
0.008703565580618	0.301511288891393\\
0.00881254723228503	0.306539528725757\\
0.00892218311986507	0.311567768560121\\
0.0090365860096572	0.316596008394484\\
0.00914274642240911	0.32161016352343\\
0.00925527748236898	0.326638403357794\\
0.00936803808027537	0.331666643192158\\
0.00948402804202233	0.336694883026522\\
0.00959561665475995	0.341709038155467\\
0.00971282703024546	0.346737277989831\\
0.00982753754816799	0.351765517824195\\
0.00994838026687649	0.356793757658559\\
0.0100695166907445	0.361821997492922\\
0.0101945499376022	0.366836152621868\\
0.0103188840822361	0.371864392456232\\
0.0104324635548294	0.376892632290596\\
0.0105507230226818	0.38192087212496\\
0.0106717306180495	0.386935027253905\\
0.0107944806783091	0.391963267088269\\
0.0109093369568365	0.396991506922633\\
0.0110407913680759	0.402019746756997\\
0.0111650621436095	0.40704798659136\\
0.011282072247286	0.412062141720306\\
0.0114090482172946	0.41709038155467\\
0.0115365450213918	0.422118621389034\\
0.0116658899849186	0.427146861223398\\
0.0118096550723639	0.432161016352343\\
0.011954566304608	0.437189256186707\\
0.0120857508772166	0.442217496021071\\
0.012200126240984	0.447245735855435\\
0.012340781914269	0.452273975689798\\
0.0124693608799393	0.457288130818744\\
0.0126100781280131	0.462316370653108\\
0.0127469323803646	0.467344610487472\\
0.0128882455962925	0.472372850321836\\
0.0130390481328805	0.477387005450781\\
0.0131554791602181	0.482415245285145\\
0.0132975907366055	0.487443485119509\\
0.0134445346942412	0.492471724953873\\
0.0135805609087756	0.497499964788236\\
0.013727052976998	0.502514119917182\\
0.0138814819646495	0.507542359751546\\
0.0140306062283022	0.51257059958591\\
0.0141862621146516	0.517598839420274\\
0.0143466462854179	0.522627079254637\\
0.0145083199300702	0.527641234383583\\
0.0146729596352501	0.532669474217947\\
0.0148343682085324	0.537697714052311\\
0.0150036975043207	0.542725953886674\\
0.0151709604821638	0.54774010901562\\
0.0153464197780719	0.552768348849984\\
0.0155111278985835	0.557796588684348\\
0.0157087264218646	0.562824828518712\\
0.015901825041998	0.567853068353075\\
0.0160827628920136	0.572867223482021\\
0.0162691220944308	0.577895463316385\\
0.0164824616424596	0.582923703150749\\
0.01669355757849	0.587951942985112\\
0.0169054415394602	0.592966098114058\\
0.0171256990379941	0.597994337948422\\
0.0173353587022988	0.603022577782786\\
0.0175632332137334	0.60805081761715\\
0.0177819911578609	0.613079057451513\\
0.0180172743420865	0.618093212580459\\
0.0182467022465962	0.623121452414823\\
0.0184806293454146	0.628149692249187\\
0.0187375573451293	0.63317793208355\\
0.0189999040980569	0.638192087212496\\
0.0192699446020687	0.64322032704686\\
0.019536877836796	0.648248566881224\\
0.0198202530222983	0.653276806715588\\
0.0200846265084425	0.658305046549951\\
0.0204086022932131	0.663319201678897\\
0.0207284680309578	0.668347441513261\\
0.02107561084275	0.673375681347625\\
0.0214210621062039	0.678403921181988\\
0.0217853339122383	0.683418076310934\\
0.0221177703508742	0.688446316145298\\
0.022474194597166	0.693474555979662\\
0.0228146971627411	0.698502795814026\\
0.0231815930198799	0.703531035648389\\
0.0235697451897969	0.708545190777335\\
0.0239595286728506	0.713573430611699\\
0.0243795870209122	0.718601670446063\\
0.024832509961166	0.723629910280426\\
0.0252764081150025	0.728644065409372\\
0.0257696027505727	0.733672305243736\\
0.0262834441045471	0.7387005450781\\
0.0268463780002419	0.743728784912464\\
0.027499147107007	0.748757024746827\\
0.0281015600853169	0.753771179875773\\
0.0287480789706855	0.758799419710137\\
0.0293974978122328	0.763827659544501\\
0.0301573011258883	0.768855899378864\\
0.0309560683744972	0.77387005450781\\
0.0317015870270016	0.778898294342174\\
0.0324329706745266	0.783926534176538\\
0.033345819399708	0.788954774010902\\
0.0342304119224992	0.793983013845265\\
0.0352389723209354	0.798997168974211\\
0.0362768613079104	0.804025408808575\\
0.0374485046735272	0.809053648642939\\
0.0386738473890684	0.814081888477302\\
0.0401105868115102	0.819096043606248\\
0.0416513088350252	0.824124283440612\\
0.0431844500359189	0.829152523274976\\
0.0448673794436025	0.83418076310934\\
0.0467892187003953	0.839209002943703\\
0.048844743743257	0.844223158072649\\
0.0509053417237759	0.849251397907013\\
0.0531619479634305	0.854279637741377\\
0.0560668359701053	0.85930787757574\\
0.0597177559736186	0.864322032704686\\
0.0640209688999538	0.86935027253905\\
0.0687103447831162	0.874378512373414\\
0.0749273615372661	0.879406752207778\\
0.0813479290827706	0.884434992042141\\
0.0878740526760736	0.889449147171087\\
0.0949634907014995	0.894477387005451\\
0.103598662256432	0.899505626839815\\
0.113780295165124	0.904533866674178\\
0.127865322482752	0.909548021803124\\
0.14823725731342	0.914576261637488\\
0.17950427058054	0.919604501471852\\
0.194095740909895	0.924632741306216\\
0.202488947358989	0.929660981140579\\
0.210831967222994	0.934675136269525\\
0.220596120575424	0.939703376103889\\
0.229860575424707	0.944731615938253\\
0.243079263661069	0.949759855772617\\
0.279686210069042	0.954774010901562\\
0.330657125177442	0.959802250735926\\
0.375841242254254	0.96483049057029\\
0.421676506239244	0.969858730404654\\
0.473792103387806	0.974886970239017\\
0.546169107914485	0.979901125367963\\
0.602693886479245	0.984929365202327\\
0.642702579192144	0.989957605036691\\
0.680415558059239	0.994985844871055\\
0.894445214571482	1\\
};

\addplot [color=mycolor1, line width=2.0pt]
  table[row sep=crcr]{%
0	0\\
0	0.00507077963236848\\
0	0.0100711317698429\\
0	0.0151419114022114\\
0	0.0201422635396859\\
0	0.0251426156771604\\
0	0.0302133953095288\\
0	0.0352137474470033\\
0	0.0402140995844778\\
0	0.0452848792168463\\
0	0.0502852313543207\\
0	0.0552855834917952\\
0	0.0603563631241637\\
0	0.0653567152616381\\
0	0.0703570673991126\\
0	0.0754278470314811\\
0	0.0804281991689556\\
0	0.08542855130643\\
0	0.0904993309387985\\
0	0.095499683076273\\
0	0.100570462708641\\
0	0.105570814846116\\
0.00253314212597733	0.11057116698359\\
0.00307144231310373	0.115641946615959\\
0.00340133280876387	0.120642298753433\\
0.00364787377135958	0.125642650890908\\
0.00384388560669031	0.130713430523276\\
0.00402641724007299	0.135713782660751\\
0.00414180633329485	0.140714134798225\\
0.0042967638743782	0.145784914430594\\
0.00441425379648349	0.150785266568068\\
0.00453063745370509	0.155785618705543\\
0.00465695203424181	0.160856398337911\\
0.00476856250140398	0.165856750475386\\
0.00488538859022292	0.17085710261286\\
0.00499381551426597	0.175927882245229\\
0.00509132629320175	0.180928234382703\\
0.00516850004596802	0.185999014015071\\
0.00525363534721227	0.190999366152546\\
0.00534200020215862	0.19599971829002\\
0.00544624304348634	0.201070497922389\\
0.00553535660997717	0.206070850059863\\
0.00560753907555416	0.211071202197338\\
0.00571872305474329	0.216141981829706\\
0.00580147528805724	0.221142333967181\\
0.0058966103116625	0.226142686104655\\
0.00597857642754643	0.231213465737024\\
0.00603559186483708	0.236213817874498\\
0.00611466030325572	0.241214170011973\\
0.00620829035169507	0.246284949644341\\
0.0062918440737323	0.251285301781816\\
0.00636169621297692	0.25628565391929\\
0.00645155504490653	0.261356433551659\\
0.00654405928668706	0.266356785689133\\
0.00665565826361079	0.271357137826608\\
0.00673445441224031	0.276427917458976\\
0.00682937946947996	0.28142826959645\\
0.00693391124192328	0.286499049228819\\
0.00702581981215589	0.291499401366293\\
0.00711712309392237	0.296499753503768\\
0.00719090191339568	0.301570533136136\\
0.00727351024164416	0.306570885273611\\
0.00738866189881256	0.311571237411085\\
0.00747655030274638	0.316642017043454\\
0.00755395982072063	0.321642369180928\\
0.00763607465805839	0.326642721318403\\
0.00775123040130142	0.331713500950771\\
0.00783784876184533	0.336713853088246\\
0.00795031393206817	0.34171420522572\\
0.00805336634458627	0.346784984858089\\
0.00814089219801582	0.351785336995563\\
0.00823245405375768	0.356785689133038\\
0.00837538840291723	0.361856468765406\\
0.00848487588107928	0.36685682090288\\
0.00858228402909478	0.371927600535249\\
0.00869096516724663	0.376927952672723\\
0.00882072880734259	0.381928304810198\\
0.00893249496140049	0.386999084442566\\
0.00902319503542844	0.391999436580041\\
0.00914152963234	0.396999788717515\\
0.00923629106870624	0.402070568349884\\
0.00933585646904733	0.407070920487358\\
0.00947037651777733	0.412071272624833\\
0.00959876781220912	0.417142052257201\\
0.00971924443586517	0.422142404394676\\
0.00984820629028511	0.42714275653215\\
0.00996246898598044	0.432213536164519\\
0.0100780090429929	0.437213888301993\\
0.0102114029953102	0.442214240439468\\
0.0103362542416394	0.447285020071836\\
0.0104587700651186	0.452285372209311\\
0.0105809944663875	0.457356151841679\\
0.0107029319186804	0.462356503979153\\
0.0108178071097287	0.467356856116628\\
0.0109783017903434	0.472427635748996\\
0.0110836700259031	0.477427987886471\\
0.0112251475248307	0.482428340023945\\
0.0113866631730234	0.487499119656314\\
0.0115079983028556	0.492499471793788\\
0.0116551806856724	0.497499823931263\\
0.011816047459168	0.502570603563631\\
0.0119508520357658	0.507570955701106\\
0.0120781845170064	0.51257130783858\\
0.0122435980744024	0.517642087470949\\
0.0123802130177955	0.522642439608423\\
0.0125537930788676	0.527642791745898\\
0.0126849262295547	0.532713571378266\\
0.0128450453939595	0.537713923515741\\
0.0129998103408425	0.542714275653215\\
0.0131656412907595	0.547785055285584\\
0.0133166772063588	0.552785407423058\\
0.0135151841555727	0.557856187055426\\
0.0136488564233153	0.562856539192901\\
0.0138424977141221	0.567856891330375\\
0.0140706153501331	0.572927670962744\\
0.0142584412118675	0.577928023100218\\
0.0144521724596629	0.582928375237693\\
0.0146400983176268	0.587999154870061\\
0.0148582727473939	0.592999507007536\\
0.0150834971696262	0.59799985914501\\
0.0152779676147537	0.603070638777379\\
0.0154746714705188	0.608070990914853\\
0.0156720457642143	0.613071343052328\\
0.0158902128454304	0.618142122684696\\
0.0161252134270362	0.623142474822171\\
0.0163249991733853	0.628142826959645\\
0.0164949643515659	0.633213606592014\\
0.0167343648219298	0.638213958729488\\
0.0169630075282919	0.643284738361857\\
0.0172195580557941	0.648285090499331\\
0.0174827150426954	0.653285442636805\\
0.0177418510880497	0.658356222269174\\
0.0179607101262799	0.663356574406648\\
0.0182470421324901	0.668356926544123\\
0.0186059508094995	0.673427706176491\\
0.0188976611010981	0.678428058313966\\
0.0191439970162554	0.68342841045144\\
0.0194539689503548	0.688499190083809\\
0.0197980716758495	0.693499542221283\\
0.0201052984257146	0.698499894358758\\
0.0204640287204078	0.703570673991126\\
0.0207670110456948	0.708571026128601\\
0.0211271264056319	0.713571378266075\\
0.021585806753406	0.718642157898444\\
0.0219767487632558	0.723642510035918\\
0.0224408702987294	0.728713289668287\\
0.0228006882993451	0.733713641805761\\
0.0232607958991184	0.738713993943235\\
0.0236660669627781	0.743784773575604\\
0.0241485920497391	0.748785125713078\\
0.0246213850883928	0.753785477850553\\
0.0250599849386724	0.758856257482921\\
0.0255218110744287	0.763856609620396\\
0.0260470779443026	0.76885696175787\\
0.026606863191603	0.773927741390239\\
0.0272687079316373	0.778928093527713\\
0.0279018107536746	0.783928445665188\\
0.0286918813232336	0.788999225297556\\
0.0293992895548192	0.793999577435031\\
0.0300366249031802	0.798999929572505\\
0.0306537451428606	0.804070709204874\\
0.0313542218201084	0.809071061342348\\
0.0321088594015848	0.814071413479822\\
0.0331374979107237	0.819142193112191\\
0.0339407888617918	0.824142545249665\\
0.0347837096299849	0.829213324882034\\
0.035624745451186	0.834213677019508\\
0.036534034135629	0.839214029156983\\
0.0377272428203295	0.844284808789351\\
0.0390195675239035	0.849285160926826\\
0.04024059810314	0.8542855130643\\
0.0415050242481395	0.859356292696669\\
0.0428308760580831	0.864356644834143\\
0.0442977012217341	0.869356996971618\\
0.0461718435917369	0.874427776603986\\
0.0476013385482482	0.879428128741461\\
0.0493778790065301	0.884428480878935\\
0.0512540245580142	0.889499260511304\\
0.053454905898036	0.894499612648778\\
0.0560818198216468	0.899499964786253\\
0.0584974573670969	0.904570744418621\\
0.0613866628303115	0.909571096556095\\
0.0647726781050622	0.914641876188464\\
0.0689770169530205	0.919642228325938\\
0.0739285606857252	0.924642580463413\\
0.0814529673488007	0.929713360095781\\
0.0895765258657653	0.934713712233256\\
0.100772545219836	0.93971406437073\\
0.118016456311878	0.944784844003099\\
0.133451732921558	0.949785196140573\\
0.161877011876475	0.954785548278048\\
0.194910891464568	0.959856327910416\\
0.228527318096775	0.964856680047891\\
0.294321390303986	0.969857032185365\\
0.355387541694006	0.974927811817734\\
0.409584166407033	0.979928163955208\\
0.465526310442827	0.984928516092683\\
0.560374160269083	0.989999295725051\\
0.68433077856361	0.994999647862525\\
0.927599490697603	1\\
};

\addplot [color=green, line width=2.0pt]
  table[row sep=crcr]{%
0	0.30975\\
0.001	0.311166666666667\\
0.002	0.325083333333333\\
0.003	0.36025\\
0.004	0.413583333333333\\
0.005	0.470333333333333\\
0.006	0.516333333333333\\
0.007	0.551333333333333\\
0.008	0.576083333333333\\
0.009	0.604\\
0.01	0.6225\\
0.011	0.642916666666667\\
0.012	0.659833333333333\\
0.013	0.6725\\
0.014	0.682833333333333\\
0.015	0.6955\\
0.016	0.706583333333333\\
0.017	0.7175\\
0.018	0.726833333333333\\
0.019	0.736833333333333\\
0.02	0.745\\
0.021	0.753583333333333\\
0.022	0.761666666666666\\
0.023	0.769416666666667\\
0.024	0.77675\\
0.025	0.783166666666666\\
0.026	0.789583333333333\\
0.027	0.796166666666667\\
0.028	0.80375\\
0.029	0.810416666666667\\
0.03	0.816416666666667\\
0.031	0.822\\
0.032	0.829833333333333\\
0.033	0.838333333333333\\
0.034	0.846333333333333\\
0.035	0.853166666666667\\
0.036	0.859833333333333\\
0.037	0.866666666666667\\
0.038	0.872166666666667\\
0.039	0.87875\\
0.04	0.886333333333333\\
0.041	0.890916666666667\\
0.042	0.898333333333333\\
0.043	0.9055\\
0.044	0.912583333333333\\
0.045	0.918166666666667\\
0.046	0.924666666666667\\
0.047	0.930333333333333\\
0.048	0.937416666666667\\
0.049	0.9435\\
0.05	0.949583333333333\\
0.051	0.9545\\
0.052	0.960416666666667\\
0.053	0.966416666666667\\
0.054	0.971666666666667\\
0.055	0.9765\\
0.056	0.980166666666667\\
0.057	0.983166666666667\\
0.058	0.986\\
0.059	0.988166666666667\\
0.06	0.98975\\
0.061	0.991\\
0.062	0.993\\
0.063	0.994416666666667\\
0.064	0.996\\
0.065	0.996916666666667\\
0.066	0.998\\
0.067	0.999083333333333\\
0.068	0.9995\\
0.069	1\\
};

\addplot [color=mycolor2, line width=2.0pt]
  table[row sep=crcr]{%
0.000210260723296888	0.13782991202346\\
0.00273338940285955	0.140762463343109\\
0.00420521446593776	0.158357771260997\\
0.00504625735912532	0.180351906158358\\
0.00609756097560976	0.206744868035191\\
0.00693860386879731	0.233137829912023\\
0.00777964676198486	0.258064516129032\\
0.00862068965517241	0.284457478005865\\
0.00904121110176619	0.306451612903226\\
0.00967199327165686	0.331378299120235\\
0.0103027754415475	0.353372434017595\\
0.0107232968881413	0.372434017595308\\
0.011354079058032	0.395894428152493\\
0.0119848612279226	0.417888563049853\\
0.0128259041211102	0.441348973607038\\
0.0136669470142977	0.458944281524927\\
0.0149285113540791	0.482404692082111\\
0.0159798149705635	0.514662756598241\\
0.0170311185870479	0.533724340175953\\
0.0185029436501262	0.555718475073314\\
0.0197645079899075	0.582111436950147\\
0.0208158116063919	0.607038123167155\\
0.022497897392767	0.633431085043988\\
0.024390243902439	0.656891495601173\\
0.0256518082422204	0.67741935483871\\
0.0277544154751892	0.697947214076246\\
0.0294365012615643	0.718475073313783\\
0.0317493692178301	0.737536656891496\\
0.033851976450799	0.756598240469208\\
0.0359545836837679	0.777126099706745\\
0.0382674516400336	0.791788856304985\\
0.0401597981497056	0.803519061583578\\
0.043313708999159	0.818181818181818\\
0.0462573591253154	0.831378299120235\\
0.0492010092514718	0.844574780058651\\
0.052565180824222	0.856304985337243\\
0.0557190916736754	0.865102639296188\\
0.0590832632464256	0.875366568914956\\
0.0632884777123633	0.8841642228739\\
0.0674936921783011	0.89149560117302\\
0.0710681244743482	0.897360703812317\\
0.0765349032800673	0.903225806451613\\
0.0805298570227082	0.910557184750733\\
0.0841042893187553	0.912023460410557\\
0.088309503784693	0.917888563049853\\
0.0927249789739277	0.92375366568915\\
0.0952481076534903	0.926686217008798\\
0.0984020185029437	0.928152492668622\\
0.103027754415475	0.932551319648094\\
0.107022708158116	0.934017595307918\\
0.110597140454163	0.936950146627566\\
0.115222876366695	0.941348973607038\\
0.118797308662742	0.941348973607038\\
0.123212783851976	0.94574780058651\\
0.126787216148024	0.947214076246334\\
0.130992430613961	0.948680351906158\\
0.135618166526493	0.950146627565982\\
0.13919259882254	0.953079178885631\\
0.144028595458368	0.954545454545455\\
0.148023549201009	0.954545454545455\\
0.151597981497056	0.954545454545455\\
0.156433978132885	0.957478005865103\\
0.160218671152229	0.957478005865103\\
0.164634146341463	0.958944281524927\\
0.169680403700589	0.958944281524927\\
0.174095878889823	0.960410557184751\\
0.178511354079058	0.961876832844575\\
0.182926829268293	0.963343108504399\\
0.187552565180824	0.963343108504399\\
0.191757779646762	0.964809384164223\\
0.197224558452481	0.966275659824047\\
0.20206055508831	0.966275659824047\\
0.205845248107653	0.966275659824047\\
0.210050462573591	0.967741935483871\\
0.214255677039529	0.967741935483871\\
0.219091673675357	0.967741935483871\\
0.223717409587889	0.967741935483871\\
0.228132884777124	0.969208211143695\\
0.232127838519765	0.970674486803519\\
0.236543313708999	0.970674486803519\\
0.24053826745164	0.972140762463343\\
0.244112699747687	0.972140762463343\\
0.247476871320437	0.972140762463343\\
0.25	0.972140762463343\\
};

\addplot [color=mycolor3, line width=2.0pt]
  table[row sep=crcr]{%
0	0.00198063\\
0.0025	0.00891285\\
0.005	0.0230661\\
0.0075	0.044523\\
0.01	0.0717017\\
0.0125	0.105579\\
0.015	0.1435\\
0.0175	0.183071\\
0.02	0.225435\\
0.0225	0.269614\\
0.025	0.314219\\
0.0275	0.357408\\
0.03	0.401051\\
0.0325	0.442864\\
0.035	0.48095\\
0.0375	0.516863\\
0.04	0.55107\\
0.0425	0.583517\\
0.045	0.614065\\
0.0475	0.642372\\
0.05	0.66867\\
0.0525	0.692699\\
0.055	0.7145\\
0.0575	0.734691\\
0.06	0.753617\\
0.0625	0.770975\\
0.065	0.785239\\
0.0675	0.798471\\
0.07	0.810726\\
0.0725	0.820808\\
0.075	0.83111\\
0.0775	0.84016\\
0.08	0.84844\\
0.0825	0.85562\\
0.085	0.862387\\
0.0875	0.868522\\
0.09	0.874147\\
0.0925	0.878631\\
0.095	0.883156\\
0.0975	0.886925\\
0.1	0.890584\\
0.1025	0.8935\\
0.105	0.896443\\
0.1075	0.898616\\
0.11	0.900762\\
0.1125	0.902578\\
0.115	0.904256\\
0.1175	0.905356\\
0.12	0.906511\\
0.1225	0.907268\\
0.125	0.908231\\
0.1275	0.908863\\
0.13	0.909537\\
0.1325	0.910101\\
0.135	0.910665\\
0.1375	0.911229\\
0.14	0.911752\\
0.1425	0.912178\\
0.145	0.912577\\
0.1475	0.912962\\
0.15	0.913485\\
0.1525	0.913829\\
0.155	0.914338\\
0.1575	0.91475\\
0.16	0.915149\\
0.1625	0.915617\\
0.165	0.915851\\
0.1675	0.916236\\
0.17	0.916552\\
0.1725	0.916868\\
0.175	0.917102\\
0.1775	0.917432\\
0.18	0.917817\\
0.1825	0.91812\\
0.185	0.918588\\
0.1875	0.918932\\
0.19	0.919275\\
0.1925	0.919564\\
0.195	0.919853\\
0.1975	0.920183\\
0.2	0.920431\\
0.2025	0.920706\\
0.205	0.921146\\
0.2075	0.921421\\
0.21	0.921627\\
0.2125	0.92193\\
0.215	0.92226\\
0.2175	0.922521\\
0.22	0.922728\\
0.2225	0.922852\\
0.225	0.923113\\
0.2275	0.923319\\
0.23	0.923539\\
0.2325	0.923814\\
0.235	0.923952\\
0.2375	0.924158\\
0.24	0.924392\\
0.2425	0.924543\\
0.245	0.924805\\
0.2475	0.925025\\
};

\addplot [color=red, line width=2.0pt]
  table[row sep=crcr]{%
0	0\\
0	0.00507077963236848\\
0	0.0100711317698429\\
0	0.0151419114022114\\
0	0.0201422635396859\\
0	0.0251426156771604\\
0	0.0302133953095288\\
0	0.0352137474470033\\
0	0.0402140995844778\\
0	0.0452848792168463\\
0	0.0502852313543207\\
0	0.0552855834917952\\
0	0.0603563631241637\\
0	0.0653567152616381\\
0	0.0703570673991126\\
0	0.0754278470314811\\
0	0.0804281991689556\\
0	0.08542855130643\\
0	0.0904993309387985\\
0	0.095499683076273\\
0	0.100570462708641\\
0	0.105570814846116\\
0	0.11057116698359\\
0	0.115641946615959\\
0	0.120642298753433\\
0	0.125642650890908\\
0	0.130713430523276\\
0	0.135713782660751\\
0	0.140714134798225\\
0	0.145784914430594\\
0	0.150785266568068\\
0	0.155785618705543\\
0	0.160856398337911\\
0	0.165856750475386\\
0	0.17085710261286\\
0	0.175927882245229\\
0	0.180928234382703\\
0	0.185999014015071\\
0	0.190999366152546\\
0	0.19599971829002\\
0	0.201070497922389\\
0	0.206070850059863\\
0	0.211071202197338\\
0	0.216141981829706\\
0	0.221142333967181\\
0	0.226142686104655\\
0	0.231213465737024\\
0	0.236213817874498\\
0	0.241214170011973\\
0	0.246284949644341\\
0	0.251285301781816\\
0	0.25628565391929\\
0	0.261356433551659\\
0	0.266356785689133\\
0	0.271357137826608\\
0	0.276427917458976\\
0	0.28142826959645\\
0	0.286499049228819\\
0	0.291499401366293\\
0	0.296499753503768\\
0	0.301570533136136\\
0	0.306570885273611\\
0.00200608953785969	0.311571237411085\\
0.00241108909354326	0.316642017043454\\
0.00270437369574142	0.321642369180928\\
0.00292872996966721	0.326642721318403\\
0.00310840919156002	0.331713500950771\\
0.00327809497085294	0.336713853088246\\
0.00346504096747256	0.34171420522572\\
0.00359832823564924	0.346784984858089\\
0.00370629611727054	0.351785336995563\\
0.00382194732617591	0.356785689133038\\
0.00393624313363827	0.361856468765406\\
0.00404561253707746	0.36685682090288\\
0.0041664502626552	0.371927600535249\\
0.00425664112406345	0.376927952672723\\
0.00434474986385312	0.381928304810198\\
0.0044250680752411	0.386999084442566\\
0.00452732961104077	0.391999436580041\\
0.00460959715256545	0.396999788717515\\
0.00471050096368811	0.402070568349884\\
0.00479317309546186	0.407070920487358\\
0.00486744345804081	0.412071272624833\\
0.0049422149133985	0.417142052257201\\
0.00501579232823438	0.422142404394676\\
0.00510055222682774	0.42714275653215\\
0.00517069925504138	0.432213536164519\\
0.00524589428517953	0.437213888301993\\
0.00530970463474109	0.442214240439468\\
0.00537768272934552	0.447285020071836\\
0.00544149120788111	0.452285372209311\\
0.00550302814493663	0.457356151841679\\
0.00557113077730754	0.462356503979153\\
0.00563736402674931	0.467356856116628\\
0.0057085773654248	0.472427635748996\\
0.00576583814319398	0.477427987886471\\
0.00584848803690166	0.482428340023945\\
0.00591872226890902	0.487499119656314\\
0.00598650002101588	0.492499471793788\\
0.00604779293571024	0.497499823931263\\
0.0061153317570588	0.502570603563631\\
0.00618709738847438	0.507570955701106\\
0.00626744165893131	0.51257130783858\\
0.00631660690672202	0.517642087470949\\
0.00638186901195083	0.522642439608423\\
0.00644335106464575	0.527642791745898\\
0.00651143478709857	0.532713571378266\\
0.00656981442900583	0.537713923515741\\
0.0066298293979365	0.542714275653215\\
0.00667742872173103	0.547785055285584\\
0.00673227074065332	0.552785407423058\\
0.00679559752477567	0.557856187055426\\
0.00686062246856918	0.562856539192901\\
0.00691988433919169	0.567856891330375\\
0.00698302602614114	0.572927670962744\\
0.00705745100334679	0.577928023100218\\
0.0071370122728811	0.582928375237693\\
0.00722472759644451	0.587999154870061\\
0.00729840206212255	0.592999507007536\\
0.00736553529209054	0.59799985914501\\
0.00743092075071114	0.603070638777379\\
0.00750179312794913	0.608070990914853\\
0.00756900159632769	0.613071343052328\\
0.00762635576345098	0.618142122684696\\
0.00769947055935219	0.623142474822171\\
0.00778413827268808	0.628142826959645\\
0.007866689279833	0.633213606592014\\
0.00794859333197656	0.638213958729488\\
0.008037547142863	0.643284738361857\\
0.00811970212780504	0.648285090499331\\
0.0081950285085867	0.653285442636805\\
0.00827101285056874	0.658356222269174\\
0.00835100481200081	0.663356574406648\\
0.00843535783031259	0.668356926544123\\
0.00852366089065446	0.673427706176491\\
0.00860663727767799	0.678428058313966\\
0.00869599099217775	0.68342841045144\\
0.0087939206644357	0.688499190083809\\
0.00890371456933399	0.693499542221283\\
0.00900517217020155	0.698499894358758\\
0.00909856265149305	0.703570673991126\\
0.0092227730687015	0.708571026128601\\
0.0093210249479187	0.713571378266075\\
0.00942001086325694	0.718642157898444\\
0.00950989386464503	0.723642510035918\\
0.00963143044009792	0.728713289668287\\
0.00973497592171183	0.733713641805761\\
0.00984539898191737	0.738713993943235\\
0.00997645207980621	0.743784773575604\\
0.0100960749362022	0.748785125713078\\
0.0102321815920169	0.753785477850553\\
0.0103516605880529	0.758856257482921\\
0.0104950712513667	0.763856609620396\\
0.0106427719610886	0.76885696175787\\
0.0107648193231213	0.773927741390239\\
0.0108959020288514	0.778928093527713\\
0.0110174787526074	0.783928445665188\\
0.0111666019655976	0.788999225297556\\
0.011334716884958	0.793999577435031\\
0.0114824145066229	0.798999929572505\\
0.0116230651732315	0.804070709204874\\
0.0117838448252635	0.809071061342348\\
0.0119593392377293	0.814071413479822\\
0.0121193715214781	0.819142193112191\\
0.0123202172879223	0.824142545249665\\
0.0125136262565852	0.829213324882034\\
0.0126630423534093	0.834213677019508\\
0.0129004056583477	0.839214029156983\\
0.013096166449718	0.844284808789351\\
0.0133192039371941	0.849285160926826\\
0.0134962859711958	0.8542855130643\\
0.0137717775651233	0.859356292696669\\
0.0139784120684028	0.864356644834143\\
0.0141678758512708	0.869356996971618\\
0.0144723048519188	0.874427776603986\\
0.0147720357367932	0.879428128741461\\
0.0150333234288875	0.884428480878935\\
0.0153049310231163	0.889499260511304\\
0.0156758031103799	0.894499612648778\\
0.0159727762384006	0.899499964786253\\
0.0163347386799996	0.904570744418621\\
0.0166867481072937	0.909571096556095\\
0.0170536344404805	0.914641876188464\\
0.0174900110709725	0.919642228325938\\
0.0179848978290161	0.924642580463413\\
0.0185681826893765	0.929713360095781\\
0.0193660742139133	0.934713712233256\\
0.0200853492332318	0.93971406437073\\
0.0209549801174974	0.944784844003099\\
0.0219168828579865	0.949785196140573\\
0.0231430559138646	0.954785548278048\\
0.0244675703829844	0.959856327910416\\
0.0262632457435425	0.964856680047891\\
0.0282364117223046	0.969857032185365\\
0.0305610908399674	0.974927811817734\\
0.0340252740441992	0.979928163955208\\
0.039017683482495	0.984928516092683\\
0.0511914706108103	0.989999295725051\\
0.145522736819623	0.994999647862525\\
0.633441329928316	1\\
};

\end{axis}
\end{tikzpicture}%

%% file: figures/curve_topkids.tikz
\definecolor{mycolor1}{rgb}{0.00000,1.00000,1.00000}%
\definecolor{mycolor2}{rgb}{0.91000,0.41000,0.17000}%
\begin{tikzpicture}

\begin{axis}[%
width=0.18\linewidth,
height=0.18\linewidth,
at={(0,0)},
scale only axis,
xmin=0,
xmax=0.25,
xlabel style={font=\color{white!15!black}},
xlabel={Relative geodesic error},
every x tick label/.append style={font=\color{black}, font=\footnotesize},
every y tick label/.append style={font=\color{black}, font=\footnotesize},
x tick label style={/pgf/number format/fixed},
ymin=0,
ymax=1,
ylabel style={font=\color{white!15!black}},
axis background/.style={fill=white},
title style={font=\bfseries},
title={TOPKIDS},
xmajorgrids,
ymajorgrids
]

\addplot [color=mycolor1, line width=2.0pt]
  table[row sep=crcr]{%
0	0\\
0	0.00504455654570102\\
0	0.0100668903753417\\
0	0.0150892242049823\\
0.000600417257331474	0.020111558034623\\
0.00110358246358106	0.0251338918642636\\
0.00141295342695854	0.0301562256939043\\
0.00165824611491152	0.035178559523545\\
0.00186018129856323	0.040223116069246\\
0.00205659706410635	0.0452454498988866\\
0.0022484330050913	0.0502677837285273\\
0.00241677251047499	0.055290117558168\\
0.00258216455546061	0.0603124513878086\\
0.00273808879388135	0.0653347852174493\\
0.00290463040706081	0.0703571190470899\\
0.003049564362325	0.0753794528767306\\
0.00318550471000649	0.0804240094224316\\
0.00337747208097702	0.0854463432520723\\
0.00355045017754878	0.0904686770817129\\
0.00374497800399233	0.0954910109113536\\
0.0039156348966647	0.100513344740994\\
0.00407718228260316	0.105535678570635\\
0.00427882913087154	0.110558012400276\\
0.00447620545889577	0.115580346229916\\
0.00472603365413084	0.120624902775617\\
0.00497492269260954	0.125647236605258\\
0.00518453974374472	0.130669570434899\\
0.00546394653940088	0.135691904264539\\
0.00568082909271784	0.14071423809418\\
0.00594571349480666	0.145736571923821\\
0.00621626186012636	0.150758905753461\\
0.00648587789130183	0.155781239583102\\
0.00682025393559164	0.160825796128803\\
0.00711623296072449	0.165848129958444\\
0.00743766902077419	0.170870463788084\\
0.00771759096782903	0.175892797617725\\
0.00799762491694405	0.180915131447365\\
0.00831528411880421	0.185937465277006\\
0.00865481596081442	0.190959799106647\\
0.00893500340961423	0.195982132936287\\
0.00921184299229588	0.201026689481988\\
0.00953432868737428	0.206049023311629\\
0.00988569143369196	0.21107135714127\\
0.0102006739238168	0.21609369097091\\
0.0105587624213516	0.221116024800551\\
0.0109249144855209	0.226138358630192\\
0.0113022984589888	0.231160692459832\\
0.011616286715493	0.236183026289473\\
0.0119870649443086	0.241227582835174\\
0.0123479019380817	0.246249916664815\\
0.0127423213005791	0.251272250494455\\
0.0131524477784571	0.256294584324096\\
0.0135536938180014	0.261316918153737\\
0.0139794976984939	0.266339251983377\\
0.0143825496306854	0.271361585813018\\
0.0148396852647559	0.276383919642659\\
0.0153029117890085	0.28142847618836\\
0.0157117661380041	0.286450810018\\
0.0162450971156865	0.291473143847641\\
0.0166928078207774	0.296495477677282\\
0.0172259514563803	0.301517811506922\\
0.0177236421472685	0.306540145336563\\
0.0182518648185808	0.311562479166204\\
0.0187263967679639	0.316584812995844\\
0.0192358235254312	0.321629369541545\\
0.0197392883605259	0.326651703371186\\
0.0203203318527601	0.331674037200827\\
0.0208070775300141	0.336696371030467\\
0.0213457441113424	0.341718704860108\\
0.0218651754285782	0.346741038689749\\
0.0224108772043058	0.351763372519389\\
0.0230091705789374	0.35678570634903\\
0.023567823832902	0.361830262894731\\
0.024073441923394	0.366852596724372\\
0.0246259255610774	0.371874930554012\\
0.0253199277544011	0.376897264383653\\
0.0258827381084259	0.381919598213294\\
0.0265044717511553	0.386941932042934\\
0.0270983258162717	0.391964265872575\\
0.0278001681990073	0.396986599702216\\
0.0284927645541656	0.402031156247917\\
0.0291537821274666	0.407053490077557\\
0.0298599865062588	0.412075823907198\\
0.0305898597478716	0.417098157736839\\
0.0312160551867231	0.422120491566479\\
0.0320260118959074	0.42714282539612\\
0.0328260948524309	0.432165159225761\\
0.0335597099741397	0.437187493055401\\
0.0342806863118088	0.442232049601102\\
0.0351710788981216	0.447254383430743\\
0.0360748692663703	0.452276717260384\\
0.0368928024397674	0.457299051090024\\
0.037735689193965	0.462321384919665\\
0.0385953375735131	0.467343718749306\\
0.0393254064042087	0.472366052578946\\
0.0401676303172645	0.477388386408587\\
0.0409123662522923	0.482432942954288\\
0.04170275958342	0.487455276783929\\
0.0425658375595237	0.492477610613569\\
0.0434598231086957	0.49749994444321\\
0.0444168513392432	0.50252227827285\\
0.0452353274427327	0.507544612102491\\
0.0461512082714074	0.512566945932132\\
0.0470971316784963	0.517589279761772\\
0.0480108915431677	0.522633836307474\\
0.0489375329447866	0.527656170137114\\
0.0500067174441905	0.532678503966755\\
0.0509300061037939	0.537700837796395\\
0.0520229214450296	0.542723171626036\\
0.0530618998681484	0.547745505455677\\
0.054344664997426	0.552767839285317\\
0.0555484726237029	0.557790173114958\\
0.0568902808655253	0.562834729660659\\
0.058453603365063	0.5678570634903\\
0.0600464267026552	0.57287939731994\\
0.0619821133780421	0.577901731149581\\
0.0636500033185139	0.582924064979222\\
0.0653599756438916	0.587946398808862\\
0.0670135847763107	0.592968732638503\\
0.0687804560633735	0.597991066468144\\
0.0707420266685148	0.603035623013845\\
0.072594181340973	0.608057956843485\\
0.0745147234168565	0.613080290673126\\
0.0761998591785774	0.618102624502767\\
0.078206787679303	0.623124958332407\\
0.0801350772368946	0.628147292162048\\
0.0820185524022415	0.633169625991689\\
0.0840445215220814	0.638191959821329\\
0.0860856550682454	0.64323651636703\\
0.0881428171790429	0.648258850196671\\
0.0903665737772187	0.653281184026312\\
0.0927305419282935	0.658303517855952\\
0.0952763668653075	0.663325851685593\\
0.0981660846195967	0.668348185515234\\
0.101029964958472	0.673370519344874\\
0.103394571076726	0.678392853174515\\
0.105681644634099	0.683437409720216\\
0.108012426225218	0.688459743549857\\
0.110299396240836	0.693482077379497\\
0.112987281634683	0.698504411209138\\
0.115698621307332	0.703526745038779\\
0.118394741115547	0.708549078868419\\
0.121089028118644	0.71357141269806\\
0.123344279746201	0.718593746527701\\
0.125526448212364	0.723638303073402\\
0.12792668460552	0.728660636903042\\
0.130644151414233	0.733682970732683\\
0.133200847512672	0.738705304562324\\
0.135674266407654	0.743727638391964\\
0.13839749529289	0.748749972221605\\
0.14146923852598	0.753772306051246\\
0.144643430945751	0.758794639880886\\
0.14943757709954	0.763839196426587\\
0.154725229622294	0.768861530256228\\
0.161944275261742	0.773883864085869\\
0.170765640921978	0.778906197915509\\
0.180559674685599	0.78392853174515\\
0.190268478778685	0.788950865574791\\
0.199826922818613	0.793973199404431\\
0.209288548150825	0.798995533234072\\
0.217495130777522	0.804040089779773\\
0.227153463610706	0.809062423609413\\
0.235583136583078	0.814084757439054\\
0.242377579349943	0.819107091268695\\
0.248999090127015	0.824129425098336\\
0.257024147895969	0.829151758927976\\
0.262944960888239	0.834174092757617\\
0.269086136260495	0.839196426587257\\
0.276461399625275	0.844240983132958\\
0.283931854549405	0.849263316962599\\
0.291039402895331	0.85428565079224\\
0.29932372246075	0.859307984621881\\
0.307293009203395	0.864330318451521\\
0.316124996126419	0.869352652281162\\
0.323805471191993	0.874374986110802\\
0.33248273762656	0.879397319940443\\
0.340353325858188	0.884441876486144\\
0.349008286231655	0.889464210315785\\
0.358888209691668	0.894486544145425\\
0.367483534051467	0.899508877975066\\
0.376943659083626	0.904531211804707\\
0.383642323878799	0.909553545634347\\
0.389453793679186	0.914575879463988\\
0.396062268446337	0.919598213293629\\
0.404311864377254	0.92464276983933\\
0.414405087291897	0.92966510366897\\
0.423475061155387	0.934687437498611\\
0.431522491042987	0.939709771328252\\
0.44303056757341	0.944732105157892\\
0.464284044061622	0.949754438987533\\
0.48349513943226	0.954776772817174\\
0.500625439848374	0.959799106646814\\
0.522342130188727	0.964843663192515\\
0.577582970206559	0.969865997022156\\
0.624120208204588	0.974888330851797\\
0.638526886138042	0.979910664681437\\
0.664799683160945	0.984932998511078\\
0.773610675467537	0.989955332340719\\
0.79559430323666	0.994977666170359\\
0.809472304000399	1\\
};

\addplot [color=green, line width=2.0pt]
  table[row sep=crcr]{%
0	0.312591531285357\\
0.01	0.454756338590483\\
0.02	0.62343542771077\\
0.03	0.715036978862531\\
0.04	0.766024171894662\\
0.05	0.796623105863754\\
0.06	0.816241010041431\\
0.07	0.829318910750074\\
0.08	0.838300034176848\\
0.09	0.844867969989308\\
0.1	0.850171715726564\\
0.11	0.854640619704382\\
0.12	0.858790169041412\\
0.13	0.862374738246235\\
0.14	0.865873349228953\\
0.15	0.86910540845052\\
0.16	0.872198401453331\\
0.17	0.875276471314084\\
0.18	0.878182207596991\\
0.19	0.881233875703617\\
0.2	0.884184086864735\\
0.21	0.887116965592408\\
0.22	0.890114058891813\\
0.23	0.892790980082497\\
0.24	0.895614571822996\\
0.25	0.898492599814893\\
};

\addplot [color=mycolor2, line width=2.0pt]
  table[row sep=crcr]{%
0	0.101253362704985\\
0.01	0.140455175839875\\
0.02	0.223635355525081\\
0.03	0.303815749580386\\
0.04	0.372965102353228\\
0.05	0.430612099627044\\
0.06	0.479731524933252\\
0.07	0.520428249453636\\
0.08	0.55392027532148\\
0.09	0.58149352327709\\
0.1	0.604236849518351\\
0.11	0.622639907968202\\
0.12	0.638372234208708\\
0.13	0.651922017414349\\
0.14	0.664134371879144\\
0.15	0.675242317566194\\
0.16	0.684530869483707\\
0.17	0.693341994958915\\
0.18	0.701356317891122\\
0.19	0.708964769162009\\
0.2	0.716249301381613\\
0.21	0.723274623421685\\
0.22	0.730321758973901\\
0.23	0.737372853135066\\
0.24	0.744914854752655\\
0.25	0.752754485097991\\
};

\addplot [color=red, line width=2.0pt]
  table[row sep=crcr]{%
0	0.150977777777778\\
0.001	0.187644444444444\\
0.002	0.280355555555556\\
0.003	0.367488888888889\\
0.004	0.444022222222222\\
0.005	0.508666666666667\\
0.006	0.565555555555556\\
0.007	0.616222222222222\\
0.008	0.661222222222222\\
0.009	0.698622222222222\\
0.01	0.729822222222222\\
0.011	0.756511111111111\\
0.012	0.780622222222222\\
0.013	0.800155555555556\\
0.014	0.817666666666667\\
0.015	0.833688888888889\\
0.016	0.8466\\
0.017	0.859288888888889\\
0.018	0.870155555555556\\
0.019	0.880355555555556\\
0.02	0.889466666666667\\
0.021	0.897244444444444\\
0.022	0.903666666666667\\
0.023	0.909622222222222\\
0.024	0.915\\
0.025	0.9202\\
0.026	0.924355555555556\\
0.027	0.9286\\
0.028	0.932555555555556\\
0.029	0.935933333333333\\
0.03	0.939444444444444\\
0.031	0.942644444444445\\
0.032	0.945511111111111\\
0.033	0.948333333333333\\
0.034	0.950911111111111\\
0.035	0.952844444444444\\
0.036	0.954311111111111\\
0.037	0.955955555555555\\
0.038	0.957377777777778\\
0.039	0.959044444444445\\
0.04	0.961088888888889\\
0.041	0.962244444444444\\
0.042	0.963444444444444\\
0.043	0.964422222222222\\
0.044	0.965422222222222\\
0.045	0.966355555555556\\
0.046	0.9674\\
0.047	0.968355555555556\\
0.048	0.969111111111111\\
0.049	0.97\\
0.05	0.970866666666667\\
0.051	0.971666666666667\\
0.052	0.972355555555555\\
0.053	0.973022222222222\\
0.054	0.973844444444444\\
0.055	0.974644444444444\\
0.056	0.975333333333333\\
0.057	0.975822222222222\\
0.058	0.976377777777778\\
0.059	0.976844444444444\\
0.06	0.977222222222222\\
0.061	0.977955555555556\\
0.062	0.978377777777778\\
0.063	0.978822222222222\\
0.064	0.979288888888889\\
0.065	0.979688888888889\\
0.066	0.98\\
0.067	0.980266666666667\\
0.068	0.980777777777778\\
0.069	0.9812\\
0.07	0.981644444444444\\
0.071	0.981866666666667\\
0.072	0.982155555555556\\
0.073	0.982333333333333\\
0.074	0.982577777777778\\
0.075	0.982733333333333\\
0.076	0.982888888888889\\
0.077	0.983044444444444\\
0.078	0.983288888888889\\
0.079	0.983355555555556\\
0.08	0.983577777777778\\
0.081	0.983888888888889\\
0.082	0.984\\
0.083	0.984088888888889\\
0.084	0.984155555555556\\
0.085	0.984311111111111\\
0.086	0.984533333333333\\
0.087	0.984644444444444\\
0.088	0.984844444444444\\
0.089	0.985\\
0.09	0.985133333333333\\
0.091	0.985244444444444\\
0.092	0.985333333333333\\
0.093	0.985466666666667\\
0.094	0.985511111111111\\
0.095	0.985555555555556\\
0.096	0.985644444444444\\
0.097	0.985688888888889\\
0.098	0.985755555555556\\
0.099	0.985888888888889\\
0.1	0.985933333333333\\
0.101	0.985933333333333\\
0.102	0.985955555555555\\
0.103	0.985955555555555\\
0.104	0.985977777777778\\
0.105	0.985977777777778\\
0.106	0.985977777777778\\
0.107	0.986066666666667\\
0.108	0.986088888888889\\
0.109	0.986133333333333\\
0.11	0.986177777777778\\
0.111	0.9862\\
0.112	0.986244444444444\\
0.113	0.986244444444444\\
0.114	0.986266666666667\\
0.115	0.986288888888889\\
0.116	0.986288888888889\\
0.117	0.986311111111111\\
0.118	0.986311111111111\\
0.119	0.986333333333333\\
0.12	0.986333333333333\\
0.121	0.986422222222222\\
0.122	0.986466666666667\\
0.123	0.986511111111111\\
0.124	0.986511111111111\\
0.125	0.986533333333333\\
0.126	0.986533333333333\\
0.127	0.986533333333333\\
0.128	0.986533333333333\\
0.129	0.986533333333333\\
0.13	0.986555555555556\\
0.131	0.9866\\
0.132	0.986644444444444\\
0.133	0.986666666666667\\
0.134	0.986711111111111\\
0.135	0.986711111111111\\
0.136	0.986711111111111\\
0.137	0.986711111111111\\
0.138	0.986711111111111\\
0.139	0.986777777777778\\
0.14	0.9868\\
0.141	0.9868\\
0.142	0.986822222222222\\
0.143	0.986844444444444\\
0.144	0.986866666666667\\
0.145	0.986888888888889\\
0.146	0.986955555555556\\
0.147	0.987\\
0.148	0.987022222222222\\
0.149	0.987066666666667\\
0.15	0.987066666666667\\
0.151	0.987066666666667\\
0.152	0.987066666666667\\
0.153	0.987111111111111\\
0.154	0.987111111111111\\
0.155	0.987111111111111\\
0.156	0.987111111111111\\
0.157	0.987111111111111\\
0.158	0.987111111111111\\
0.159	0.987133333333333\\
0.16	0.987177777777778\\
0.161	0.987177777777778\\
0.162	0.987222222222222\\
0.163	0.987222222222222\\
0.164	0.987244444444444\\
0.165	0.987244444444444\\
0.166	0.987244444444444\\
0.167	0.987266666666667\\
0.168	0.987288888888889\\
0.169	0.987288888888889\\
0.17	0.987288888888889\\
0.171	0.987288888888889\\
0.172	0.987288888888889\\
0.173	0.987333333333333\\
0.174	0.987355555555555\\
0.175	0.987377777777778\\
0.176	0.987444444444444\\
0.177	0.987466666666667\\
0.178	0.987488888888889\\
0.179	0.987555555555556\\
0.18	0.987555555555556\\
0.181	0.987577777777778\\
0.182	0.987577777777778\\
0.183	0.987666666666667\\
0.184	0.987711111111111\\
0.185	0.987755555555556\\
0.186	0.987755555555556\\
0.187	0.9878\\
0.188	0.987844444444444\\
0.189	0.987955555555556\\
0.19	0.988133333333333\\
0.191	0.988355555555556\\
0.192	0.988488888888889\\
0.193	0.9886\\
0.194	0.988755555555556\\
0.195	0.988844444444444\\
0.196	0.988933333333333\\
0.197	0.989\\
0.198	0.989\\
0.199	0.989044444444445\\
0.2	0.989111111111111\\
0.201	0.9892\\
0.202	0.989266666666667\\
0.203	0.989311111111111\\
0.204	0.989311111111111\\
0.205	0.989333333333333\\
0.206	0.989355555555556\\
0.207	0.989377777777778\\
0.208	0.989422222222222\\
0.209	0.989488888888889\\
0.21	0.989533333333333\\
0.211	0.989533333333333\\
0.212	0.989622222222222\\
0.213	0.989666666666667\\
0.214	0.989711111111111\\
0.215	0.989733333333333\\
0.216	0.989755555555556\\
0.217	0.9898\\
0.218	0.9898\\
0.219	0.9898\\
0.22	0.989844444444444\\
0.221	0.989866666666667\\
0.222	0.989911111111111\\
0.223	0.989911111111111\\
0.224	0.989955555555556\\
0.225	0.99\\
0.226	0.990044444444444\\
0.227	0.990088888888889\\
0.228	0.990155555555556\\
0.229	0.990244444444444\\
0.23	0.990288888888889\\
0.231	0.990377777777778\\
0.232	0.9904\\
0.233	0.990444444444445\\
0.234	0.990466666666667\\
0.235	0.990488888888889\\
0.236	0.990488888888889\\
0.237	0.990488888888889\\
0.238	0.990533333333333\\
0.239	0.990533333333333\\
0.24	0.990555555555556\\
0.241	0.990577777777778\\
0.242	0.9906\\
0.243	0.990644444444444\\
0.244	0.990688888888889\\
0.245	0.990733333333333\\
0.246	0.990733333333333\\
0.247	0.990733333333333\\
0.248	0.990755555555556\\
0.249	0.990755555555556\\
0.25	0.990755555555556\\
};

\end{axis}
\end{tikzpicture}%

%% file: figures/conf_dist_SCAPE.tikz
\definecolor{mycolor1}{rgb}{0.00000,1.00000,1.00000}%
\begin{tikzpicture}

\begin{axis}[%
width=0.23\linewidth,
at={(0,0)},
scale only axis,
xmin=0,
xmax=2,
xlabel style={font=\color{white!15!black}},
xlabel={Conformal distortion factor},
ymin=0,
ymax=1,
ylabel style={font=\color{white!15!black}},
axis background/.style={fill=white},
title style={font=\bfseries},
title={SCAPE},
xmajorgrids,
ymajorgrids,
legend style={at={(0.55,0.02)}, anchor=south west, legend cell align=left, align=left, draw=white!15!black,font=\tiny}
]
\addplot [color=red, dashed, line width=2.0pt]
  table[row sep=crcr]{%
2.97526132619907e-08	0\\
0.000154658706660982	0.00502519357897566\\
0.000308669305284504	0.0100503871579513\\
0.0004664046382028	0.015075580736927\\
0.000621929958259493	0.0201007743159026\\
0.000781089898679088	0.0251259678948783\\
0.000942681920521604	0.0301511614738539\\
0.00110931472616382	0.0351763550528296\\
0.00127367842596771	0.0402015486318053\\
0.00144077556796507	0.0452261787851078\\
0.00161065351370393	0.0502513723640834\\
0.00178563395581754	0.0552765659430591\\
0.00196047620121176	0.0603017595220347\\
0.00213960438136285	0.0653269531010104\\
0.00232197391244204	0.070352146679986\\
0.00250360263204463	0.0753773402589617\\
0.00268655267429496	0.0804025338379374\\
0.00287501201043483	0.0854271639912399\\
0.00306707694661146	0.0904523575702155\\
0.00325985244871596	0.0954775511491912\\
0.00345558150493641	0.100502744728167\\
0.00365358138910876	0.105527938307142\\
0.00385070340224569	0.110553131886118\\
0.00405352446638618	0.115578325465094\\
0.00426004984578654	0.120603519044069\\
0.00446657394698446	0.125628149197372\\
0.00467758032260246	0.130653342776348\\
0.0048918619931051	0.135678536355323\\
0.00510794252011459	0.140703729934299\\
0.00532454667126769	0.145728923513275\\
0.00554999089100727	0.15075411709225\\
0.00578378788399148	0.155779310671226\\
0.00601568250083151	0.160804504250202\\
0.00625081612028433	0.165829697829177\\
0.00648612477079258	0.17085432798248\\
0.00673223828500236	0.175879521561455\\
0.00697293049688152	0.180904715140431\\
0.00722603080954354	0.185929908719407\\
0.00747383676648061	0.190955102298382\\
0.00773025462937227	0.195980295877358\\
0.00798830374027037	0.201005489456334\\
0.00824936641470497	0.206030683035309\\
0.00852283729411152	0.211055313188612\\
0.00879394962528002	0.216080506767587\\
0.00906449006357501	0.221105700346563\\
0.00933604475401673	0.226130893925539\\
0.00962020243713546	0.231156087504514\\
0.00991072285998928	0.23618128108349\\
0.0102052758170172	0.241206474662466\\
0.0105014955972393	0.246231668241441\\
0.0108026965325974	0.251256298394744\\
0.0111063642322971	0.25628149197372\\
0.0114193811810916	0.261306685552695\\
0.0117353084405027	0.266331879131671\\
0.0120592878790924	0.271357072710647\\
0.0123832020151546	0.276382266289622\\
0.0127113731679853	0.281407459868598\\
0.0130500977274517	0.286432653447574\\
0.0133859096780018	0.291457847026549\\
0.0137284431738354	0.296482477179852\\
0.0140811584235485	0.301507670758827\\
0.0144387402500645	0.306532864337803\\
0.0148070715536104	0.311558057916779\\
0.0151788315309775	0.316583251495754\\
0.0155530617928132	0.32160844507473\\
0.0159341565870492	0.326633638653706\\
0.0163140266724242	0.331658832232681\\
0.016705705856646	0.336683462385984\\
0.017098535210831	0.341708655964959\\
0.017506842860294	0.346733849543935\\
0.0179137664190332	0.351759043122911\\
0.0183320056674989	0.356784236701886\\
0.0187608096710541	0.361809430280862\\
0.0191983044658897	0.366834623859838\\
0.0196357381954817	0.371859817438813\\
0.0200824681668839	0.376884447592116\\
0.0205353849116259	0.381909641171092\\
0.0209989934716406	0.386934834750067\\
0.0214738321292676	0.391960028329043\\
0.0219498434899412	0.396985221908018\\
0.0224438257121116	0.402010415486994\\
0.0229464818266782	0.40703560906597\\
0.0234590342223835	0.412060802644945\\
0.0239843483193836	0.417085432798248\\
0.0245163497719814	0.422110626377224\\
0.0250536873795397	0.427135819956199\\
0.025600137703758	0.432161013535175\\
0.0261648646661086	0.437186207114151\\
0.026730375215986	0.442211400693126\\
0.027314350920506	0.447236594272102\\
0.0278952187711758	0.452261787851078\\
0.0284961570614386	0.457286981430053\\
0.0291044313258535	0.462311611583356\\
0.0297342148213633	0.467336805162331\\
0.0303777129380411	0.472361998741307\\
0.0310261829377674	0.477387192320283\\
0.0316854236486335	0.482412385899258\\
0.0323650637474868	0.487437579478234\\
0.0330603779463985	0.49246277305721\\
0.03375713083557	0.497487966636185\\
0.034457855819555	0.502512596789488\\
0.0351934154992604	0.507537790368463\\
0.0359377915933621	0.512562983947439\\
0.036699925611503	0.517588177526415\\
0.0374874666602083	0.52261337110539\\
0.0382869779544865	0.527638564684366\\
0.0390954411157312	0.532663758263342\\
0.0399270528139186	0.537688951842317\\
0.0407800399217058	0.54271358199562\\
0.0416457024622159	0.547738775574596\\
0.0425247267154356	0.552763969153571\\
0.0434195236115928	0.557789162732547\\
0.0443327150750461	0.562814356311523\\
0.0452623083030748	0.567839549890498\\
0.0462110302875653	0.572864743469474\\
0.0471823347311005	0.577889937048449\\
0.0481756918133973	0.582915130627425\\
0.0491933231050123	0.587939760780728\\
0.0502482388304988	0.592964954359703\\
0.0513174786611712	0.597990147938679\\
0.0524202378670267	0.603015341517655\\
0.0535224265449932	0.60804053509663\\
0.0546641932481662	0.613065728675606\\
0.0558575357951674	0.618090922254582\\
0.0570588880458511	0.623116115833557\\
0.0582996573581451	0.62814074598686\\
0.0595924063677584	0.633165939565835\\
0.0608717044640219	0.638191133144811\\
0.0622081816435744	0.643216326723787\\
0.0636100582370696	0.648241520302762\\
0.065022974117805	0.653266713881738\\
0.0664969038709105	0.658291907460714\\
0.0680228508787217	0.663317101039689\\
0.0695617809289901	0.668341731192992\\
0.071142518945122	0.673366924771968\\
0.072791385785886	0.678392118350943\\
0.074495782192463	0.683417311929919\\
0.0762380196917496	0.688442505508895\\
0.0780390543067928	0.69346769908787\\
0.0799141062124615	0.698492892666846\\
0.0818234545026453	0.703518086245821\\
0.0838179601732372	0.708542716399124\\
0.0858263567167783	0.7135679099781\\
0.0879806708982347	0.718593103557075\\
0.0901518537895778	0.723618297136051\\
0.0923954249955079	0.728643490715027\\
0.09474383036103	0.733668684294002\\
0.0971690680300887	0.738693877872978\\
0.0996641271243695	0.743719071451954\\
0.10233473778993	0.748744265030929\\
0.105059979093716	0.753768895184232\\
0.107885482009965	0.758794088763207\\
0.110777976585143	0.763819282342183\\
0.113836952496463	0.768844475921159\\
0.117022636869665	0.773869669500134\\
0.120352118276827	0.77889486307911\\
0.123778096158059	0.783920056658086\\
0.127367899974546	0.788945250237061\\
0.131156604629153	0.793969880390364\\
0.13509700353404	0.79899507396934\\
0.139253057419493	0.804020267548315\\
0.143579310879563	0.809045461127291\\
0.14811380548309	0.814070654706266\\
0.152891268604832	0.819095848285242\\
0.157952548909885	0.824121041864218\\
0.163255436876391	0.829146235443193\\
0.168861447713287	0.834170865596496\\
0.174734769625672	0.839196059175472\\
0.181135406086101	0.844221252754447\\
0.187835919602164	0.849246446333423\\
0.194920855177856	0.854271639912399\\
0.202611468930781	0.859296833491374\\
0.210900080984869	0.86432202707035\\
0.219672283700622	0.869347220649326\\
0.229221023589612	0.874372414228301\\
0.239449331375527	0.879397044381604\\
0.250471361661605	0.884422237960579\\
0.262493103485687	0.889447431539555\\
0.275553096543293	0.894472625118531\\
0.289789631047499	0.899497818697506\\
0.30534839539157	0.904523012276482\\
0.322781629851914	0.909548205855458\\
0.341894438940649	0.914573399434433\\
0.363109312444997	0.919598029587736\\
0.386906467346877	0.924623223166711\\
0.413463003013503	0.929648416745687\\
0.443145895238018	0.934673610324663\\
0.47909169187233	0.939698803903638\\
0.520198364527395	0.944723997482614\\
0.568799906878295	0.94974919106159\\
0.62707545516462	0.954774384640565\\
0.698753221282308	0.959799014793868\\
0.789148577832322	0.964824208372844\\
0.905532792902064	0.969849401951819\\
1.06175306962506	0.974874595530795\\
1.28325104196059	0.97989978910977\\
1.61897465148612	0.984924982688746\\
2	0.989950176267722\\
2	0.994975369846698\\
2	1\\
};
\addlegendentry{$\text{Ours}^*$}

\addplot [color=black, line width=2.0pt]
  table[row sep=crcr]{%
1.42168473882265e-07	0\\
0.00088841935888917	0.00502519357897566\\
0.00179735272871273	0.0100503871579513\\
0.00271701585154416	0.015075580736927\\
0.00367476416421475	0.0201007743159026\\
0.00465644918495345	0.0251259678948783\\
0.00563957293480799	0.0301511614738539\\
0.00663208070064858	0.0351763550528296\\
0.00763554418111889	0.0402015486318053\\
0.00866889251995318	0.0452261787851078\\
0.00971349634516816	0.0502513723640834\\
0.0107512174188775	0.0552765659430591\\
0.0118293519186565	0.0603017595220347\\
0.0129444053360701	0.0653269531010104\\
0.0140306797562335	0.070352146679986\\
0.0151450498426264	0.0753773402589617\\
0.0162568921652002	0.0804025338379374\\
0.0173720078421975	0.0854271639912399\\
0.0185153092759962	0.0904523575702155\\
0.0196711710868742	0.0954775511491912\\
0.0208192040421555	0.100502744728167\\
0.022024914457428	0.105527938307142\\
0.0232069495336025	0.110553131886118\\
0.0244061333236427	0.115578325465094\\
0.0256200584234225	0.120603519044069\\
0.0268577991900969	0.125628149197372\\
0.0280949698677753	0.130653342776348\\
0.0293589176725813	0.135678536355323\\
0.0305993880618778	0.140703729934299\\
0.0318940051812846	0.145728923513275\\
0.0331952044600117	0.15075411709225\\
0.03447831612212	0.155779310671226\\
0.0357991158949673	0.160804504250202\\
0.0371189216677381	0.165829697829177\\
0.0384436491257523	0.17085432798248\\
0.0397855864409538	0.175879521561455\\
0.0411327298597111	0.180904715140431\\
0.0425173338559084	0.185929908719407\\
0.0439138071860055	0.190955102298382\\
0.0453128973453194	0.195980295877358\\
0.0467321824467049	0.201005489456334\\
0.0481617996559751	0.206030683035309\\
0.0496045896711048	0.211055313188612\\
0.0510599769694955	0.216080506767587\\
0.0525470418432246	0.221105700346563\\
0.0540248695616774	0.226130893925539\\
0.0555498168644815	0.231156087504514\\
0.0570606009541312	0.23618128108349\\
0.0586127739419808	0.241206474662466\\
0.060159129687384	0.246231668241441\\
0.0617319254028876	0.251256298394744\\
0.0633083224678876	0.25628149197372\\
0.0649161195501389	0.261306685552695\\
0.0665378562767027	0.266331879131671\\
0.0681713427851864	0.271357072710647\\
0.0698317674229054	0.276382266289622\\
0.0714937547093726	0.281407459868598\\
0.0731753891112454	0.286432653447574\\
0.0749022614943544	0.291457847026549\\
0.0766249369080483	0.296482477179852\\
0.0783896336825114	0.301507670758827\\
0.0801542146615364	0.306532864337803\\
0.0819392005269148	0.311558057916779\\
0.0837501115412529	0.316583251495754\\
0.0856210941222115	0.32160844507473\\
0.0874858191081751	0.326633638653706\\
0.0893776209537909	0.331658832232681\\
0.0912640541514276	0.336683462385984\\
0.0931802040428931	0.341708655964959\\
0.0951067429939445	0.346733849543935\\
0.0970698630262756	0.351759043122911\\
0.0990379609058771	0.356784236701886\\
0.101041609531022	0.361809430280862\\
0.103086537173741	0.366834623859838\\
0.105143167870047	0.371859817438813\\
0.107260582946209	0.376884447592116\\
0.109365416209262	0.381909641171092\\
0.111510101310354	0.386934834750067\\
0.113676360477314	0.391960028329043\\
0.115883110645275	0.396985221908018\\
0.118092575654628	0.402010415486994\\
0.120375137016586	0.40703560906597\\
0.122651381753923	0.412060802644945\\
0.124960161236283	0.417085432798248\\
0.127344302459157	0.422110626377224\\
0.129751393636455	0.427135819956199\\
0.132154995189162	0.432161013535175\\
0.134622167090932	0.437186207114151\\
0.137073591091773	0.442211400693126\\
0.13953695649868	0.447236594272102\\
0.142005159367451	0.452261787851078\\
0.144550589309737	0.457286981430053\\
0.147200539751432	0.462311611583356\\
0.149867296918007	0.467336805162331\\
0.152545624325786	0.472361998741307\\
0.155274241511787	0.477387192320283\\
0.158025734662477	0.482412385899258\\
0.160865630179743	0.487437579478234\\
0.163721995429714	0.49246277305721\\
0.166662986526829	0.497487966636185\\
0.169617070645713	0.502512596789488\\
0.172657401446784	0.507537790368463\\
0.175707506243939	0.512562983947439\\
0.178794676654379	0.517588177526415\\
0.181955431552849	0.52261337110539\\
0.185159172476316	0.527638564684366\\
0.188370336890253	0.532663758263342\\
0.191668236090623	0.537688951842317\\
0.195046195628759	0.54271358199562\\
0.19849868378753	0.547738775574596\\
0.202032089980436	0.552763969153571\\
0.205559297947585	0.557789162732547\\
0.209144202748171	0.562814356311523\\
0.212744955131479	0.567839549890498\\
0.216424301907176	0.572864743469474\\
0.220203773013684	0.577889937048449\\
0.224057982262014	0.582915130627425\\
0.228071827896283	0.587939760780728\\
0.232038649723715	0.592964954359703\\
0.236143104060279	0.597990147938679\\
0.240407747893669	0.603015341517655\\
0.244712066510454	0.60804053509663\\
0.249205456923377	0.613065728675606\\
0.253643121466347	0.618090922254582\\
0.258135503943143	0.623116115833557\\
0.262876000528881	0.62814074598686\\
0.267594459246549	0.633165939565835\\
0.272443600412712	0.638191133144811\\
0.277403073671431	0.643216326723787\\
0.282507289737654	0.648241520302762\\
0.287808688605295	0.653266713881738\\
0.293183951694276	0.658291907460714\\
0.298618861398806	0.663317101039689\\
0.304307546677634	0.668341731192992\\
0.310045363925882	0.673366924771968\\
0.315976568114113	0.678392118350943\\
0.321980546032044	0.683417311929919\\
0.3281823031076	0.688442505508895\\
0.334369649028385	0.69346769908787\\
0.340997657449516	0.698492892666846\\
0.347583121789258	0.703518086245821\\
0.354538024061696	0.708542716399124\\
0.3615822144825	0.7135679099781\\
0.368674643576406	0.718593103557075\\
0.37621694753424	0.723618297136051\\
0.384039429622297	0.728643490715027\\
0.392099636889252	0.733668684294002\\
0.40044048886163	0.738693877872978\\
0.408921225227286	0.743719071451954\\
0.417589067818262	0.748744265030929\\
0.426612164497916	0.753768895184232\\
0.436122553987516	0.758794088763207\\
0.44580436102441	0.763819282342183\\
0.455935740953413	0.768844475921159\\
0.466114684452383	0.773869669500134\\
0.476796702698811	0.77889486307911\\
0.488066972250603	0.783920056658086\\
0.499857544231958	0.788945250237061\\
0.511967975567641	0.793969880390364\\
0.524502871231731	0.79899507396934\\
0.537860848101815	0.804020267548315\\
0.551595630769607	0.809045461127291\\
0.565860761635207	0.814070654706266\\
0.580510870508239	0.819095848285242\\
0.596167391499635	0.824121041864218\\
0.613118212708877	0.829146235443193\\
0.630453960081148	0.834170865596496\\
0.648483079452274	0.839196059175472\\
0.667773940093485	0.844221252754447\\
0.68792568812631	0.849246446333423\\
0.709723668450631	0.854271639912399\\
0.732734207503936	0.859296833491374\\
0.756978451043424	0.86432202707035\\
0.782737807563875	0.869347220649326\\
0.810303583708579	0.874372414228301\\
0.840404752783118	0.879397044381604\\
0.871933610244265	0.884422237960579\\
0.906209501610764	0.889447431539555\\
0.943429639138571	0.894472625118531\\
0.983659492910806	0.899497818697506\\
1.02873017733013	0.904523012276482\\
1.07727525640767	0.909548205855458\\
1.13051310418144	0.914573399434433\\
1.18893865286665	0.919598029587736\\
1.25404833969404	0.924623223166711\\
1.32832186012357	0.929648416745687\\
1.41279840877867	0.934673610324663\\
1.50991356619971	0.939698803903638\\
1.62072988387555	0.944723997482614\\
1.75009837174705	0.94974919106159\\
1.90745787499825	0.954774384640565\\
2	0.959799014793868\\
2	0.964824208372844\\
2	0.969849401951819\\
2	0.974874595530795\\
2	0.97989978910977\\
2	0.984924982688746\\
2	0.989950176267722\\
2	0.994975369846698\\
2	1\\
};
\addlegendentry{GT \cite{anguelov2005scape}}

\addplot [color=red, line width=2.0pt]
  table[row sep=crcr]{%
2.34232468976359e-08	0\\
0.00152237050632076	0.00502519357897566\\
0.00316692790909778	0.0100503871579513\\
0.00488106788067055	0.015075580736927\\
0.00657797438375418	0.0201007743159026\\
0.0083289456229072	0.0251259678948783\\
0.0101571838236678	0.0301511614738539\\
0.0120193984067525	0.0351763550528296\\
0.0139088461151204	0.0402015486318053\\
0.0158857579341429	0.0452261787851078\\
0.0178878482807785	0.0502513723640834\\
0.0199302563214401	0.0552765659430591\\
0.0219892189342925	0.0603017595220347\\
0.0241078315106051	0.0653269531010104\\
0.0262481164486212	0.070352146679986\\
0.0284422653204737	0.0753773402589617\\
0.0306911810575166	0.0804025338379374\\
0.0329418340799812	0.0854271639912399\\
0.035285725840478	0.0904523575702155\\
0.0375991654821042	0.0954775511491912\\
0.0399478795150028	0.100502744728167\\
0.0423173473846825	0.105527938307142\\
0.044742185071565	0.110553131886118\\
0.0472449923280385	0.115578325465094\\
0.0497493119862544	0.120603519044069\\
0.0523278473760973	0.125628149197372\\
0.0548867418787902	0.130653342776348\\
0.057490081263166	0.135678536355323\\
0.0601683771032624	0.140703729934299\\
0.0628427488598686	0.145728923513275\\
0.0655616219222916	0.15075411709225\\
0.0683415068841069	0.155779310671226\\
0.0711332169310288	0.160804504250202\\
0.0739813453854881	0.165829697829177\\
0.0768779836821873	0.17085432798248\\
0.079806752472988	0.175879521561455\\
0.0828021730467596	0.180904715140431\\
0.0858354385354714	0.185929908719407\\
0.0889596029119053	0.190955102298382\\
0.0920456246286174	0.195980295877358\\
0.0951603015192912	0.201005489456334\\
0.0983192330607867	0.206030683035309\\
0.101557359534515	0.211055313188612\\
0.104917446430719	0.216080506767587\\
0.108288443709094	0.221105700346563\\
0.111677589087753	0.226130893925539\\
0.115132117613569	0.231156087504514\\
0.118646753384545	0.23618128108349\\
0.122217328629557	0.241206474662466\\
0.12582761367447	0.246231668241441\\
0.129524272740867	0.251256298394744\\
0.133344338161276	0.25628149197372\\
0.137148990505639	0.261306685552695\\
0.140990150619809	0.266331879131671\\
0.144917907996398	0.271357072710647\\
0.148831173601174	0.276382266289622\\
0.152865701231362	0.281407459868598\\
0.156976802369651	0.286432653447574\\
0.161143713176706	0.291457847026549\\
0.165401677692933	0.296482477179852\\
0.169721055092	0.301507670758827\\
0.174046959919647	0.306532864337803\\
0.178588200863375	0.311558057916779\\
0.183125948822101	0.316583251495754\\
0.187715622722971	0.32160844507473\\
0.192395096422664	0.326633638653706\\
0.1972642162447	0.331658832232681\\
0.202137173219722	0.336683462385984\\
0.206976422577569	0.341708655964959\\
0.211984975901994	0.346733849543935\\
0.217171631267541	0.351759043122911\\
0.222410205650831	0.356784236701886\\
0.227786914391831	0.361809430280862\\
0.233161977708552	0.366834623859838\\
0.238696310914174	0.371859817438813\\
0.244254053594781	0.376884447592116\\
0.249975606273177	0.381909641171092\\
0.25566605559302	0.386934834750067\\
0.261668860174568	0.391960028329043\\
0.267715095773241	0.396985221908018\\
0.273839019058415	0.402010415486994\\
0.28014644083705	0.40703560906597\\
0.286574189280157	0.412060802644945\\
0.293353023368188	0.417085432798248\\
0.299960810217898	0.422110626377224\\
0.306625504090392	0.427135819956199\\
0.313639938453409	0.432161013535175\\
0.320866467692349	0.437186207114151\\
0.328168698114875	0.442211400693126\\
0.335661011043975	0.447236594272102\\
0.343460472156814	0.452261787851078\\
0.351276937130407	0.457286981430053\\
0.359232635309187	0.462311611583356\\
0.367357841522788	0.467336805162331\\
0.375721960810961	0.472361998741307\\
0.384114537600064	0.477387192320283\\
0.392701813116813	0.482412385899258\\
0.401610461780972	0.487437579478234\\
0.410643470407433	0.49246277305721\\
0.420047875063543	0.497487966636185\\
0.429571725491291	0.502512596789488\\
0.439500738789774	0.507537790368463\\
0.449398917323558	0.512562983947439\\
0.459534674143378	0.517588177526415\\
0.469911593835981	0.52261337110539\\
0.480570660021387	0.527638564684366\\
0.491429081266176	0.532663758263342\\
0.502830572566602	0.537688951842317\\
0.514301171043706	0.54271358199562\\
0.526182956019169	0.547738775574596\\
0.53867506017933	0.552763969153571\\
0.551522514310863	0.557789162732547\\
0.564657471232951	0.562814356311523\\
0.577925779350736	0.567839549890498\\
0.591716021318927	0.572864743469474\\
0.606294415656235	0.577889937048449\\
0.621176244738096	0.582915130627425\\
0.636217960669606	0.587939760780728\\
0.651969156801434	0.592964954359703\\
0.668435061998184	0.597990147938679\\
0.685383937513556	0.603015341517655\\
0.702904857405039	0.60804053509663\\
0.72138493256235	0.613065728675606\\
0.74005187370901	0.618090922254582\\
0.759628778864088	0.623116115833557\\
0.779969808332999	0.62814074598686\\
0.801249530147475	0.633165939565835\\
0.823778962491137	0.638191133144811\\
0.846933782536977	0.643216326723787\\
0.871030447087799	0.648241520302762\\
0.896457867186649	0.653266713881738\\
0.923113843542821	0.658291907460714\\
0.950717423051501	0.663317101039689\\
0.980284923801191	0.668341731192992\\
1.01116583121418	0.673366924771968\\
1.04391635731092	0.678392118350943\\
1.07886690422895	0.683417311929919\\
1.11557236509785	0.688442505508895\\
1.15449981851341	0.69346769908787\\
1.19633185175049	0.698492892666846\\
1.24136776605837	0.703518086245821\\
1.29012840021365	0.708542716399124\\
1.34273140981964	0.7135679099781\\
1.3998079900941	0.718593103557075\\
1.46127062241622	0.723618297136051\\
1.52902734302029	0.728643490715027\\
1.60415786086142	0.733668684294002\\
1.68785793197386	0.738693877872978\\
1.78410077298615	0.743719071451954\\
1.8914709499676	0.748744265030929\\
2	0.753768895184232\\
2	0.758794088763207\\
2	0.763819282342183\\
2	0.768844475921159\\
2	0.773869669500134\\
2	0.77889486307911\\
2	0.783920056658086\\
2	0.788945250237061\\
2	0.793969880390364\\
2	0.79899507396934\\
2	0.804020267548315\\
2	0.809045461127291\\
2	0.814070654706266\\
2	0.819095848285242\\
2	0.824121041864218\\
2	0.829146235443193\\
2	0.834170865596496\\
2	0.839196059175472\\
2	0.844221252754447\\
2	0.849246446333423\\
2	0.854271639912399\\
2	0.859296833491374\\
2	0.86432202707035\\
2	0.869347220649326\\
2	0.874372414228301\\
2	0.879397044381604\\
2	0.884422237960579\\
2	0.889447431539555\\
2	0.894472625118531\\
2	0.899497818697506\\
2	0.904523012276482\\
2	0.909548205855458\\
2	0.914573399434433\\
2	0.919598029587736\\
2	0.924623223166711\\
2	0.929648416745687\\
2	0.934673610324663\\
2	0.939698803903638\\
2	0.944723997482614\\
2	0.94974919106159\\
2	0.954774384640565\\
2	0.959799014793868\\
2	0.964824208372844\\
2	0.969849401951819\\
2	0.974874595530795\\
2	0.97989978910977\\
2	0.984924982688746\\
2	0.989950176267722\\
2	0.994975369846698\\
2	1\\
};
\addlegendentry{Ours}

\addplot [color=green, line width=2.0pt]
  table[row sep=crcr]{%
7.72588113040484e-07	0\\
0.00251173475586652	0.00502519357897566\\
0.00537607170326604	0.0100503871579513\\
0.00833966336119563	0.015075580736927\\
0.011516355242029	0.0201007743159026\\
0.0148183953936716	0.0251259678948783\\
0.0182207101276535	0.0301511614738539\\
0.0217335736312569	0.0351763550528296\\
0.0253382330304111	0.0402015486318053\\
0.029019237721255	0.0452261787851078\\
0.0328356464287118	0.0502513723640834\\
0.0366327239174136	0.0552765659430591\\
0.0405252036135781	0.0603017595220347\\
0.0444956028506276	0.0653269531010104\\
0.0485805483983803	0.070352146679986\\
0.0527554809758652	0.0753773402589617\\
0.056967623535995	0.0804025338379374\\
0.0612598437820018	0.0854271639912399\\
0.0656531864183032	0.0904523575702155\\
0.0700312944890009	0.0954775511491912\\
0.0744385091819373	0.100502744728167\\
0.0789978533046169	0.105527938307142\\
0.0836731613829169	0.110553131886118\\
0.0882448501303821	0.115578325465094\\
0.0930599919898425	0.120603519044069\\
0.0978957819238402	0.125628149197372\\
0.102747442685683	0.130653342776348\\
0.107732549353568	0.135678536355323\\
0.112842415344683	0.140703729934299\\
0.11804133384504	0.145728923513275\\
0.123289355113368	0.15075411709225\\
0.128591973932095	0.155779310671226\\
0.133976668880442	0.160804504250202\\
0.139257375058078	0.165829697829177\\
0.144831068836242	0.17085432798248\\
0.150347611144423	0.175879521561455\\
0.156054608817702	0.180904715140431\\
0.161780555719041	0.185929908719407\\
0.167661276024943	0.190955102298382\\
0.173569276855986	0.195980295877358\\
0.179528290966377	0.201005489456334\\
0.185592386493253	0.206030683035309\\
0.191723952722669	0.211055313188612\\
0.197981826031702	0.216080506767587\\
0.204278074775512	0.221105700346563\\
0.210657751248662	0.226130893925539\\
0.217201783097674	0.231156087504514\\
0.223792601187345	0.23618128108349\\
0.230528885644642	0.241206474662466\\
0.237259716797393	0.246231668241441\\
0.244301373410599	0.251256298394744\\
0.251433831218848	0.25628149197372\\
0.258563014978101	0.261306685552695\\
0.265690977978673	0.266331879131671\\
0.272939993455978	0.271357072710647\\
0.280255897133778	0.276382266289622\\
0.287741338507872	0.281407459868598\\
0.29530372038404	0.286432653447574\\
0.302944999853369	0.291457847026549\\
0.310630001281529	0.296482477179852\\
0.318479362301723	0.301507670758827\\
0.326496399020036	0.306532864337803\\
0.334465187390428	0.311558057916779\\
0.342683735254269	0.316583251495754\\
0.350941256953512	0.32160844507473\\
0.35934427519272	0.326633638653706\\
0.367984790132191	0.331658832232681\\
0.376682607657381	0.336683462385984\\
0.385590609943236	0.341708655964959\\
0.394566289949735	0.346733849543935\\
0.403441706499156	0.351759043122911\\
0.412429112756166	0.356784236701886\\
0.421666950143595	0.361809430280862\\
0.431073523725512	0.366834623859838\\
0.440803664266372	0.371859817438813\\
0.450723886058473	0.376884447592116\\
0.46056848248412	0.381909641171092\\
0.470723679489598	0.386934834750067\\
0.480831794206625	0.391960028329043\\
0.491200763610355	0.396985221908018\\
0.501573116052919	0.402010415486994\\
0.512097155664269	0.40703560906597\\
0.522925200832027	0.412060802644945\\
0.533710004593658	0.417085432798248\\
0.544795041593845	0.422110626377224\\
0.556330328791372	0.427135819956199\\
0.567652904844901	0.432161013535175\\
0.579267624641679	0.437186207114151\\
0.591121994956996	0.442211400693126\\
0.603259338847562	0.447236594272102\\
0.615301295736233	0.452261787851078\\
0.627661899416366	0.457286981430053\\
0.640187322307408	0.462311611583356\\
0.652944449045105	0.467336805162331\\
0.665820048576238	0.472361998741307\\
0.67902725824859	0.477387192320283\\
0.692599587970207	0.482412385899258\\
0.70624800837526	0.487437579478234\\
0.720142452738018	0.49246277305721\\
0.734160130428492	0.497487966636185\\
0.748602102070996	0.502512596789488\\
0.763307385553582	0.507537790368463\\
0.777927660460971	0.512562983947439\\
0.793031799522392	0.517588177526415\\
0.808425789531426	0.52261337110539\\
0.824017224747171	0.527638564684366\\
0.839960691329521	0.532663758263342\\
0.856509850560557	0.537688951842317\\
0.873066492941601	0.54271358199562\\
0.889808994430525	0.547738775574596\\
0.907328687218105	0.552763969153571\\
0.925077195660015	0.557789162732547\\
0.942747372698256	0.562814356311523\\
0.961456275229049	0.567839549890498\\
0.980669734934184	0.572864743469474\\
0.999472812169485	0.577889937048449\\
1.01888666857345	0.582915130627425\\
1.03812488598579	0.587939760780728\\
1.05861970155074	0.592964954359703\\
1.07953991292091	0.597990147938679\\
1.10046849986628	0.603015341517655\\
1.12254475499562	0.60804053509663\\
1.14485452729282	0.613065728675606\\
1.16749415975623	0.618090922254582\\
1.1906788560791	0.623116115833557\\
1.21443953484625	0.62814074598686\\
1.23849701935691	0.633165939565835\\
1.26348434049841	0.638191133144811\\
1.28905351687249	0.643216326723787\\
1.31540020212605	0.648241520302762\\
1.3426000604633	0.653266713881738\\
1.36974119437403	0.658291907460714\\
1.39816921093693	0.663317101039689\\
1.42744633747815	0.668341731192992\\
1.45740181365293	0.673366924771968\\
1.48796182595101	0.678392118350943\\
1.51920322115801	0.683417311929919\\
1.55188064183529	0.688442505508895\\
1.58587145029765	0.69346769908787\\
1.62064733663781	0.698492892666846\\
1.65626171476311	0.703518086245821\\
1.69325722044396	0.708542716399124\\
1.73130604502614	0.7135679099781\\
1.7697492116343	0.718593103557075\\
1.81046858163543	0.723618297136051\\
1.85181443933574	0.728643490715027\\
1.89521974853913	0.733668684294002\\
1.93996582852869	0.738693877872978\\
1.98650519563997	0.743719071451954\\
2	0.748744265030929\\
2	0.753768895184232\\
2	0.758794088763207\\
2	0.763819282342183\\
2	0.768844475921159\\
2	0.773869669500134\\
2	0.77889486307911\\
2	0.783920056658086\\
2	0.788945250237061\\
2	0.793969880390364\\
2	0.79899507396934\\
2	0.804020267548315\\
2	0.809045461127291\\
2	0.814070654706266\\
2	0.819095848285242\\
2	0.824121041864218\\
2	0.829146235443193\\
2	0.834170865596496\\
2	0.839196059175472\\
2	0.844221252754447\\
2	0.849246446333423\\
2	0.854271639912399\\
2	0.859296833491374\\
2	0.86432202707035\\
2	0.869347220649326\\
2	0.874372414228301\\
2	0.879397044381604\\
2	0.884422237960579\\
2	0.889447431539555\\
2	0.894472625118531\\
2	0.899497818697506\\
2	0.904523012276482\\
2	0.909548205855458\\
2	0.914573399434433\\
2	0.919598029587736\\
2	0.924623223166711\\
2	0.929648416745687\\
2	0.934673610324663\\
2	0.939698803903638\\
2	0.944723997482614\\
2	0.94974919106159\\
2	0.954774384640565\\
2	0.959799014793868\\
2	0.964824208372844\\
2	0.969849401951819\\
2	0.974874595530795\\
2	0.97989978910977\\
2	0.984924982688746\\
2	0.989950176267722\\
2	0.994975369846698\\
2	1\\
};
\addlegendentry{KM \cite{kernel17}}

\addplot [color=blue, line width=2.0pt]
  table[row sep=crcr]{%
7.2816250984431e-07	0\\
0.00449537608039385	0.00502519357897566\\
0.00904672884356339	0.0100503871579513\\
0.0137730054282743	0.015075580736927\\
0.0185957364938218	0.0201007743159026\\
0.0234524797948046	0.0251259678948783\\
0.0283932075113493	0.0301511614738539\\
0.0334309098357739	0.0351763550528296\\
0.03850458627717	0.0402015486318053\\
0.0435800042814192	0.0452261787851078\\
0.048750444096632	0.0502513723640834\\
0.0540727951809545	0.0552765659430591\\
0.0595154410634171	0.0603017595220347\\
0.0649908328771147	0.0653269531010104\\
0.0704365780188416	0.070352146679986\\
0.0759999657080175	0.0753773402589617\\
0.0817164578187533	0.0804025338379374\\
0.0875030794161424	0.0854271639912399\\
0.093404248665613	0.0904523575702155\\
0.0993306739105959	0.0954775511491912\\
0.105400317899351	0.100502744728167\\
0.111566436024306	0.105527938307142\\
0.117706901164139	0.110553131886118\\
0.123885546192896	0.115578325465094\\
0.130241049226916	0.120603519044069\\
0.136729184328444	0.125628149197372\\
0.143282918304302	0.130653342776348\\
0.149995393979341	0.135678536355323\\
0.156819821086168	0.140703729934299\\
0.163762937791222	0.145728923513275\\
0.170692229515793	0.15075411709225\\
0.177736619004897	0.155779310671226\\
0.184833934778253	0.160804504250202\\
0.192022158166121	0.165829697829177\\
0.199416842259027	0.17085432798248\\
0.206783084834245	0.175879521561455\\
0.214425462373063	0.180904715140431\\
0.222011285606865	0.185929908719407\\
0.229883745846726	0.190955102298382\\
0.237882058890166	0.195980295877358\\
0.245951255072878	0.201005489456334\\
0.253990355498185	0.206030683035309\\
0.26221410164124	0.211055313188612\\
0.270376670883261	0.216080506767587\\
0.278970522644582	0.221105700346563\\
0.287461493935035	0.226130893925539\\
0.296045628706608	0.231156087504514\\
0.304816412013952	0.23618128108349\\
0.313877831893189	0.241206474662466\\
0.322919508661032	0.246231668241441\\
0.332182182155358	0.251256298394744\\
0.341826600655663	0.25628149197372\\
0.351363474136755	0.261306685552695\\
0.361078828320004	0.266331879131671\\
0.370902217090083	0.271357072710647\\
0.380833210023213	0.276382266289622\\
0.39084749923239	0.281407459868598\\
0.4010719694157	0.286432653447574\\
0.411355899243234	0.291457847026549\\
0.421883250368408	0.296482477179852\\
0.432386516603366	0.301507670758827\\
0.443137452569996	0.306532864337803\\
0.454069184373115	0.311558057916779\\
0.465210379131151	0.316583251495754\\
0.476767645562798	0.32160844507473\\
0.488369821799155	0.326633638653706\\
0.500215083096723	0.331658832232681\\
0.512053842966895	0.336683462385984\\
0.524470529971692	0.341708655964959\\
0.537202178355632	0.346733849543935\\
0.549746772297252	0.351759043122911\\
0.56255347358168	0.356784236701886\\
0.57585893076084	0.361809430280862\\
0.589471882108938	0.366834623859838\\
0.603206130079176	0.371859817438813\\
0.617030083460091	0.376884447592116\\
0.631352847584568	0.381909641171092\\
0.645819211971242	0.386934834750067\\
0.66073538670529	0.391960028329043\\
0.675578949436146	0.396985221908018\\
0.691013128269325	0.402010415486994\\
0.706508071385191	0.40703560906597\\
0.722452044591731	0.412060802644945\\
0.738779518791295	0.417085432798248\\
0.755346337436619	0.422110626377224\\
0.772613079174562	0.427135819956199\\
0.789873992200286	0.432161013535175\\
0.807416807214834	0.437186207114151\\
0.825542682592045	0.442211400693126\\
0.843974101539011	0.447236594272102\\
0.863107407487595	0.452261787851078\\
0.882330355005208	0.457286981430053\\
0.902150689014809	0.462311611583356\\
0.922349932437092	0.467336805162331\\
0.942692983125694	0.472361998741307\\
0.964057431796104	0.477387192320283\\
0.985851673385147	0.482412385899258\\
1.00824246464683	0.487437579478234\\
1.03092193624587	0.49246277305721\\
1.05456763306956	0.497487966636185\\
1.07849067310938	0.502512596789488\\
1.10291782914623	0.507537790368463\\
1.1284401868403	0.512562983947439\\
1.15439281604553	0.517588177526415\\
1.18104459258313	0.52261337110539\\
1.20875608010643	0.527638564684366\\
1.23744570977412	0.532663758263342\\
1.26647279821323	0.537688951842317\\
1.29619474272699	0.54271358199562\\
1.32730323126673	0.547738775574596\\
1.35943181031486	0.552763969153571\\
1.39183189228846	0.557789162732547\\
1.4256275309282	0.562814356311523\\
1.46023892213238	0.567839549890498\\
1.49665928473668	0.572864743469474\\
1.53312936888953	0.577889937048449\\
1.57145079150046	0.582915130627425\\
1.61125610458938	0.587939760780728\\
1.65199308981537	0.592964954359703\\
1.69503290290532	0.597990147938679\\
1.73976153840452	0.603015341517655\\
1.78567014552739	0.60804053509663\\
1.83418247228921	0.613065728675606\\
1.88245521852329	0.618090922254582\\
1.93415665017339	0.623116115833557\\
1.98689891238235	0.62814074598686\\
2	0.633165939565835\\
2	0.638191133144811\\
2	0.643216326723787\\
2	0.648241520302762\\
2	0.653266713881738\\
2	0.658291907460714\\
2	0.663317101039689\\
2	0.668341731192992\\
2	0.673366924771968\\
2	0.678392118350943\\
2	0.683417311929919\\
2	0.688442505508895\\
2	0.69346769908787\\
2	0.698492892666846\\
2	0.703518086245821\\
2	0.708542716399124\\
2	0.7135679099781\\
2	0.718593103557075\\
2	0.723618297136051\\
2	0.728643490715027\\
2	0.733668684294002\\
2	0.738693877872978\\
2	0.743719071451954\\
2	0.748744265030929\\
2	0.753768895184232\\
2	0.758794088763207\\
2	0.763819282342183\\
2	0.768844475921159\\
2	0.773869669500134\\
2	0.77889486307911\\
2	0.783920056658086\\
2	0.788945250237061\\
2	0.793969880390364\\
2	0.79899507396934\\
2	0.804020267548315\\
2	0.809045461127291\\
2	0.814070654706266\\
2	0.819095848285242\\
2	0.824121041864218\\
2	0.829146235443193\\
2	0.834170865596496\\
2	0.839196059175472\\
2	0.844221252754447\\
2	0.849246446333423\\
2	0.854271639912399\\
2	0.859296833491374\\
2	0.86432202707035\\
2	0.869347220649326\\
2	0.874372414228301\\
2	0.879397044381604\\
2	0.884422237960579\\
2	0.889447431539555\\
2	0.894472625118531\\
2	0.899497818697506\\
2	0.904523012276482\\
2	0.909548205855458\\
2	0.914573399434433\\
2	0.919598029587736\\
2	0.924623223166711\\
2	0.929648416745687\\
2	0.934673610324663\\
2	0.939698803903638\\
2	0.944723997482614\\
2	0.94974919106159\\
2	0.954774384640565\\
2	0.959799014793868\\
2	0.964824208372844\\
2	0.969849401951819\\
2	0.974874595530795\\
2	0.97989978910977\\
2	0.984924982688746\\
2	0.989950176267722\\
2	0.994975369846698\\
2	1\\
};
\addlegendentry{BCICP \cite{ren2018orientation}}

\addplot [color=mycolor1, line width=2.0pt]
  table[row sep=crcr]{%
1.42168473882265e-07	0\\
0.00384253861127526	0.00502519357897566\\
0.0080030871923733	0.0100503871579513\\
0.0124682271530587	0.015075580736927\\
0.0170022071926939	0.0201007743159026\\
0.0217134617184001	0.0251259678948783\\
0.0265806266621977	0.0301511614738539\\
0.0315582835537196	0.0351763550528296\\
0.0366230153563958	0.0402015486318053\\
0.0417693761481526	0.0452261787851078\\
0.0471010803552869	0.0502513723640834\\
0.0526422713945713	0.0552765659430591\\
0.0581957090457328	0.0603017595220347\\
0.0638682391286456	0.0653269531010104\\
0.0696501660913342	0.070352146679986\\
0.0757965808614203	0.0753773402589617\\
0.0818738306767344	0.0804025338379374\\
0.0881694890980071	0.0854271639912399\\
0.0943680450792721	0.0904523575702155\\
0.100696258747842	0.0954775511491912\\
0.10735640540922	0.100502744728167\\
0.114003361691351	0.105527938307142\\
0.120961019274902	0.110553131886118\\
0.127931606921627	0.115578325465094\\
0.135248041037197	0.120603519044069\\
0.142463270052558	0.125628149197372\\
0.149765461548716	0.130653342776348\\
0.157383221503977	0.135678536355323\\
0.165089923107106	0.140703729934299\\
0.17304863826095	0.145728923513275\\
0.1810503751336	0.15075411709225\\
0.189182878072486	0.155779310671226\\
0.197460642848605	0.160804504250202\\
0.20588035254408	0.165829697829177\\
0.214417025324274	0.17085432798248\\
0.223248541008771	0.175879521561455\\
0.232037711134363	0.180904715140431\\
0.241210125946196	0.185929908719407\\
0.25054029529301	0.190955102298382\\
0.260001627280414	0.195980295877358\\
0.269488150427527	0.201005489456334\\
0.279232071088893	0.206030683035309\\
0.289212858989359	0.211055313188612\\
0.299336386913399	0.216080506767587\\
0.309778343140015	0.221105700346563\\
0.320298473947532	0.226130893925539\\
0.331023368908977	0.231156087504514\\
0.342034927532203	0.23618128108349\\
0.353020828717521	0.241206474662466\\
0.364213453988485	0.246231668241441\\
0.375935586390834	0.251256298394744\\
0.387903234690693	0.25628149197372\\
0.399897813508877	0.261306685552695\\
0.411982700115927	0.266331879131671\\
0.424684044326789	0.271357072710647\\
0.437388261213526	0.276382266289622\\
0.450314139060856	0.281407459868598\\
0.463591959246997	0.286432653447574\\
0.47700513532082	0.291457847026549\\
0.490691447421548	0.296482477179852\\
0.504793684854836	0.301507670758827\\
0.518745791579164	0.306532864337803\\
0.533446073415423	0.311558057916779\\
0.548065771722207	0.316583251495754\\
0.563462315027539	0.32160844507473\\
0.579161595593258	0.326633638653706\\
0.595164186330225	0.331658832232681\\
0.61157208881278	0.336683462385984\\
0.628173943182757	0.341708655964959\\
0.645029957892835	0.346733849543935\\
0.662414304063143	0.351759043122911\\
0.679818997727737	0.356784236701886\\
0.69813842905954	0.361809430280862\\
0.716661202712294	0.366834623859838\\
0.735567403063648	0.371859817438813\\
0.755042979407038	0.376884447592116\\
0.775152781848001	0.381909641171092\\
0.795583604573332	0.386934834750067\\
0.816501032686119	0.391960028329043\\
0.837912090291913	0.396985221908018\\
0.859803160428291	0.402010415486994\\
0.882619639689179	0.40703560906597\\
0.905865644421771	0.412060802644945\\
0.929846285668614	0.417085432798248\\
0.954023754682406	0.422110626377224\\
0.978644047383723	0.427135819956199\\
1.00499058774383	0.432161013535175\\
1.03197250027092	0.437186207114151\\
1.05891527633756	0.442211400693126\\
1.08754231449396	0.447236594272102\\
1.11683463852862	0.452261787851078\\
1.14700470402927	0.457286981430053\\
1.17868004683462	0.462311611583356\\
1.21078716617276	0.467336805162331\\
1.24411289132597	0.472361998741307\\
1.27804908866625	0.477387192320283\\
1.313812301995	0.482412385899258\\
1.35066441441888	0.487437579478234\\
1.38946298868371	0.49246277305721\\
1.43013018795521	0.497487966636185\\
1.472301534933	0.502512596789488\\
1.51599742332281	0.507537790368463\\
1.56198415991603	0.512562983947439\\
1.60961043390445	0.517588177526415\\
1.65867842905493	0.52261337110539\\
1.71065564048347	0.527638564684366\\
1.76531944199543	0.532663758263342\\
1.82248563120529	0.537688951842317\\
1.8827550232179	0.54271358199562\\
1.94505978449832	0.547738775574596\\
2	0.552763969153571\\
2	0.557789162732547\\
2	0.562814356311523\\
2	0.567839549890498\\
2	0.572864743469474\\
2	0.577889937048449\\
2	0.582915130627425\\
2	0.587939760780728\\
2	0.592964954359703\\
2	0.597990147938679\\
2	0.603015341517655\\
2	0.60804053509663\\
2	0.613065728675606\\
2	0.618090922254582\\
2	0.623116115833557\\
2	0.62814074598686\\
2	0.633165939565835\\
2	0.638191133144811\\
2	0.643216326723787\\
2	0.648241520302762\\
2	0.653266713881738\\
2	0.658291907460714\\
2	0.663317101039689\\
2	0.668341731192992\\
2	0.673366924771968\\
2	0.678392118350943\\
2	0.683417311929919\\
2	0.688442505508895\\
2	0.69346769908787\\
2	0.698492892666846\\
2	0.703518086245821\\
2	0.708542716399124\\
2	0.7135679099781\\
2	0.718593103557075\\
2	0.723618297136051\\
2	0.728643490715027\\
2	0.733668684294002\\
2	0.738693877872978\\
2	0.743719071451954\\
2	0.748744265030929\\
2	0.753768895184232\\
2	0.758794088763207\\
2	0.763819282342183\\
2	0.768844475921159\\
2	0.773869669500134\\
2	0.77889486307911\\
2	0.783920056658086\\
2	0.788945250237061\\
2	0.793969880390364\\
2	0.79899507396934\\
2	0.804020267548315\\
2	0.809045461127291\\
2	0.814070654706266\\
2	0.819095848285242\\
2	0.824121041864218\\
2	0.829146235443193\\
2	0.834170865596496\\
2	0.839196059175472\\
2	0.844221252754447\\
2	0.849246446333423\\
2	0.854271639912399\\
2	0.859296833491374\\
2	0.86432202707035\\
2	0.869347220649326\\
2	0.874372414228301\\
2	0.879397044381604\\
2	0.884422237960579\\
2	0.889447431539555\\
2	0.894472625118531\\
2	0.899497818697506\\
2	0.904523012276482\\
2	0.909548205855458\\
2	0.914573399434433\\
2	0.919598029587736\\
2	0.924623223166711\\
2	0.929648416745687\\
2	0.934673610324663\\
2	0.939698803903638\\
2	0.944723997482614\\
2	0.94974919106159\\
2	0.954774384640565\\
2	0.959799014793868\\
2	0.964824208372844\\
2	0.969849401951819\\
2	0.974874595530795\\
2	0.97989978910977\\
2	0.984924982688746\\
2	0.989950176267722\\
2	0.994975369846698\\
2	1\\
};
\addlegendentry{Zoomout \cite{melzi2019zoomout}}

\end{axis}

%\begin{axis}[%
%width=12.635in,
%at={(0in,0in)},
%scale only axis,
%xmin=0,
%xmax=1,
%ymin=0,
%ymax=1,
%axis line style={draw=none},
%ticks=none,
%axis x line*=bottom,
%axis y line*=left,
%legend style={legend cell align=left, align=left, draw=white!15!black}
%]
%\end{axis}
\end{tikzpicture}%

%% file: figures/michael_runtimes.tikz
\definecolor{mycolor1}{rgb}{0.00000,1.00000,1.00000}%
\begin{tikzpicture}

\begin{axis}[%
width=0.25\linewidth,
height=0.25\linewidth,
at={(0,0)},
scale only axis,
xmode=log,
xmin=531,
xmax=52565,
xminorticks=true,
xlabel style={font=\color{white!15!black}},
xlabel={Number of vertices $N$},
ymode=log,
ymin=1,
ymax=1000000,
yminorticks=true,
ylabel style={font=\color{white!15!black}},
ylabel={Runtime},
axis background/.style={fill=white},
xmajorgrids,
xminorgrids,
ymajorgrids,
yminorgrids,
legend style={at={(0.03,0.6)}, anchor=south west, legend cell align=left, align=left, draw=white!15!black,font=\footnotesize}
]
\addplot [color=red, line width=2.0pt]
  table[row sep=crcr]{%
531	23.4421\\
749	28.751142\\
1026	32.037538\\
1375	35.403819\\
1808	42.619912\\
2337	50.08412\\
2974	59.974895\\
3731	72.155195\\
4626	87.093155\\
5672	105.773681\\
6882	129.036547\\
8280	157.933971\\
9881	193.552407\\
11703	236.974113\\
13765	290.633409\\
16091	360.896505\\
18700	443.369173\\
21614	547.343061\\
24855	667.098761\\
28445	820.783531\\
32413	1013.885574\\
36784	1239.23965\\
41580	1516.873842\\
46831	1877.613832\\
52565	2322.589896\\
};
\addlegendentry{Ours}

\addplot [color=blue, line width=2.0pt]
  table[row sep=crcr]{%
531	42.068735\\
749	48.983985\\
1026	69.938947\\
1375	130.224975\\
1808	171.974938\\
2337	267.584313\\
2974	488.289205\\
3731	460.356644\\
4626	567.492774\\
5672	881.113771\\
6882	1748.742538\\
8280	2592.061693\\
9881	6715.747306\\
11703	8344.361924\\
13765	9322.575229\\
16091	12989.698751\\
18700	18227.571345\\
21614	24870.503611\\
24855	39882.752091\\
28445	54595.270729\\
32413	58890.713086\\
36784	85813.649346\\
41580	97403.924165\\
46831	132248.173013\\
52565	176815.726157\\
};
\addlegendentry{BCICP}

\addplot [color=green, line width=2.0pt]
  table[row sep=crcr]{%
531	7.808096\\
749	40.393861\\
1026	36.564303\\
1375	22.820831\\
1808	25.184204\\
2337	38.564042\\
2974	59.202759\\
3731	103.74142\\
4626	133.123835\\
5672	157.117701\\
6882	269.345386\\
8280	276.679094\\
9881	348.535867\\
11703	747.2758695\\
13765	795.5034195\\
16091	771.983541\\
18700	865.748946\\
21614	1299.57555\\
24855	1547.13621\\
28445	2240.201181\\
32413	2421.712041\\
36784	2519.2464315\\
41580	2549.2108125\\
46831	2904.422022\\
52565	4408.733214\\
};
\addlegendentry{KM}

\addplot [color=mycolor1, line width=2.0pt]
  table[row sep=crcr]{%
531	6.378672\\
749	8.803426\\
1026	5.155134\\
1375	6.72214\\
1808	9.182955\\
2337	11.41599\\
2974	15.435126\\
3731	20.033631\\
4626	25.796376\\
5672	33.512451\\
6882	42.727163\\
8280	56.735666\\
9881	73.842474\\
11703	86.130905\\
13765	105.221119\\
16091	128.128474\\
18700	171.840975\\
21614	192.101413\\
24855	231.378318\\
28445	277.310831\\
32413	345.427879\\
36784	424.333539\\
41580	527.690346\\
46831	652.546223\\
52565	845.716182\\
};
\addlegendentry{Zoomout}

\end{axis}

\end{tikzpicture}%

%% file: figures/michael_errors.tikz
\definecolor{mycolor1}{rgb}{0.00000,1.00000,1.00000}%
\begin{tikzpicture}

\begin{axis}[%
width=0.25\linewidth,
height=0.25\linewidth,
at={(0,0)},
scale only axis,
xmode=log,
xmin=531,
xmax=52565,
xminorticks=true,
xlabel style={font=\color{white!15!black}},
xlabel={Number of vertices $N$},
ymode=log,
ymin=0.001,
ymax=1,
yminorticks=true,
ylabel style={font=\color{white!15!black}},
ylabel={Mean error},
axis background/.style={fill=white},
xmajorgrids,
xminorgrids,
ymajorgrids,
yminorgrids
]
\addplot [color=red, line width=2.0pt]
  table[row sep=crcr]{%
531	0.213451959865233\\
749	0.456342680703689\\
1026	0.0151795420837401\\
1375	0.017180745997122\\
1808	0.00833182003606327\\
2337	0.0118944146996376\\
2974	0.00935960943435102\\
3731	0.00522533208679137\\
4626	0.00489287419040221\\
5672	0.00462566357479715\\
6882	0.00836845851352045\\
8280	0.00852751430510763\\
9881	0.00832778174356993\\
11703	0.00672871205084277\\
13765	0.00563476968314634\\
16091	0.00324430585723986\\
18700	0.00473226294047524\\
21614	0.00941557332088712\\
24855	0.00779258202087824\\
28445	0.00679673965944521\\
32413	0.00891829924013035\\
36784	0.011533251507039\\
41580	0.00932301380354445\\
46831	0.0112418907205284\\
52565	0.0085538371000685\\
};

\addplot [color=blue, line width=2.0pt]
  table[row sep=crcr]{%
531	0.216280830457024\\
749	0.0179902117803898\\
1026	0.0282239485540255\\
1375	0.0227792361328724\\
1808	0.028808112372294\\
2337	0.168713959784288\\
2974	0.0307556459501246\\
3731	0.0312179380816466\\
4626	0.0213291398916308\\
5672	0.0181876601797644\\
6882	0.021292431276246\\
8280	0.0381220852797411\\
9881	0.15010426661054\\
11703	0.0161396063019281\\
13765	0.0260869260495808\\
16091	0.0156420459327324\\
18700	0.0206481108904385\\
21614	0.0139377898416742\\
24855	0.011609784724447\\
28445	0.0138385441073814\\
32413	0.0109693872249586\\
36784	0.0106359899806298\\
41580	0.0104312836912507\\
46831	0.0110916053569455\\
52565	0.0120881281674647\\
};

\addplot [color=green, line width=2.0pt]
  table[row sep=crcr]{%
531	0.0398192292540886\\
749	0.248368607370298\\
1026	0.0176083591776697\\
1375	0.219470704950858\\
1808	0.0246186751047829\\
2337	0.0188102743820919\\
2974	0.0262884569172345\\
3731	0.0133026709975743\\
4626	0.0103090409544601\\
5672	0.012144982532891\\
6882	0.0107468177005755\\
8280	0.0131605518787766\\
9881	0.00986343329626045\\
11703	0.0205021639006473\\
13765	0.00835766738062149\\
16091	0.0130114869939306\\
18700	0.010203845186433\\
21614	0.0096299520321333\\
24855	0.0130766045237416\\
28445	0.00937774513720372\\
32413	0.014210700380918\\
36784	0.0132640277488892\\
41580	0.0168809784882932\\
46831	0.0133855906253853\\
52565	0.0165323767368732\\
};

\addplot [color=mycolor1, line width=2.0pt]
  table[row sep=crcr]{%
531	0.385463563449414\\
749	0.533986496830477\\
1026	0.306181576924201\\
1375	0.161487314348578\\
1808	0.445692085036591\\
2337	0.100668553441821\\
2974	0.252213592647139\\
3731	0.0486915228358915\\
4626	0.0637084514149531\\
5672	0.0598250318192625\\
6882	0.0685740826537842\\
8280	0.081909684212381\\
9881	0.0691379510799004\\
11703	0.077571506109744\\
13765	0.0656457270436377\\
16091	0.053251490656793\\
18700	0.0865733265236391\\
21614	0.0588463220363531\\
24855	0.0687954295227358\\
28445	0.0628498090592165\\
32413	0.0722336111929021\\
36784	0.106415185501688\\
41580	0.0834021037343214\\
46831	0.0827589866954773\\
52565	0.0785422205034527\\
};

\end{axis}
\end{tikzpicture}%